\documentclass[10pt,journal,compsoc]{IEEEtran}
\ifCLASSOPTIONcompsoc
  \usepackage[nocompress]{cite}
\else
  \usepackage{cite}
\fi

\usepackage[top=0.4in,bottom=0.4in,left=0.42in,right=0.42in]{geometry}

\ifCLASSINFOpdf
\else
\fi

\usepackage{amsmath,amsfonts}
\usepackage{algorithmic}
\usepackage{algorithm}
\usepackage{array}
\usepackage[caption=false,font=normalsize,labelfont=sf,textfont=sf]{subfig}
\usepackage{textcomp}
\usepackage{stfloats}
\usepackage{url}
\usepackage{verbatim}
\usepackage{graphicx}
\usepackage{cite}
\usepackage{multirow}
\usepackage{pifont}
\usepackage{xcolor}
\usepackage{makecell}
\usepackage[colorlinks, linkcolor=red, citecolor=green]{hyperref}

\begin{document}
%
\title{Probing Deep into Temporal Profile Makes the Infrared Small Target Detector Much Better}

\author{Ruojing~Li, Wei~An, Yingqian~Wang, Xinyi~Ying, Yimian~Dai, Longguang~Wang, Miao~Li, Yulan~Guo, ~Li~Liu 
\IEEEcompsocitemizethanks{\IEEEcompsocthanksitem This paper was partially supported by the National Natural Science Foundations of China under Grants 62376283, 62401590, 42501589, and 62401591, and the Key Stone grant (JS2023-03) of the National University of Defense Technology (NUDT).
\IEEEcompsocthanksitem  R. Li, W. An, Y. Wang, X. Ying, M. Li, and L. Liu are with the College of Electronic Science and Technology, NUDT, Changsha 410073, China. Y. Dai is with Nankai University, Tianjin 300071, China. L. Wang is with Aviation University of Air Force, Changchun 130012, China. Y. Guo is with Sun Yat-Sen University, Shenzhen 518107, China. Corresponding authors are Miao Li, Yingqian Wang, and Li Liu.
}
}

\markboth{Submitted to IEEE Trans. Pattern Analysis and Machine Intelligence}%
{Li \MakeLowercase{\textit{et al.}}: Probing Deep into Temporal Profile Makes the Infrared Small Target Detector Much Better}

\IEEEtitleabstractindextext{%
\begin{abstract}
Infrared small target (IRST) detection is challenging in simultaneously achieving precise, robust, and efficient performance due to extremely dim targets and strong interference. Current learning-based methods attempt to leverage ``more'' information from both the spatial and the short-term temporal domains, but suffer from unreliable performance under complex conditions while incurring computational redundancy. In this paper, we explore the ``more essential'' information from a more crucial domain for the detection. Through theoretical analysis, we reveal that the global temporal saliency and correlation information in the temporal profile demonstrate significant superiority in distinguishing target signals from other signals. To investigate whether such superiority is preferentially leveraged by well-trained networks, we built the first prediction attribution tool in this field and verified the importance of the temporal profile information. Inspired by the above conclusions, we remodel the IRST detection task as a one-dimensional signal anomaly detection task, and propose an efficient deep temporal probe network (DeepPro) that only performs calculations in the time dimension for IRST detection. We conducted extensive experiments to fully validate the effectiveness of our method. The experimental results are exciting, as our DeepPro outperforms existing state-of-the-art IRST detection methods on widely-used benchmarks with extremely high efficiency, and achieves a significant improvement on dim targets and in complex scenarios. We provide a new modeling domain, a new insight, a new method, and a new performance, which can promote the development of IRST detection. Codes are available at \href{https://tinalrj.github.io/DeepPro/}{https://tinalrj.github.io/DeepPro/}.
\end{abstract}

\begin{IEEEkeywords}
Infrared small target detection, temporal profile, anomaly detection, remote sensing object detection, temporal probes, attribution analysis.
\end{IEEEkeywords}}

\maketitle

\IEEEdisplaynontitleabstractindextext
\IEEEpeerreviewmaketitle

\IEEEraisesectionheading{\section{Introduction}\label{sec:introduction}}
The objective of Infrared Small Target (IRST) detection \cite{chapple1999target, kou2023infrared} is to identify and localize moving targets that are typically ``small'' and ``dim''\footnote{``Small'' means the target usually occupies a small portion of the overall image, and ``dim'' indicates that the target exhibits a weak signal intensity and low contrast relative to its background.} from infrared images. It has been a longstanding, fundamental yet challenging problem in infrared remote sensing \cite{wang2024robust, ying2025infrared}, and has broad applicability in both civilian and military domains, including reconnaissance, precision guidance \cite{sun2020infrared, li2023direction}, maritime surveillance \cite{li2023sparse, hou2024unsupervised}, early fire warnings \cite{hua2017progress, zhao2022single}, vehicle detection \cite{wang2022review, sun2024multi}, and UAV monitoring \cite{zhang2022visible, huang2023anti}. In recent years, IRST detection has attracted increasing attention and made measurable progress over the past few decades; however, most IRST detection methods have not been capable of simultaneously achieving outstanding performance in \emph{precision}, \emph{robustness}, and \emph{efficiency} due to the following challenges \cite{zhao2022single, cheng2023towards, li2024mixed, yang2025deep}. 

\begin{figure*}[!t]
    \centering
    \includegraphics[width=1\linewidth]{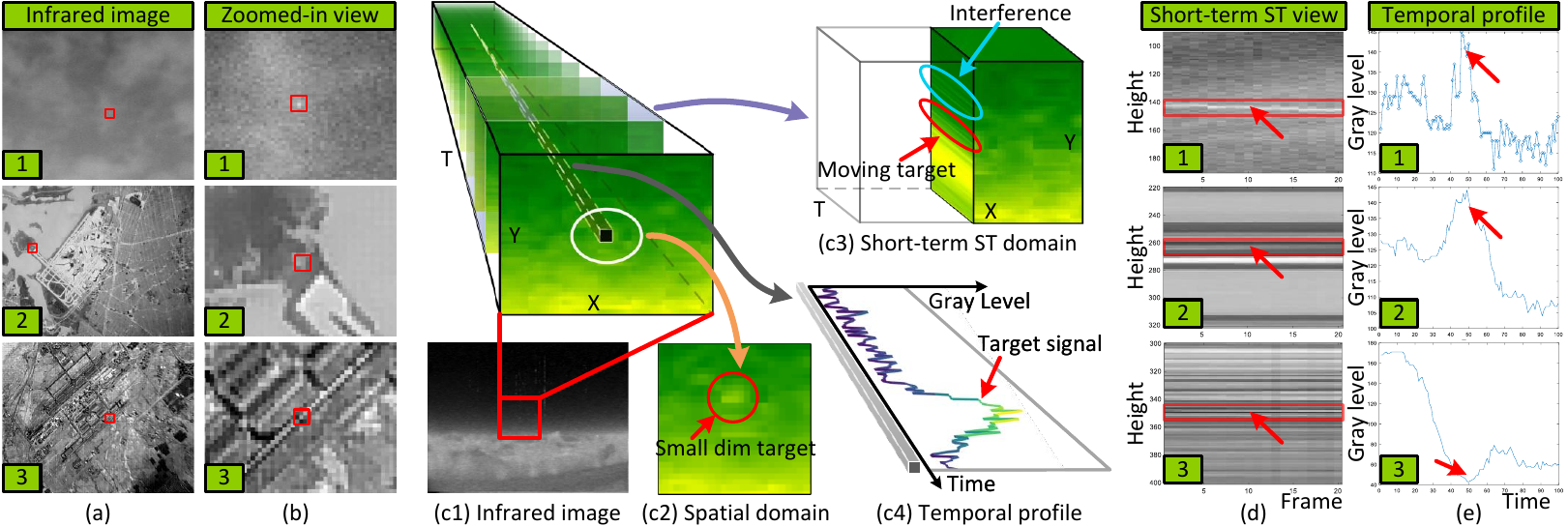}
    \caption{Visualizations of IRSTs in different scenarios and typical detection difficulties in different domains. (a, b) Distinct targets in different scenarios (Cases 1, 2, and 3): in appearance, targets and clutters show small inter-class differences, while different targets exhibit marked intra-class variations. (c1) Infrared image. (c2) Spatial domain: small dim targets are barely observable. (c3) Short-term spatial-temporal (ST) domain: targets are indistinguishable from interferences. (c4) Temporal profile: records dynamic statistical details of all signals at fixed spatial location, where small dim target signals are complete and more prominent. The temporal profile contains more essential information. (d, e) More contrasts of the three cases. For clarity, targets are marked with red annotations. Corresponding dynamic demos are available at \href{https://tinalrj.github.io/DeepPro/}{https://tinalrj.github.io/DeepPro/}. 
    }
	\label{Fig1}
\end{figure*}

\begin{itemize}
  \item \emph{High-precision related challenges.} On the one hand, the targets themselves are usually very small (e.g., typically occupy $\leq9\times 9$ pixels), dim (e.g., with an SNR $\leq 3$) \cite{li2023direction, liu2023infrared, fan2024diffusion}, and with insufficient appearance features, such as shape and texture information. Such characteristics increase the difficulty of developing effective feature representations. On the other hand, the background is usually nonsmooth and nonuniform, containing clutter and noise. Clutter refers to imaged real things, while noise refers to electronic noise and inaccuracies of the signal processor. They almost occupy the entire image, and some of them are strong and dynamic, similar to the targets as shown in Fig. \ref{Fig1}(a) and (b). This complicates the effective suppression of false alarms.
  \item \emph{High-robustness related challenges.} Apart from the aforementioned challenges, the target characteristics evolve temporally under motion conditions. The intensity of the target varies, and the local background neighborhood of the target also changes dynamically. Furthermore, practical scenarios involve diverse complex backgrounds with distinct clutter properties and various targets with different grayscale distributions, as shown in Fig. \ref{Fig1}(a) and (b). These complex factors make it difficult to detect targets continuously and stably.
  \item \emph{High-efficiency related challenges.} The application of IRST detection requires real-time processing of massive data streams, while IRST is extremely sparse in the spatial domain. It is difficult to balance sufficient feature extraction with efficient computation. 
\end{itemize}

Therefore, in diverse complex scenarios, \textit{which modeling domain can better exhibit the targets? Which prior information can more effectively capture the essential differences between target and background?} Exploring these questions is critical to achieve precise, robust, and efficient detection.

Existing deep learning-based methods focus on extracting ``more'' information to improve the detection performance. For example, many single-frame methods \cite{liu2023infrared, yuan2024sctransnet, zhang2025irmamba} extensively explore the local saliency and global information in the spatial domain via the complex network structure.
Besides, some multi-frame methods \cite{du2021spatial, yan2023stdmanet, tong2024st, zhang2025mocid} explore motion information in the short-term spatial-temporal domain. Although performance improvements are achieved in some scenarios, these methods suffer from severe performance degradation under complex conditions. This comes from the limitation that the target information is weak and that other discriminative information is insufficient in both domains, as illustrated in Fig. \ref{Fig1}(c2) and (c3). In addition, extracting ``more'' information inevitably expands receptive fields, causing severe computational redundancy that fundamentally undermines real-time processing capabilities. Therefore, given our previous analysis, the focus of this work is to explore the ``more essential'' information from a more crucial profile and develop a highly effective and efficient method for IRST detection, especially under complex conditions.

In this paper, we study the inherent characteristics of IRST detection task in depth. We reveal that the temporal profile is crucial especially under complex conditions, which dramatically amplifies target saliency as shown in Fig. \ref{Fig1}(c4) and (e). Through further theoretical analysis, we find that the temporal profile contains global temporal saliency information of target signals and correlation information between different signals, which are more essential for IRST detection. To validate the importance of the temporal profile in well-trained networks, we built the first prediction attribution \cite{sundararajan2017axiomatic, xu2020attribution} tool that can analyze which input pixels impact the predictions of IRST detection networks. The attribution results show that the most influential pixels are concentrated in a small area centered on the target and continuously distributed along the time axis, which illustrates the importance of the temporal profile. 

\begin{figure}[!t]
    \vspace{-5pt}
    \centering
    \includegraphics[width=1\linewidth]{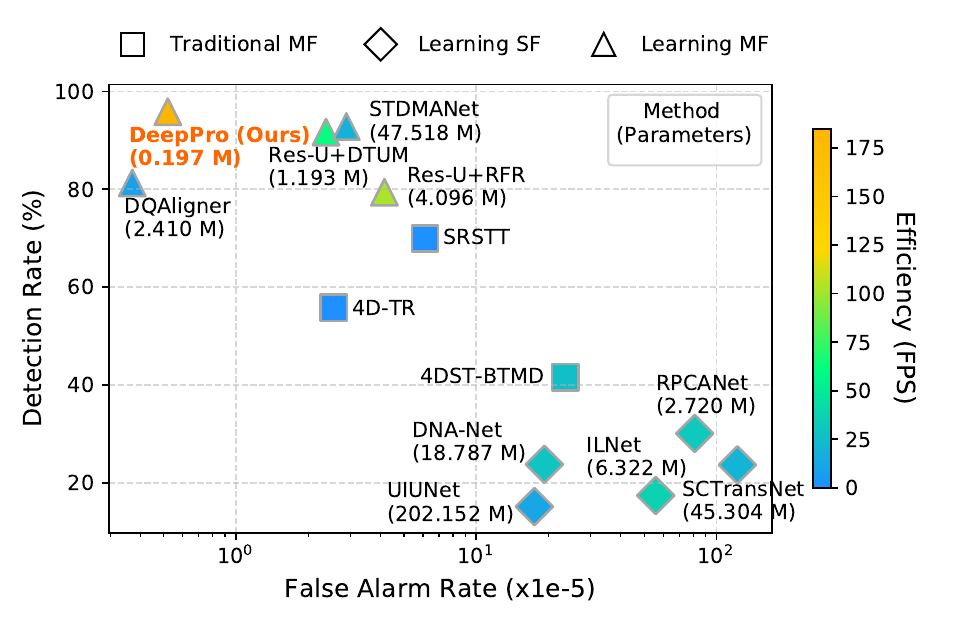}
    \vspace{-20pt}
    \caption{Comparison of the detection performance, computational efficiency (Frames Per Second, FPS), and parameter counts of different deep learning-based methods on dim targets ($SNR \leq 3$) in the NUDT-MIRSDT dataset. The color of each symbol represents the computational efficiency of the method. The yellower the color, the higher efficiency the method holds. Our proposed method is highly efficient and effective compared with other methods, including traditional multi-frame (MF) methods (SRSTT \cite{li2023sparse}, 4DST-BTMD \cite{luo20234dst}, and 4D-TR \cite{wu2023infrared}), deep learning-based single-frame (SF) methods (DNA-Net \cite{li2022dense}, UIUNet \cite{wu2022uiu}, SCTransNet \cite{yuan2024sctransnet}, RPCANet \cite{wu2024rpcanet}, and ILNet \cite{li2025ilnet}), and deep learning-based multi-frame methods (Res-UNet+DTUM \cite{li2023direction}, STDMANet \cite{yan2023stdmanet}, Res-UNet+RFR \cite{ying2025infrared},and DQAligner \cite{deng2026learning}).}
	\label{Fig2}
\end{figure}

Inspired by the above research, we remodel the IRST detection task as a one-dimensional signal anomaly detection task in the temporal profile, where targets manifest as statistical outliers in sequential signals. Then, we design a pixel-wise temporal probe mechanism (TPro) to extract essential information in the temporal profile. Based on TPro, we propose an effective and efficient deep temporal probe network (i.e., DeepPro) for IRST detection, which performs multiplications and additions only in the time dimension. We conducted numerous experiments on widely-used benchmarks. As shown in Fig. \ref{Fig2}, our DeepPro achieves significantly superior performance as compared to existing state-of-the-art methods employing complex spatial-temporal computations. Meanwhile, our DeepPro exhibits higher robustness on dim targets and in complex scenarios with strong noise, achieving an 87.6\% reduction in parameter count and a 13-FPS inference gain on $256 \times 256 $ images compared to the existing lightest and second-fastest single-frame IRST (SIRST) detection network (i.e., ACM \cite{dai2021attentional}).

The main contributions of this work are as follows.
\begin{itemize}
    \item We reveal some new insights from a more crucial profile (i.e., temporal profile): long-term temporal information is much more essential for IRST detection, which includes global temporal saliency of target signals and correlation information between different signals. We validated the importance of the temporal profile in IRST detection, by developing the first predictive attribution tool.
    \item Inspired by our research, we remodel the IRST detection task as a one-dimensional signal anomaly detection task. Then, we propose DeepPro to leverage the essential temporal profile information with multiply-add operations only in the time dimension.
    \item Experimental results show that DeepPro achieves a significant performance improvement in IRST detection with extremely high efficiency, and has high robustness to dim targets and scenes with strong noise.
\end{itemize}

\section{Related work}
In this section, we briefly review the major works in SIRST detection, multi-frame IRST (MIRST) detection, and gradient-based attribution for deep networks.

\subsection{Single-frame infrared small target detection}
Traditional SIRST detection methods mainly include spatial-filtering-based methods \cite{soni1993performance, deshpande1999max, bai2010analysis, zhu2020balanced}, sparsity-and-low-rank-based methods \cite{2013Infrared_IPI, dai2017reweighted_RIPT, 2018Infrared_NRAM, zhao2021three}, and human-visual-system-based (HVS) methods \cite{chen2013local, 2015Infrared_PLCM, liu2018infrared, yang2023small}. Recently, deep learning-based methods achieved better detection performance than traditional ones due to their strong feature learning abilities. Drawing inspiration from traditional methods, some networks are integrated with domain-specific priors for robust SIRST detection, such as the local spatial saliency differences between targets and backgrounds \cite{dai2021attentional, zhang2022learning, yang2024pinwheel}, the sparsity of targets and the low rankness of backgrounds \cite{zhou2023deep, wu2024rpcanet, liu2024infrared}. Dai et al. \cite{dai2021attentional} incorporated local contrast measure \cite{wei2016multiscale} into the network. Zhang et al. \cite{zhang2022learning} converted the local contrast measure into a nonlocal quadrature difference measure in deep feature space. Zhou et al. \cite{zhou2023deep} developed a deep unrolling network to exploit the low-rankness of backgrounds and the sparsity of targets. Liu et at. \cite{liu2024infrared} combined low rankness and local smoothness prior of IRSTs to better model the background. Wu et al. \cite{wu2024rpcanet} formulated the IRST detection task as sparse target extraction, low-rank background estimation, and image reconstruction in a relaxed robust principle component analysis model. For performance improvements, some methods combine IRST properties with the attention mechanism \cite{vaswani2017attention} to learn correlations among pixels and pay attention to key features. Li et al. \cite{li2022dense} cascaded channel and spatial attention module to enhance multi-level features. Zhang et al. \cite{zhang2022isnet} proposed a Taylor finite difference-inspired edge block and a two-orientation attention aggregation block to highlight the shape of the targets. Zhang et al. \cite{zhang2023dim2clear} introduced a context mixer decoder based on spatial and frequency attention to modulate the low-level features for feature representation enhancement. Liu et al. \cite{liu2023infrared} adopted the self-attention mechanism of Transformer to learn the non-local correlation of image features. Beside feature extraction, Liu et al. \cite{liu2024infrared} proposed a scale and location sensitive loss function to boost detection performance. Zhang et al. \cite{zhang2025saist} integrated textual information with image modalities to enhance IRSTD performance. Zhang et al. \cite{zhang2024irprunedet} proposed a wavelet-based pruning method for efficient detection. Li et al. \cite{li2026dynamic} enabled the convolution to explicitly and differentially model high-frequency components. Moreover, some researchers investigated the weakly-supervised detection \cite{ying2023mapping, li2023monte, zhang2025semi}, in which the pseudo-masks are generated based on the local saliency of the IRST. SIRST detection is applied to locate the salient targets, but is not suitable for targets with low SNR.

\subsection{Multi-frame infrared small target detection}
MIRST detection aims at exploiting both spatial and temporal information from input image tensors, and is expected to achieve better and more robust performance than SIRST detection. Existing MIRST detection methods can be categorized into filtering-based methods \cite{reed1983application, reed1990recursive, gao2017tvpcf, li2021infrared, tzannes2002detecting, liu2015moving, niu2024high}, sparsity-and-low-rank-based methods \cite{zhang2020edge, wang2021infrared, luo2022imnn, wu2023infrared, li2023sparse, luo2023spatial, liu2025graph}, and deep learning-based methods \cite{sun2021small, yan2023stdmanet, li2023direction, tong2024st, zhang2024explore, xiao2024highly, wei2020end}.

Filtering-based methods are the earlier developed methods, which mainly uses the temporal information among frames. Reed et al. \cite{reed1983application, reed1990recursive} proposed a series of 3D matched filtering methods to highlight the targets through spatial filtering and temporal energy accumulation. Gao et al. \cite{gao2017tvpcf} proposed a time variance and patch contrast filter to improve the local contrast of targets. Li et al. \cite{li2021infrared} achieved MIRST detection by using the directional morphological filtering and spatial-temporal semantic information (e.g., the consistency of targets and the fluctuation of clutters). Mooney et al. \cite{mooney1995point} and Silverman et al. \cite{silverman1996temporal} early recognized the effectiveness of temporal profile filtering for detecting weak point IRSTs, making the ``temporal profile'' concept popular in IRST detection. Building on this foundation, some temporal profile-based filters \cite{tzannes2002detecting, liu2015moving, niu2024high} have been developed. These filters offer advantages in both detection performance and computational cost compared to other filtering-based methods.

Sparsity-and-low-rank-based methods decompose the low-rank input tensors, which is developed from the corresponding methods in SIRST detection, by further considering the time dimension. Sun et al. \cite{sun2020infrared} extended the spatial matrix to a spatial-temporal tensor to fit highly heterogeneous scenes. Wu et al. \cite{wu2023infrared} introduced a 4-D infrared image patch tensor structure to accommodate both local and global features in the space and time dimension. Liu et al. \cite{liu2025graph} proposed graph Laplacian regularization to describe the low rankness of backgrounds, which improved the speed of the detection and the robustness of the method.

Deep learning-based methods mainly include spatial-temporal saliency feature-based methods \cite{du2021spatial, yan2023stdmanet, tong2024st} and motion feature-based methods \cite{sun2021small, guo2023small, li2023direction, zhang2024explore}. Spatial-temporal saliency feature-based methods mainly utilize the saliency information of targets in a single frame and the difference information between two frames. Du et al. \cite{du2021spatial} built an end-to-end spatial-temporal feature extraction and target detection framework based on an energy accumulation enhancement mechanism between frames. Yan et al.\cite{yan2023stdmanet} proposed to use a temporal multiscale feature extractor to obtain spatial-temporal features from multiple time scales. Tong et al. \cite{tong2024st} established the spatial-temporal dependencies among frames by using temporal information. Motion feature-based methods improve the IRST detection performance especially in suppressing false alarms by using the continuity of the target motion in the space-time dimension. Sun et al. \cite{sun2021small} modeled the local smoothness and global continuous characteristic of the target trajectory as short-strict and long-loose constraints, to detect the true targets from numerous target candidates. Guo et al. \cite{guo2023small} applied a trained classification convolution neural network to identify the trajectory mask images, and linked adjacent trajectory segments physically. The above methods leverage the motion features only in track association, but not in target candidates detection, which increases the difficulty of trajectory management. Then, Li et al. \cite{li2023direction} designed a direction-coded convolution block to encode the motion direction into features and extract the motion information of targets for MIRST detection. Zhang et al. \cite{zhang2024explore} proposed a hybrid modeling method with smoothed-particle hydrodynamics for motion trajectory simulation and Markov decision processes for optimal action selection. In conclusion, the above methods are based on the spatial and temporal saliency information and motion information. However, these information are weak and broken under extremely complex conditions (e.g., dim targets and strong interference), resulting in failed detection. Moreover, there are amounts of computational redundancy to extract this information due to the large receptive field of the network and the sparsity of the target.

\subsection{Gradient-based attribution for deep networks}
Attribution \cite{abhishek2022attribution, wang2024gradient} is an approach of network interpretation to understand the predictions given by the models, through highlighting input features (pixels in the case of image input) that strongly influence the network outputs. Gradient-based attribution \cite{gevrey2003review, simonyan2014deep, zeiler2014visualizing, shrikumar2017learning} is a typical branch of attribution, which exploits the information flow pathway in deep networks and generate saliency maps \cite{ancona2019gradient}. Here, we review two major types of approaches, i.e., vanilla gradient-based methods \cite{gevrey2003review, simonyan2014deep, zeiler2014visualizing, springenberg2014striving, zhou2016learning, selvaraju2017grad} and integrated gradient-based methods \cite{sundararajan2017axiomatic, xu2020attribution, sturmfels2020visualizing, miglani2020investigating}.

Vanilla gradient-based methods explain the models by directly using gradients. Gevrey et al. \cite{gevrey2003review} and Simonyan et al. \cite{simonyan2014deep} proposed to consider both the gradients and their variants to explain the models. Zeiler et al. \cite{zeiler2014visualizing} proposed the deconvolutional network (DeconvNet) to approximately visualize the input stimuli that excite individual feature maps in a specified layer. Considering that DeconvNet performs poorly without max-pooling layers, Springenberg et al. \cite{springenberg2014striving} proposed the guided backpropagation algorithm including an additional guidance signal from the higher layers. Zhou et al. \cite{zhou2016learning} developed class activation mapping (CAM) to identify discriminative regions for predictions via the linearly weighted summation of activation maps from the last convolution layer. Then, Selvaraju et al. \cite{selvaraju2017grad} introduced gradient-weighted CAM (Grad-CAM) to use the gradients w.r.t. the activation map to measure the channel importance. Based on Grad-CAM, a series of methods were proposed, e.g., Grad-CAM++ \cite{chattopadhay2018grad}, Smooth Grad-CAM++ \cite{omeiza2019smooth}. However, this type of methods may fail in some cases due to the saturation issue of the gradients \cite{sundararajan2017axiomatic, sturmfels2020visualizing}.

Integrated gradient-based methods explain the models by accumulating gradients along a specified path from a baseline \cite{shrikumar2016not, shrikumar2017learning, binder2016layer} to the input pixels. Sundararajan et al. \cite{sundararajan2017axiomatic} first proposed axiomatic attribution (i.e., integrated gradients) to address the problem of gradient saturation. Xu et al. \cite{xu2020attribution} proposed the blur integrated gradients to explain the prediction by accumulating gradients from both the frequency and space domains. Sturmfels et al. \cite{sturmfels2020visualizing} proposed alternative choices for baselines in integrated gradients, considering that the all-zero baseline is unsuitable for black targets. Miglani et al. \cite{miglani2020investigating} observed that the accumulated gradients within the saturation regions also have a non-negligible contribution on saliency maps, but not on improving the model prediction.

However, all the above methods are aimed at classification tasks, and only a few works consider explaining dense-prediction networks \cite{hoyer2019grid, gu2021interpreting, chrabaszcz2024aggregated}. Hoyer et al. \cite{hoyer2019grid} proposed the first attribution method to generate grid saliency maps for dense prediction networks. Gu et al. \cite{gu2021interpreting} succeeded in applying the integrated gradients to super-resolution networks by building suitable baseline images and a gradient detector. Chrabaszcz et al. \cite{chrabaszcz2024aggregated} introduced $\rm AGG^2 EXP$ to aggregate fine-grained voxel attributions of 3D segmentation models. However, the attribution of IRST detection networks lacks exploration.

\section{Methodology}
\label{method}
In this section, we present the more essential information for IRST detection especially under complex conditions, and introduce our deep temporal probe network in detail. We first analyze the temporal profile information supporting the detection, and then verify the importance of the information through building the first attribution tool and applying it to IRST detection networks. Finally, we describe our temporal probe mechanism and our DeepPro with only temporal calculations.

\begin{figure}[!t]
    \centering
    \includegraphics[width=1\linewidth]{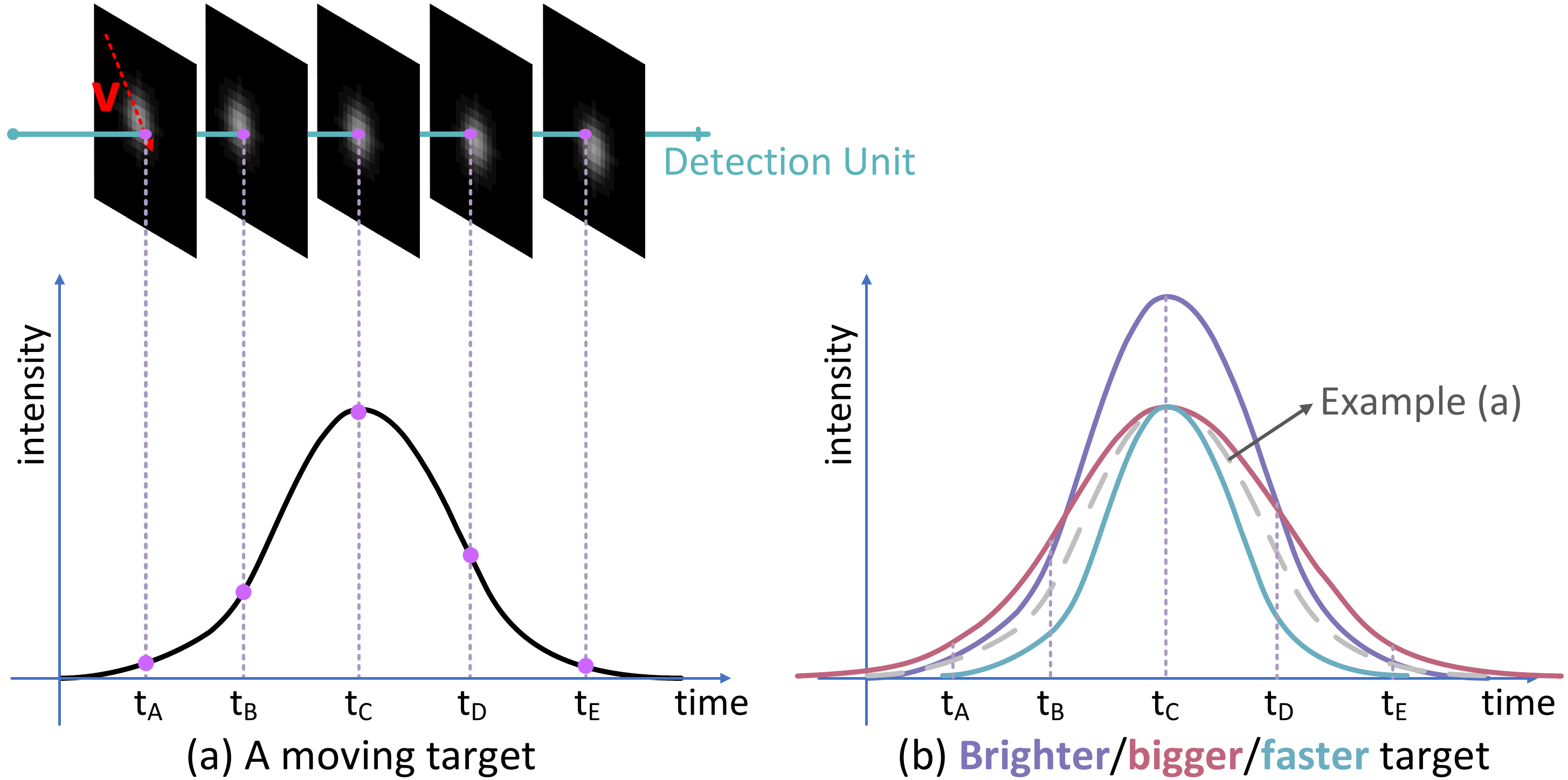}
    \caption{Toy examples of the temporal profiles of different targets.
    }
    \label{fig:toy_examples}
\end{figure}

\subsection{Temporal profile information}
\label{Temporal-profile-analysis}
Given the loss of spatial saliency, local temporal saliency, and motion characteristics of some IRSTs, we explore essential statistical information at the signal level to support target predictions. In this part, we shift focus from the spatial domain and the short-term spatial-temporal domain to the temporal profile. Then, we delve into what information is in the temporal profile and how it supports the detection of IRST.

\subsubsection{Global temporal saliency}
We analyzed the waveform characteristics of target signals in the temporal profile. Given that the velocity of small targets exhibits minimal variation over the short time they pass through a detection unit completely, we modeled the moving targets in uniform linear motions. In Fig. \ref{fig:toy_examples}, we display some toy examples of the temporal profiles of the targets with different intensities, sizes, and speeds. The fluctuation of a target signal $G$ is regular with a single up-down process. As a target moves into and out of a detection unit, the target intensity in this unit first increases and then decreases. The amplitude of $G$ is related to the maximum intensity of the part of the target passing through the detection unit, and a brighter target usually has a higher peak value. Besides, the width of $G$ is related to the size and speed of the target. Bigger targets typically exhibit wider signals in the temporal profile, while targets with higher speeds under the same conditions produce narrower signals. Therefore, assuming that the fluctuations of other signals can be ignored, the target signals are saliency in the temporal profile.

\begin{figure}[!t]
    \centering
    \includegraphics[width=1\linewidth]{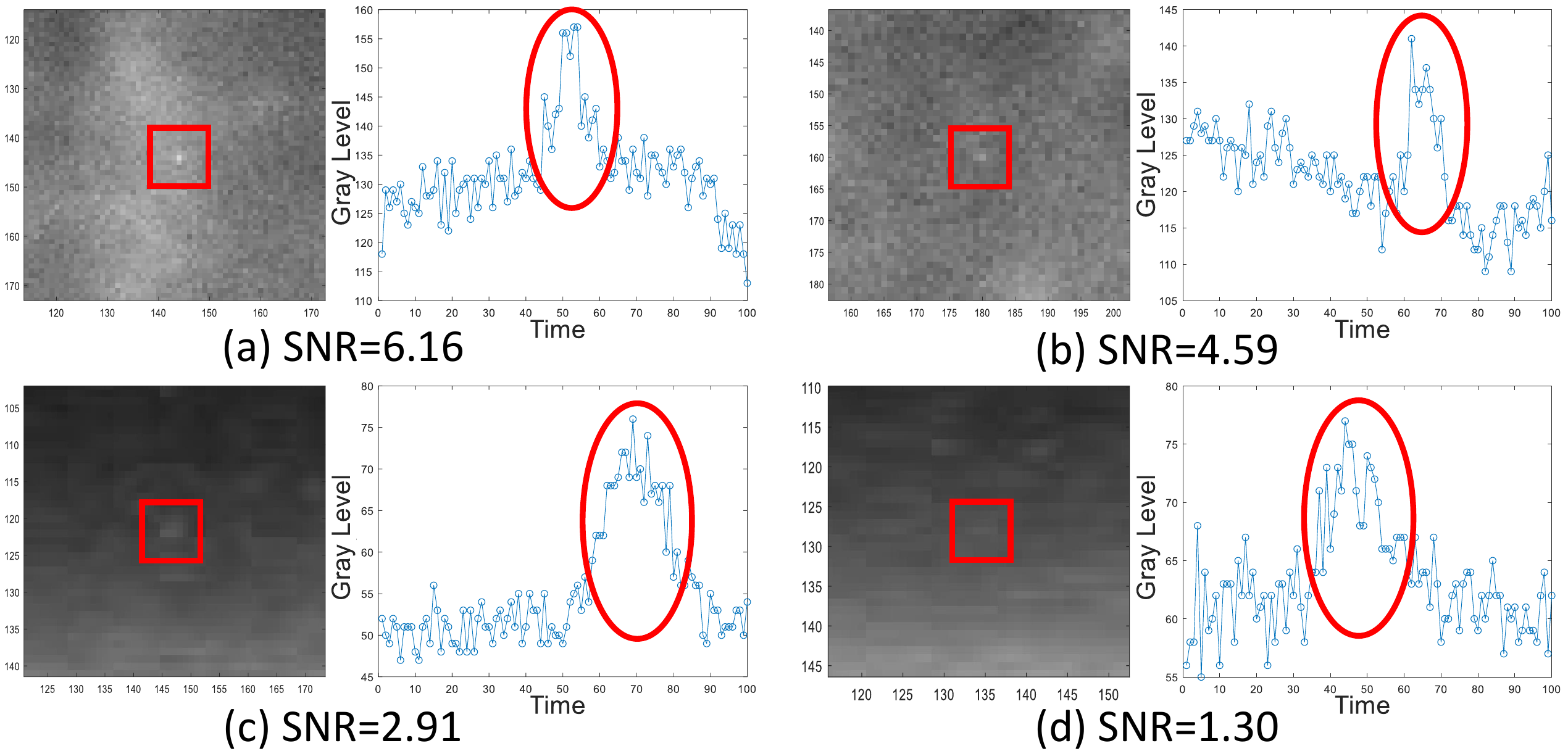}
    \caption{Spatial and temporal-profile visualizations of targets in real scenes with noise and clutter. The red annotations mark the targets.}
    \label{fig:temporal_profile_analysis_real}
\end{figure}

To investigate whether the saliency remains in the practical complex scenarios, we visualized the temporal profiles of real target signals with different SNR values, and compared them with the spatial domains of the same targets. The visualizations are shown in Fig. \ref{fig:temporal_profile_analysis_real}. It can be seen that the fluctuations of the target signals are consistent with those of the toy examples, i.e., first increasing and then decreasing. When the target SNR is higher than 3, the target is significant in both the spatial domain and the temporal profile. However, when the target SNR is less than 3, the target is so dim that it is barely visible in the spatial domain, but is still prominent in the temporal profile. This demonstrates that \textit{the temporal profile holds the global temporal saliency of the target signals, and is more robust to distinguish the targets from strong noise and clutter than other domains}.

\subsubsection{Correlation information}
\begin{figure*}[!t]
    \centering
    \includegraphics[width=1\linewidth]{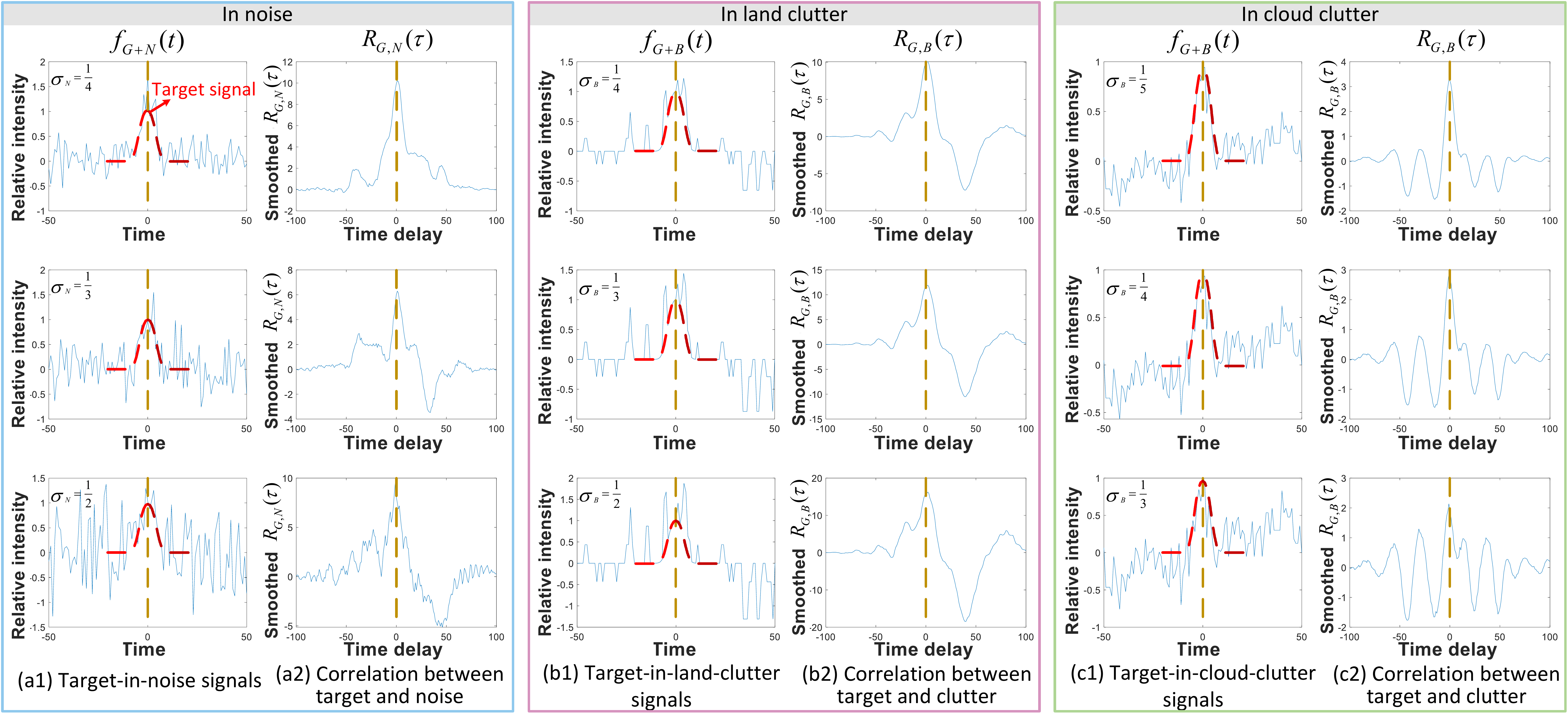}
    \caption{Visualizations of the temporal profiles of target-in-noise(/-clutter) signals and correlation analyses between noise, clutter and target in the temporal profile. The target signals are same with fixed intensity (maximum is 1), and added to noise and clutter signals with different standard deviations (i.e., $\sigma_{N}$ and $\sigma_{B}$, respectively).
    }
    \label{fig:temporal_profile_analysis_corr}
\end{figure*}
As the intensities of noise and clutter signals increase, the global temporal saliency becomes weak and damaged (e.g., Fig. \ref{fig:temporal_profile_analysis_real}(d)). Therefore, we deeply extract other essential distinctive characteristics between noise $N$, clutter $B$, and target $G$ in the temporal profile.

Firstly, to further analyze the differences between target and noise signals, we mixed target and noise, and analyzed the correlation function of the mixed signals (i.e., $f_{G+N}(t)$) which quantifies the similarity of signals at different time points. We added the same target signal to the simulated noise signals\footnote{According to \cite{wang2015dsNoise, 2016InfraredTargetDetection, liuT2023infrared}, the combination of various noises in real scenes follows a Gaussian distribution, and can be modeled by Gaussian white noise.} with different standard deviations $\sigma_{N}$, as shown in Fig. \ref{fig:temporal_profile_analysis_corr}(a1). The target signal still has significance and the same up-down process when the noise is relatively weak (e.g., $\sigma_{N}=\frac{1}{4}$) in the time dimension. As the noise becomes stronger (e.g., $\sigma_{N}=\frac{1}{2}$), these features gradually become less pronounced. Then, we analyzed the correlation of $f_{G+N}(t)$ to find if there are still some distinctive characteristics. The correlation function $R_{G,N}(\tau)$ is calculated by measuring the inner product of $f_{G+N}(t)$ and $f_{G+N}(-t)$ under the time delay $\tau$, i.e.,
\begin{equation}
    \label{corGN}
    \begin{array}{rl}
        R_{G,N}(\tau) & = \int_{-\infty}^{+\infty} f_{G+N}(t) \cdot f_{G+N}(-t+\tau) dt \\
        & = f_{G+N}(\tau)*f_{G+N}(\tau).
    \end{array}
\end{equation}
After smoothing, the correlation function is shown in Fig.~\ref{fig:temporal_profile_analysis_corr}(a2). When there is no time delay (i.e., $\tau=0$), the correlation value is maximum since the target is autocorrelated in the time dimension. When $\tau\neq0$, the correlation value is close to zero. This demonstrates that the target signal is uncorrelated with the noise signal, which is also supported in \cite{2016InfraredTargetDetection}. Even for strong noise (e.g., $\sigma_N=\frac{1}{2}$), the target still maintains high autocorrelation characteristics and distinctiveness from the noise in the temporal profile. Therefore, \textit{the temporal profile holds the correlation information between target and noise, which is advantageous to predict the target signal even mixed with extremely strong noise}.

Then, to investigate the differences between target and clutter, we mixed target and different clutters (e.g., land clutter and moving cloud clutter), and analyzed the correlation function of the mixed signals (i.e., $f_{G+B}(t)$). At first, we got different clutters from the real video scenarios. After decentralizing and scaling, we added the fixed-intensity target signal to a series of clutter signals with different standard deviations $\sigma_{B}$\footnote{Note that the standard deviation of clutter in the time dimension is different from its fluctuation intensity in the space dimension. Usually, the temporal standard deviation of a clutter is smaller than its local spatial fluctuation since the background is slowly changing.}, as shown in Fig. \ref{fig:temporal_profile_analysis_corr}(b1) and (c1). The typical clutters usually include low-frequency signals (e.g., slowly-changing cloud in c1) and high-frequency signals (e.g., camera blind elements and flickering background in b1). High-frequency signals are more likely to be falsely detected as targets due to their spatial and temporal saliency. When the clutter signal in the time dimension is not strong, there are some obviously distinctive characteristics between targets and clutters. Usually, the fluctuations of the clutter are dense, and they are correlated with each other on the amplitude in a period but uncorrelated on the occurrence moment. However, these characteristics are still inadequate for the detection when the clutter has a high standard deviation. Following the above analysis, we calculated the correlation function $R_{G,B}(\tau)$ of $f_{G+B}(t)$, i.e.,
\begin{equation}
    \label{corGN}
    \begin{array}{rl}
        R_{G,B}(\tau) & = \int_{-\infty}^{+\infty} f_{G+B}(t) \cdot f_{G+B}(-t+\tau) dt \\
        & = f_{G+B}(\tau)*f_{G+B}(\tau).
    \end{array}
\end{equation}
The smoothed correlation function is shown in Fig. \ref{fig:temporal_profile_analysis_corr}(b2) and (c2). It can be observed that the target retains significant autocorrelation characteristics even if mixed with high-intensity land clutters or cloud clutters. Besides, the target signal is usually uncorrelated with the clutter signal as the correlation value at $\tau\neq0$ is close to zero. Therefore, \textit{the temporal profile contains a lot of correlation information between target and clutter, which is reliable to detect the target from even high-intensity clutters}.

According to the above analyses, target, noise, and clutter signals exhibit different characteristics in the temporal profile. Specifically, target signals are autocorrelated with a regular process and are irrelevant to noise and clutter signals. These distinct characteristics make the temporal profile more robust than other domains for IRST detection.

\subsection{Attribution verification}
From the theoretical analysis, we conclude that the temporal profile information is very important for IRST detection. As for the deep networks learning features from numerous labeled data, it is still unclear whether the temporal profile information is really essential for model predictions, whether it is more robust than the information in the spatial domain and the short-term spatial-temporal domain, and whether every moment in the temporal profile is important. In this section, we verify the importance of the temporal profile information in IRST detection networks.
\subsubsection{Attribution approach for IRST detection networks}
Before introducing the attribution approach in this work, we briefly summarize the preliminary \cite{sundararajan2017axiomatic} in classification networks. Let a function $F:\mathbb{R}^n\longrightarrow [0,1]$ and an input $\mathbf{x}=(x_1,...,x_n)\in\mathbb{R}^n$ denote a detection model and its input. Assume that $a_i$ is the contribution of $x_i$ to the prediction $F(\mathbf{x})$. Then, the attribution of the prediction at input $\mathbf{x}$ is defined as
\begin{equation}
    A_F(\mathbf{x},\mathbf{x'}) = (a_1,...,a_n)\in\mathbb{R}^n.
\end{equation}
The prediction attribution is relative to a baseline input $\mathbf{x'}$ which simulates the absence of important features and is used to provide a comparative reference. Consequently, the model prediction on a baseline input is usually zero (i.e., $F(\mathbf{x'})=0$). The necessity of a baseline has been proved in \cite{shrikumar2016not, binder2016layer}. Generally, the baseline input is an all-zero matrix or a matrix without information that can affect the prediction. Integrated gradients (IG) \cite{sundararajan2017axiomatic} is a typical attribution approach, and is formulated as
\begin{equation}
    IG(\mathbf{x}) = (\mathbf{x}-\mathbf{x'}) \int_0^1 \frac{\partial F(\mathbf{x'}+\alpha \times (\mathbf{x}-\mathbf{x'}))}{\partial \mathbf{x}} d\alpha.
\end{equation}
Here, $\frac{\partial F(\mathbf{x})}{\partial \mathbf{x}}$ is the gradient of $F(\mathbf{x})$ w.r.t. the input, which measures the local impact of input variations on the output within a narrow vicinity around $\mathbf{x}$.

Based on integrated gradients, we developed the first prediction attribution analysis tool for IRST detection networks. Unlike classification networks, segmentation-based IRST detection networks generate the prediction of each spatial pixel in the input $\mathbf{S}\in\mathbb{R}^{T\times H\times W}$, and can be represented as $D:\mathbb{R}^{T\times H\times W}\longrightarrow \mathbb{R}^{H\times W}$. $T$ denotes the temporal length of the input, while $H\times W$ denotes the input's spatial size. To interpret $D$, we attribute the prediction of a single target globally. Specifically, we need a converter to quantify the prediction of a target, $C: \mathbb{R}^{H\times W}\longrightarrow \mathbb{R}$. In this work, we achieved the converter by summing the predictive scores of all pixels in this target. That is,
\begin{equation}
    \label{attr}
    C(D(\mathbf{S}))=\sum\limits_{i=1}^{H} \sum\limits_{j=1}^{W} D(\mathbf{S})\odot M_k,
\end{equation}
where $M_k$ represents the mask of the $k^{th}$ target, and $\odot$ donates Hadamard product.

To generate the attribution map of $C(D(\mathbf{S}))$, we accumulated the gradients along a specified path from a baseline input $\mathbf{S'}$ to $\mathbf{S}$. The path function is represented as $\gamma(\alpha):[0,1]\longrightarrow \mathbb{R}^{T\times H\times W}$. Then, the attribution for IRST detection models is formulated as
\begin{equation}
    \label{attr}
    A_{D,C}^{\gamma}(\mathbf{S}, \mathbf{S'}) := \int_0^1 \frac{\partial C(D(\gamma(\alpha)))}{\partial \gamma(\alpha)}\times \frac{\partial \gamma(\alpha)}{\partial \alpha} d\alpha ,
\end{equation}
where $\mathbf{S'}\in\mathbb{R}^{T\times H\times W}$. Since the targets always present as high-frequency components in the input, the baseline should avoid containing high-frequency components. Meantime, arbitrary fluctuation in the background can affect the target prediction. Therefore, the constant tensor is suitable as a baseline for IRST detection, and the constant should be lower than the gray average of the target since integrated gradients fail to highlight black targets \cite{wang2024gradient}. The path used in this work is a straight line, i.e.,
\begin{equation}
    \gamma(\alpha) = \mathbf{S'}+\alpha(\mathbf{S}-\mathbf{S'}).
\end{equation}

With this attribution approach, the predictions of arbitrary segmentation-based IRST networks can be attributed.

\subsubsection{Attribution analysis} \label{attrRes_Analysis}
We analyzed the attribution results of a typical CNN-based MIRST detection network. We selected a widely used open source network, i.e., STDMANet \cite{yan2023stdmanet}, which takes 19 adjacent reference frames and a predicted frame as its input and detects the targets in the predicted frame.

\begin{figure}[!t]
    \centering
    \includegraphics[width=1\linewidth]{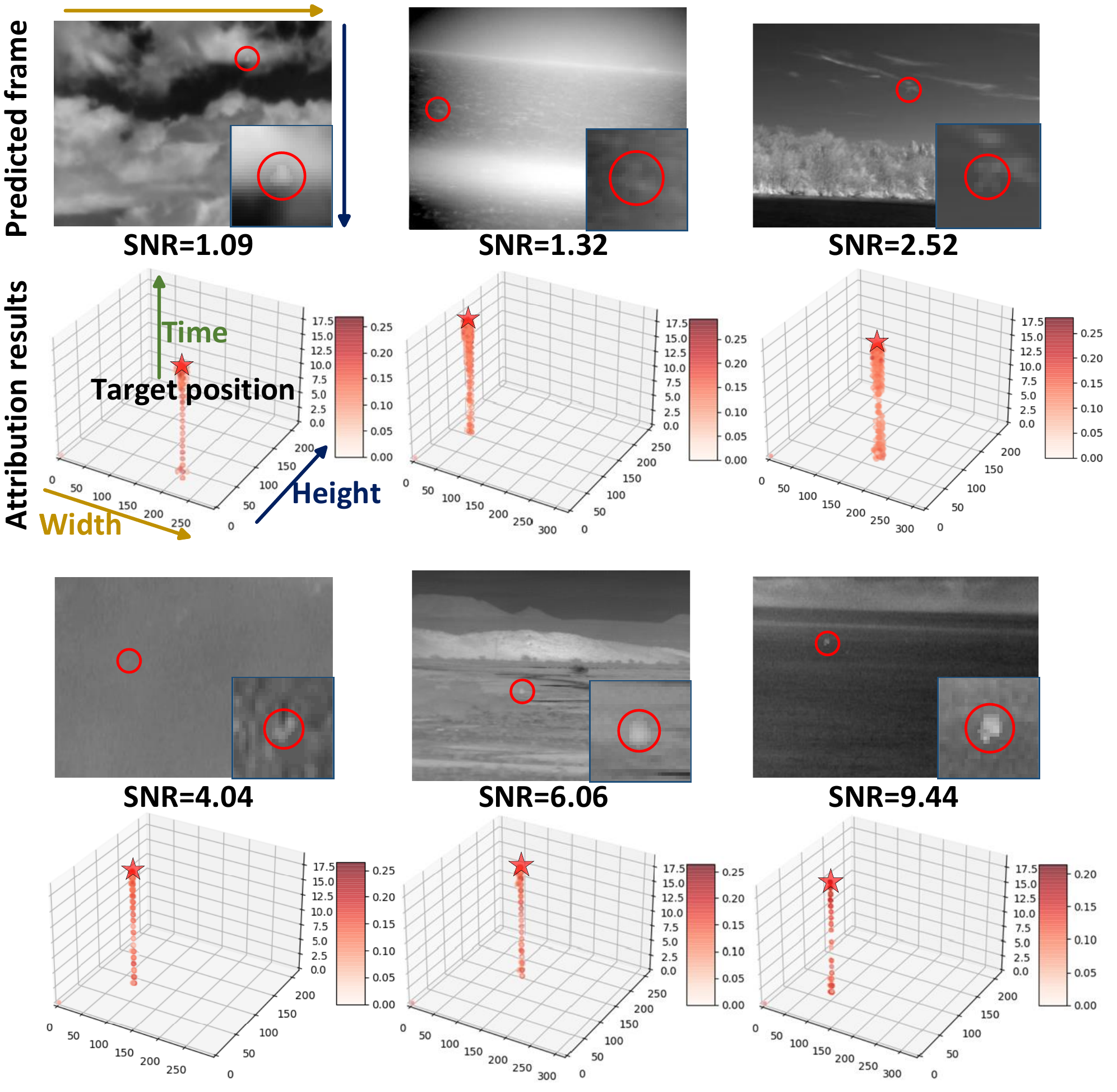}
    \caption{Attribution visualizations of reference frames at targets with different SNRs. For the predictions of the salient targets and the dim targets, the influential pixels are almost concentrated on the temporal profile of the target.}
	\label{attributionVis}
\end{figure}

We set up different seeds (e.g., 1202 \cite{yan2023stdmanet}, 3407 \cite{picard2021torch}, 42, 500, 1500, 2000, 2500, 3000) to train the network and observe the attribution results in different scenes. We first visualized the distribution of the input pixels in reference frames that significantly influence the prediction. The visualization is shown in Fig. \ref{attributionVis}, from which we observe Phenomenon 1 and verify that the temporal profile information is more important than the spatial information.

\vspace{0.2cm}\noindent\textbf{Phenomenon 1.} \textit{The influential pixels in reference frames are mainly distributed around the temporal profile of the target. The core influence region is close to a cylindrical zone along the time axis.}

As aforementioned, many IRST detection methods expand the spatial receptive field and supplementing the limited temporal continuous receptive field for more information. Following these thoughts, the influential pixels should be distributed over a wide spatial-temporal (especially spatial) field covering the target trajectory. However, Fig. \ref{attributionVis} shows that the influential pixels in reference frames are concentrated on the temporal profile of the target in the predicted frame. The influence region is small in the space dimension but covers all reference moments, and the region does not cover the target trajectory since the region is along the time axis. The results reveal that the thoughts in existing methods are one-sided under no matter simple scenes or extremely complex conditions, and \textit{the temporal profile information of the target is more important for its prediction}.

\begin{figure}[!t]
    \centering
    \includegraphics[width=1\linewidth]{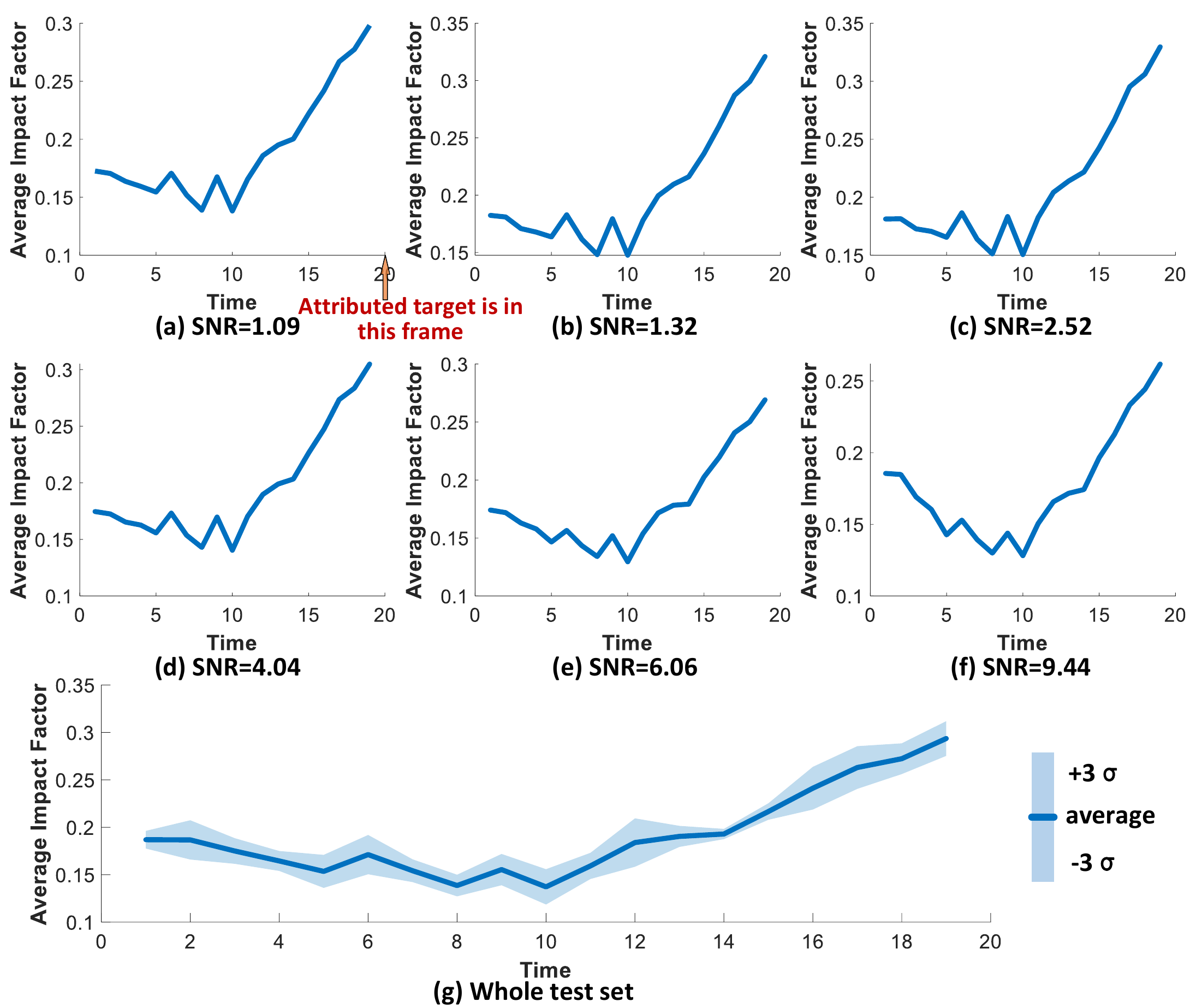}
    \caption{Temporal variations of the average influence and the influence range of all random seeds in different scenes. (a)-(f) show the temporal variations on the targets with SNR of 1.09, 1.32, 2.52, 4.04, 6.06, and 9.44, respectively. (g) illustrates the temporal variation in the whole test set.}
	\label{frameMAXAttri}
\end{figure}

Since the phenomenon demonstrates the significance of the temporal profile, a natural question arises: Is the influence of each reference frame on the predicted frame related to its temporal distance from the predicted moment? Therefore, we statistically assess the average influence (across a whole sequence) of different reference frames on the prediction, and the results are shown in Fig. \ref{frameMAXAttri}. From the results, we observe Phenomenon 2 and verify that long-term temporal information is important.

\vspace{0.2cm}\noindent\textbf{Phenomenon 2.} \textit{The importance of a reference frame changes over time, following a U-shaped curve. Distant information is crucial for predictions like recent information in a period.}

Researchers conventionally feed adjacent frames into networks to augment temporal context. There is a potential cognition that temporally proximate information is more important, and temporally distant information is not important. In this way, the curve should increase monotonically. However, there is something different from the above intuition in practical curves. That is, the importance degrees of 19 reference frames first decrease and then increase over time, exhibiting a U-shaped pattern. The difference reveals that a few adjacent frames are not enough, and distant information is also important for model prediction. In other words, \textit{the long-term variations of the signals in the temporal profile are crucial for IRST detection.}

\subsection{Temporal probe mechanism}  
\begin{figure*}[!t]
    \centering
    \includegraphics[width=1\linewidth]{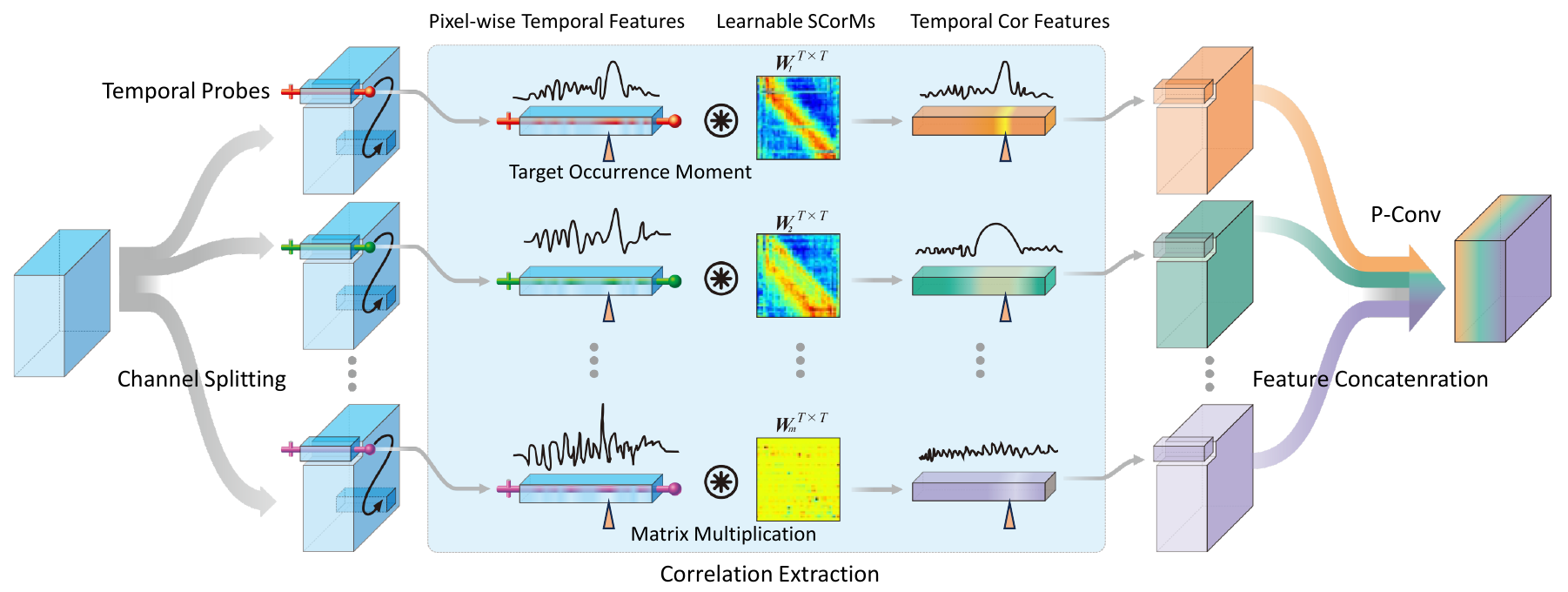}
    \caption{Visualization of the structure of TPro. Firstly, temporal probes are employed to pull out complete temporal features of a single pixel from the split input features. Then, multiple learnable SCorMs are applied to the gotten features to extract temporal correlation features. This process is performed pixel-wise across all split input features. Finally, all temporal correlation features are concatenated and fused to generate output features. Due to the learned correlationship among signals, target features are enhanced, and background features are suppressed. Notably, the entire pipeline strictly excludes any space-dimension computations.
    }
    \label{fig:architectureTPM}
\end{figure*}
Inspired by prior research, we remodel the IRST detection task as a one-dimensional (i.e., time-dimension) signal anomaly detection task, treating target signals as temporal anomalies. As previously analyzed, it is important for the networks to extract long-term temporal correlation information for this task. Ordinary 3D convolution operations focus only on short-term features and require stacking multiple convolution layers to achieve long temporal coverage. However, the temporal receptive field gradually expands as the number of stacked layers increases, and thus a amount of feature extraction computation is still based on short-term features. Moreover, multiple 3D convolutional layers bring substantial computational overhead for large-scale feature tensors. Therefore, in this part, we propose a temporal probe mechanism (TPro) to efficiently extract long-term temporal correlation features and to probe whether the temporal features of a pixel exist anomaly. The structure of TPro is shown in Fig. \ref{fig:architectureTPM}.

As the IRST detection task is remodeled as a one-dimensional (i.e., time-dimension) signal anomaly detection task, TPro is designed to focus on the long-term temporal features of an arbitrary pixel. The input is some basic features $F^b \in\mathbb{R}^{C\times T\times H\times W}$ of a series of input frames (with the temporal length of $T$ and the spatial size of $H\times W$), which are also the outputs of the models before TPro in a network. Its output $F^{tp} \in\mathbb{R}^{C\times T\times H\times W}$ is the features containing some temporal profile information. According to the analysis in Section \ref{Temporal-profile-analysis}, the target, background, and noise signals have different characteristics in the temporal correlation. Therefore, we used the learnable statistics-based correlation matrix (SCorM) $W^{T\times T} \in\mathbb{R}^{T\times T}$ to perceive temporal profile information and distinguish target signals from background signals. A matrix can learn a set of statistical properties that describe the correlation between signals at different moments. To fully learn these properties, it is essential to apply multiple SCorMs (i.e., $W_1^{T\times T}, W_2^{T\times T}, ..., W_m^{T\times T}, m \textgreater 1$). The operation details of TPro are described below.

As shown in Fig. \ref{fig:architectureTPM}, the input basic features are first divided into $m$ features along the channel dimension, i.e., $F^b_i \in\mathbb{R}^{\frac{C}{m}\times T\times H\times W}$ ($i=1,2,...,m$). To perform computations on the long-term temporal features of a pixel, temporal probes are utilized to individually extract the complete temporal features of each pixel, i.e., $F^b_i(x,y) \in\mathbb{R}^{\frac{C}{m}\times T\times 1\times 1}$ ($x\in [1,W], y\in [1,H]$). Subsequently, the correlation matrix is applied to the extracted features for the temporal correlation features $F^{cor}_i(x,y) \in\mathbb{R}^{\frac{C}{m}\times T\times 1\times 1}$. The process can be described as
\begin{equation}
    \label{equ_TP}
    F^{cor}_i(x,y) = F^{b}_i(x,y)\cdot W_{i}^{T\times T} ,
\end{equation}
where $W_{i}^{T\times T}$ represents the $i^{th}$ correlation matrix. After repeating the process, the temporal correlation features (i.e., $F^{cor}_i$) of all pixels are obtained. To fuse temporal correlation information from different channels and correlation matrixes, a pointwise convolution \cite{chollet2017xception} (P-Conv, i.e., $1\times 1\times 1$ convolution) layer $\phi(\cdot)$ is applied to these features (i.e., $F^{cor}_1, F^{cor}_2, ..., F^{cor}_m$) after concatenation along the channel dimension. This process can be described as
\begin{equation}
    \label{equ_Fusion}
    F^{tp} = \phi\big([F_{i,1}^{cor}, F_{i,2}^{cor}, ..., F_{i,m}^{cor}] \big),
\end{equation}
where $\phi(\cdot)$ includes convolution operation, batch normalization, and ReLU activation \cite{nair2010rectified}. Finally, the long-term temporal correlation features are generated with highlighting the anomaly signals.

\subsection{Deep temporal probe network for IRST detection}
\label{DeepPro}
\begin{figure}[!t]
    \centering
    \includegraphics[width=1\linewidth]{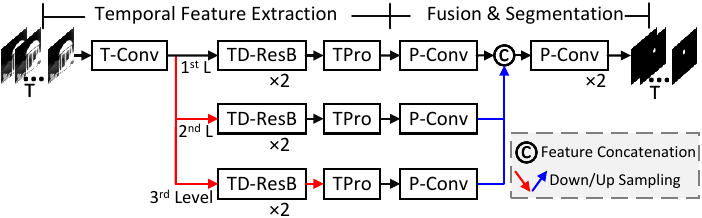}
    \caption{An overview of our DeepPro. There are three similar levels in our DeepPro. Note that this is the first IRST detection network with calculations (e.g., multiplications and additions) only in the time dimension.
    }
    \label{fig:architecture}
\end{figure}
The superior temporal profile information and the attribution verification inspire us to develop a new framework that emphasizes the temporal profile information. Therefore, based on TPro, we propose the deep temporal probe network called DeepPro. Our DeepPro consists of temporal feature extraction module (including TPro) and fusion \& segmentation module. There are three similar level structures operating in parallel, as shown in Fig. \ref{fig:architecture}. The input is a long sequence $\mathbf{S}\in\mathbb{R}^{T\times H\times W}$, and is first sent to the temporal feature extraction module for temporal features. After feature fusion and segmentation operations, i.e., a series of $1\times 1\times 1$ convolution layers, the final confidence maps of all input frames are generated, i.e., $\mathbf{C}\in\mathbb{R}^{T\times H\times W}$. There is no deep spatial-temporal feature extraction structure employed in most MIRST detection networks for a large spatial and short temporal receptive field, but only few temporal convolutions and TPro for long-term temporal information (i.e., temporal profile information). Our DeepPro achieves unprecedented simplicity in IRST detection, which is the first IRST detection network with calculations only in the time dimension. The details of the temporal feature extraction are described below.

Specifically, tens of frames $\mathbf{S}$ are first fed to a temporal convolution (T-Conv) layer\footnote{Each convolution layer consists of a convolution operation, batch normalization, and ReLU activation.} with a kernel $\mathbf{W_{T}}\in\mathbb{R}^{l\times 1\times 1}$ ($l=5$). Then, the features are sent to a few identical temporal difference residual blocks (TD-ResBlock) to roughly extract the basic temporal features $F^b$. Each TD-ResBlock contains a temporal difference convolution (TD-Conv) layer $\psi(\cdot)$ with a $3\times 1\times 1$ kernel $\mathbf{W_{TD}}$ and a temporal dilation of 2, a T-Conv layer ($l=3$), a P-Conv layer and the skip connection. TD-Conv serves to maintain and emphasize the temporal saliency of targets roughly. Inspired by the temporal-difference design in \cite{yu2021searching}, it accomplishes this by applying convolution to multiple temporal difference features (i.e., $[f_t-f_{t-1}, f_t-f_{t+1}]$), which can be described as
\begin{equation}
    \label{equ_TD}
    \begin{array}{rl}
    \psi(\cdot, \mathbf{W_{TD}}) & =\sum\limits_{t=1}^{l}w_t \cdot (2\cdot f_t - f_{t+1} - f_{t-1}) \\
     & = \sum\limits_{t=1}^{l}(2\cdot w_t - w_{t+1} - w_{t-1}) \cdot f_t,
    \end{array}
\end{equation}
where $f_t$ is the feature value of the $t^{th}$ ($t=1,2,3$) dimension in the time dimension within a sliding window. $w_t$ is the $t^{th}$ weight of the convolution kernel $\mathbf{W_{TD}}$.

To obtain the basic temporal features of the entire target region, the above process is performed multiple times at different levels. The only difference between these levels lies in the size of the features they process. The sizes are different due to the different number of max-pooling operations applied in a specific level. After multiple above processes, a series of basic temporal features are obtained, i.e., $F^b_j\in\mathbb{R}^{C\times T\times H\times W}$ ($j=1,2,3$). They are sent to TPro respectively for long-term temporal correlation feature extraction. Finally, various features containing temporal profile information are acquired, i.e., $F^{tp}_1, F^{tp}_2, F^{tp}_3$.

\section{Experiments}\label{experiments}
\subsection{Experimental setup}

\subsubsection{Datasets}
We evaluated our method on the NUDT-MIRSDT dataset \cite{li2023direction}, the IRDST dataset, \cite{sun2023receptive} and the real scenarios from the RGBT-Tiny dataset \cite{ying2025visible}. The IRDST-simulation dataset contains 102077 frames from 316 sequences. The NUDT-MIRSDT dataset contains 120 sequences, and each includes 100 frames. The test set of the NUDT-MIRSDT dataset is divided into two subsets according to their SNR. One consists of 8 sequences with SNR lower than 3, and the other consists of 12 sequences with SNR ranging from 3 to 10.

\begin{figure}[!t]
    \centering
    \includegraphics[width=1\linewidth]{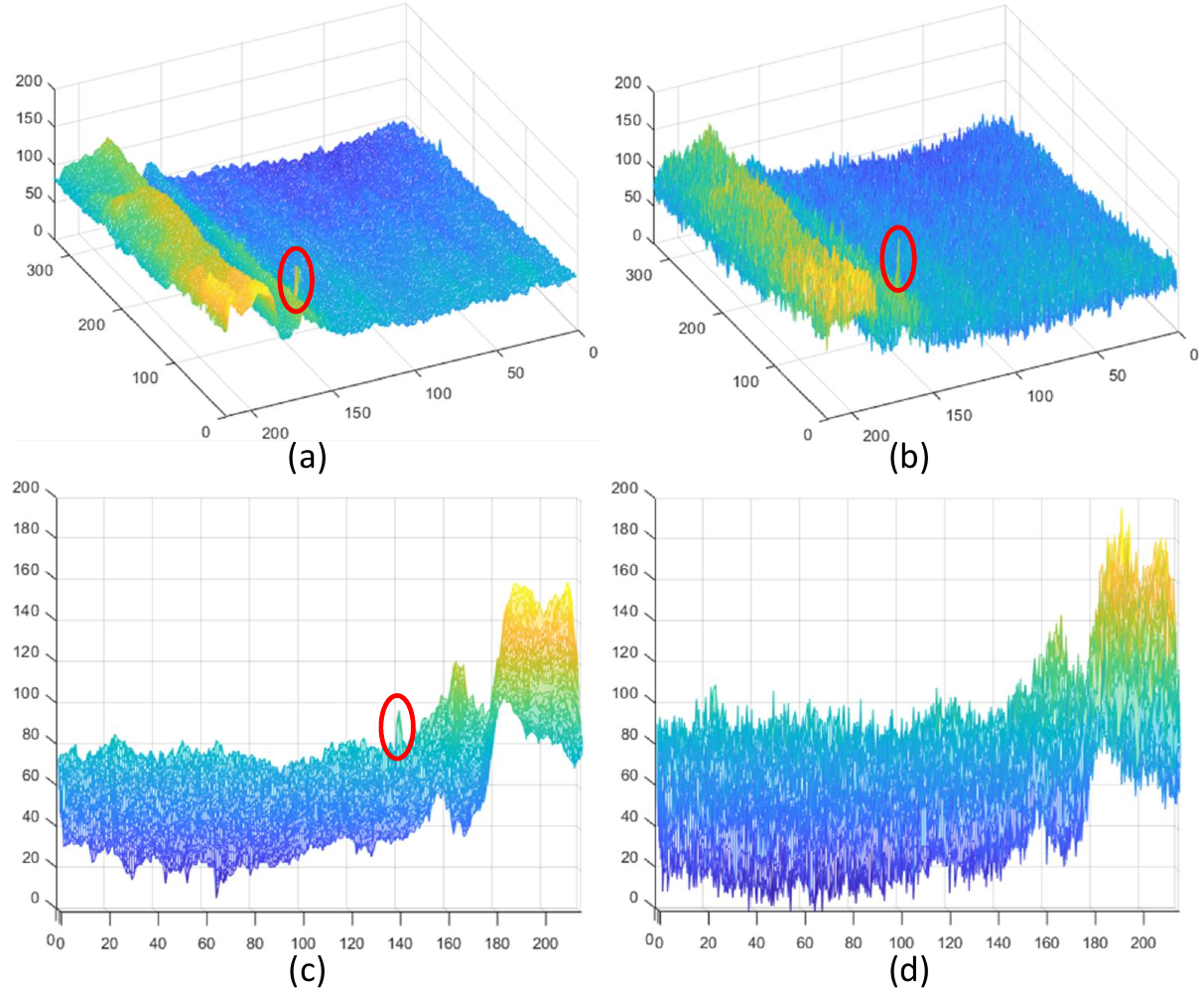}
    \caption{3D visualization of same targets in the NUDT-MIRSDT dataset and the NUDT-MIRSDT-HiNo dataset. (a) The bright target in the NUDT-MIRSDT dataset. (b) The same bright target in the NUDT-MIRSDT-HiNo dataset. (c) The dim target in the NUDT-MIRSDT dataset. (d) The same dim target in the NUDT-MIRSDT-HiNo dataset.
    }
    \label{fig:3dNoise}
\end{figure}

To further verify the robustness of our DeepPro to dim targets and complex scenarios, we built a dataset with strong noise by adding high-intensity noise to the NUDT-MIRSDT dataset, called NUDT-MIRSDT-HiNo dataset. The noise model is described in the Appendices. Detailed information on the NUDT-MIRSDT-HiNo dataset is reported in Table \ref{tab:NUDT-MIRSDT-HiNo}. Fig. \ref{fig:3dNoise} shows the effect of the strong noise on the targets. The noise makes the bright targets dim and makes the dim targets invisible in the space dimension.

\begin{table}[h]
\caption{Parameters related to noise of the datasets. The standard deviation of noise ($\sigma_n$), the distribution range of the multiplicative noise ($\sigma_g$), the standard deviation of the additive noise ($\sigma_o$), the pattern of non-uniformity, and the average SNR values of the whole dataset and the test set.}\label{tab:NUDT-MIRSDT-HiNo}
\centering
\setlength{\tabcolsep}{3.4pt}{
\begin{tabular}{c c c c c c c}
\hline
Datasets & $\sigma_n$ & $\sigma_g$ & $\sigma_o$ & non-uniformity & SNR & SNR (test) \\ \hline  
NUDT-MIRSDT      & 0.1 & 0 & 0 & none & 11.67 & 7.00 \\
NUDT-MIRSDT-HiNo & 8.0 & 0.15 & 1.3 & pixel-wise & 3.86 & 3.24 \\  \hline
\end{tabular}}
\end{table}

\subsubsection{Metrics}
We followed \cite{li2023direction} to use the probability of detection ($P_d$),  false alarm rate ($F_a$), and area under curve (AUC) as quantitative metrics for the evaluation of IRST detection performance. $P_d$ measures the ability of a method to detect targets. $F_a$ measures the ability of a method to correctly identify the background, such as clutter. AUC is the area value under the receiver operating characteristics (ROC) curve \cite{fawcett2006introduction,deng2017entropy}, reflecting the detection performance under different thresholds. Besides, the number of parameters (\#Params), GFLOPs, and FPS are measured to evaluate the scale, computational complexity, and computational efficiency of the model, respectively. Specifically, GFLOPs and FPS are calculated based on input images of size $256\times 256 $. In the following result tables, $P_d$ is expressed in units of $10^{-2}$, $F_a$ in $10^{-5}$, \#Params in millions (M), GFLOPs in GFLOPs per frame, and FPS in frames per second.

\subsubsection{Implementation details} \label{Implementation_details} 
All deep learning-based methods were implemented in PyTorch on a computer with an Intel Xeon Gold 6328H CPU @ 2.80GHz and two Tesla V100s PCIe 32GB GPUs. In detail, the networks were trained for 32 epochs with a batch size of 4. We used the Adam optimizer \cite{bearman2016s} with an initial learning rate of 0.001. The learning rate decayed after every 10 epochs with a decaying rate of 0.7. We initialized the weights of the convolution layers using the Kaiming method \cite{he2015delving}.

The input of our DeepPro was set to 40 consecutive frames ($T=40$) in a sequence, and the number of temporal probes was set to 4 ($m=4$). For data augmentation, we randomly selected one frame as the first frame in the input. If the number of subsequent frames is less than $T$, images with all zero values are filled. We adopted random cropping with a patch size of $128\times 128$, and ensured that the probability of the existence of targets in a sample is higher than 75\%. The Soft-IoU \cite{rahman2016optimizing} loss function was used for loss calculation. In the inference phase, we set the repetition rate at 10\%\footnote{For example, the first sample is continuous frames from the $1^{st}$ to the $T^{th}$, and the next sample is continuous frames from the $\big( (1-10\%)T+1 \big)^{th}$ to the $\big( (1-10\%)T+T \big)^{th}$.} to ensure that the targets at both ends of the input sequence can be well detected. When the remaining frames are less than $T$, overlapping windows are used to form complete frames. The final detection results for frames input multiple times through overlapping regions are combined using a union operation.

\begin{table*}[t!]
\caption{Detection results achieved by different state-of-the-art methods on the NUDT-MIRSDT dataset and the NUDT-MIRSDT-HiNo dataset. The best results are in bold, and the second-best results are underlined. \textit{SF} and \textit{MF} refer to single-frame and multi-frame methods, respectively.}\label{tab:SOTA1}
\centering
\renewcommand{\arraystretch}{1.1}
\begin{tabular}{c c l c c c c c c c c r c}
\hline
\multicolumn{3}{c}{\multirow{2}*{Methods}} & \multicolumn{2}{c}{$SNR\leq 3$} & \multicolumn{3}{c}{NUDT-MIRSDT} & \multicolumn{3}{c}{NUDT-MIRSDT-HiNo} & \multirow{2}*{\makecell{\#Params\\(M)}} &\multirow{2}*{FPS}\\
\cline{4-11}
\multicolumn{3}{c}{~}                    & $P_d$  & $F_a$        & $P_d$  & $F_a$ & AUC       & $P_d$  & $F_a$  & AUC & & \\ \hline
\multirow{8}*{\rotatebox{90}{Traditional Methods}} & \multirow{8}{*}{\rotatebox{90}{\textit{MF}}}  & MSLSTIPT \cite{sun2020infrared} \textit{(TGRS'20)} & 4.16 & 21.70 & 18.97 & 15.37 & 0.9404 & 3.93 & 73.76 & 0.9185 & - & 0.17 \\
                                                  & & IMNN-LWEC \cite{luo2022imnn} \textit{(TGRS'22)} & 0.00 & 7.22 & 26.43 & 10.74 & 0.6734 & 4.97 & 83.28 & 0.5394 & - & 30.31 \\
                                                  & & SRSTT \cite{li2023sparse} \textit{(TGRS'23)} & 69.94 & 6.12 & 90.63 & 3.35 & \textbf{0.9989} & 4.34 & 55.04 & 0.5358 & - & 0.06 \\  
                                                  & & 4DST-BTMD \cite{luo20234dst} \textit{(TGRS'23)} & 41.58 & 23.45 & 44.77 & 74.95 & 0.8488 & 4.80 & 77.29 & 0.6651 & - & 26.09 \\  
                                                  & & STRL-LBCM \cite{luo2023spatial} \textit{(TAES'23)} & 5.48 & 85.53 & 19.03 & 34.05 & 0.5972 & 2.55 & 77.78 & 0.5238 & - & 0.87 \\  
                                                  & & 4D-TR \cite{wu2023infrared} \textit{(TGRS'23)} & 55.77 & 2.55 & 55.70 & 3.19 & 0.9946 & 4.63 & 120.16 & 0.6633 & - & 0.36 \\  
                                                  & & 4D-TT \cite{wu2023infrared} \textit{(TGRS'23)} & 24.95 & 1.67 & 30.89 & 3.21 & 0.8287 & 6.94 & 73.18 & 0.5347 & - & 10.82 \\  
                                                  & & NFTDGSTV \cite{liuT2023infrared} \textit{(TGRS'23)} & 1.51 & 32.31 & 13.77 & 35.32 & 0.8613 & 11.56 & 43.16 & \underline{0.9524} & - & 0.58 \\ \hline  
\multirow{16}*{\rotatebox{90}{Deep-Learning Methods}}  & \multirow{11}{*}{\rotatebox{90}{\textit{SF}}}  & ACM \cite{dai2021asymmetric} \textit{(WACV'21)} & 7.75 & 22.88 & 51.53 & 17.52 & 0.9298 & 0 & - & - & \underline{1.592} & 171.31 \\
                                                      & & ALCNet \cite{dai2021attentional} \textit{(TGRS'21)} & 3.97 & 37.10 & 52.57 & 25.50 & 0.8435 & 36.09 & 91.99 & 0.9326 & 3.457 & 142.51 \\
                                                      & & Res-UNet \cite{xiao2018weighted} \textit{(ITME'18)} & 15.83 & 30.32 & 63.27 & 40.83 & 0.9198 & 35.51 & 22.55 & 0.9391 & 3.656 & \textbf{222.85} \\
                                                      & & DNA-Net \cite{li2022dense} \textit{(TIP'22)} & 23.74 & 19.23 & 67.38 & 15.07 & 0.8843 & 49.16 & 60.98 & 0.9373 & 18.787 & 29.87 \\
                                                      & & ISNet \cite{zhang2022isnet} \textit{(CVPR'22)} & 17.96 & 8.53 & 65.99 & 19.25 & 0.9123 & 28.40 & 90.17 & 0.9224 & 4.363 & 52.25 \\
                                                      & & UIUNet \cite{wu2022uiu} \textit{(TIP'22)} & 15.12 & 17.46 & 61.25 & 14.42 & 0.9436 & 43.67 & 28.87 & 0.9246 & 202.152 & 12.46 \\
                                                      & & AGPCNet \cite{zhang2023attention} \textit{(TAES'23)} & 31.76 & 176.38 & 55.47 & 85.56 & 0.9443 & 42.452 & 13655.20 & 0.7348 & 49.442 & 24.13 \\
                                                      & & MSHNet \cite{liu2024infraredMSHNet} \textit{(CVPR'24)} & 2.46 & 78.07 & 36.78 & 41.91 & 0.7966 & 21.81 & 44.92 & 0.6748 & 16.262 & 69.35 \\
                                                      & & SCTransNet \cite{yuan2024sctransnet} \textit{(TGRS'24)} & 23.63 & 122.24 & 62.81 & 74.20 & 0.9320 & 20.65 & 35.37 & 0.8283 & 45.304 & 20.07 \\
                                                      & & RPCANet \cite{wu2024rpcanet} \textit{(WACV'24)} & 30.06 & 81.21 & 61.13 & 41.28 & 0.8694 & 21.81 & 198.14 & 0.8786 & 2.720 & 31.47 \\ 
                                                      & & ILNet \cite{li2025ilnet} \textit{(TAES'25)} & 17.39 & 55.89 & 57.90 & 34.09 & 0.7572 & 34.11 & 56.53 & 0.6551 & 6.322 & 36.82 \\ \cline{2-13}
                                                      & \multirow{5}{*}{\rotatebox{90}{\textit{MF}}}
                                                      & Res-U+DTUM \cite{li2023direction} \textit{(TNNLS'23)} & 91.68 & 2.37 & \underline{97.46} & 3.00 & 0.9967 & 43.90 & 4.86 & 0.9413 & 1.193 & 62.65 \\
                                                      & & STDMANet \cite{yan2023stdmanet} \textit{(TGRS'23)} & \underline{92.82} & 2.88 & 96.59 & 3.40 & 0.9908 & 51.65 & \underline{1.95} & 0.8766 & 47.518 & 18.66 \\ 
                                                      & & Res-U+RFR \cite{ying2025infrared} \textit{(TGRS'25)} & 79.39 & 4.15 & 93.35 & 1.95 & 0.9657 & \underline{55.64} & 111.23 & 0.7839 & 4.096 & 102.88 \\ 
                                                      & & DQAligner \cite{deng2026learning} \textit{(TGRS'26)} & 81.29 & \textbf{0.37} & 94.22 & \textbf{0.15} & 0.9419 & 33.41 & 2.95 & 0.8022 & 2.410 & 10.01 \\
                                                      & & DeepPro \textit{(ours)} & \textbf{95.84} & \underline{0.52} & \textbf{98.50} & \underline{0.72} & \underline{0.9973} & \textbf{59.17} & \textbf{1.76} & \textbf{0.9638} & \textbf{0.197} & \underline{184.55} \\ \hline
\end{tabular}
\end{table*}

\begin{table}[t!]
\caption{Detection results achieved by different state-of-the-art methods on the IRDST-simulation dataset. The best results are in bold, and the second-best results are underlined.}\label{tab:SOTA2}
\centering
\renewcommand{\arraystretch}{1.1}
\begin{tabular}{c l c c c c}
\hline
\multicolumn{2}{c}{Methods} & $P_d$  & $F_a$ & AUC & GFLOPs \\ \hline  
\multirow{10}{*}{\rotatebox{90}{\textit{SF}}}
                                              & ALCNet \cite{dai2021attentional} & 75.06 & 1.25 & 0.9587 & \textbf{0.45} \\  
                                              & Res-UNet \cite{xiao2018weighted} & 82.10 & 1.34 & 0.9729 & 1.30 \\  
                                              & DNA-Net \cite{li2022dense} & 81.44 & 1.12 & 0.9724 & 14.28 \\  
                                              & ISNet \cite{zhang2022isnet} & 82.02 & 1.06 & 0.9630 & 30.63 \\  
                                              & UIUNet \cite{wu2022uiu} & 87.58 & 1.69 & 0.9729 & 54.50 \\  
                                              & AGPCNet \cite{zhang2023attention} & 59.48 & 94.15 & 0.9529 & 43.18 \\
                                              & MSHNet \cite{liu2024infraredMSHNet} & 78.01 & 1.10 & 1.8413 & 6.11 \\
                                              & SCTransNet \cite{yuan2024sctransnet} & 83.67 & 6.34 & 0.9564 & 10.12 \\
                                              & RPCANet \cite{wu2024rpcanet} & 88.06 & 281.96 & 0.9686 & 44.57 \\  
                                              & ILNet \cite{li2025ilnet} & 41.55 & 0.33 & 0.6964 & 20.97 \\ \hline
\multirow{5}{*}{\rotatebox{90}{\textit{MF}}}  & Res-U+DTUM \cite{li2023direction} & \underline{99.48} & \textbf{0} & 0.9730 & 6.28 \\  
                                              & STDMANet \cite{yan2023stdmanet} & 98.88 & \textbf{0} & \underline{0.9978} & 95.54 \\  
                                              & Res-U+RFR \cite{ying2025infrared} & 98.90 & 9.36 & 0.9950 & 9.43 \\  
                                              & DQAligner \cite{deng2026learning} & 96.86 & \textbf{0} & 0.9742 & 27.56 \\
                                              & DeepPro & \textbf{99.85} & \underline{0.13} & \textbf{0.9993} & \underline{1.01} \\ \hline
\end{tabular}
\end{table}

\subsection{Comparison to the state-of-the-arts}\label{ComparisonSORT}
\begin{figure*}[t]
    \centering
    \includegraphics[width=1\linewidth]{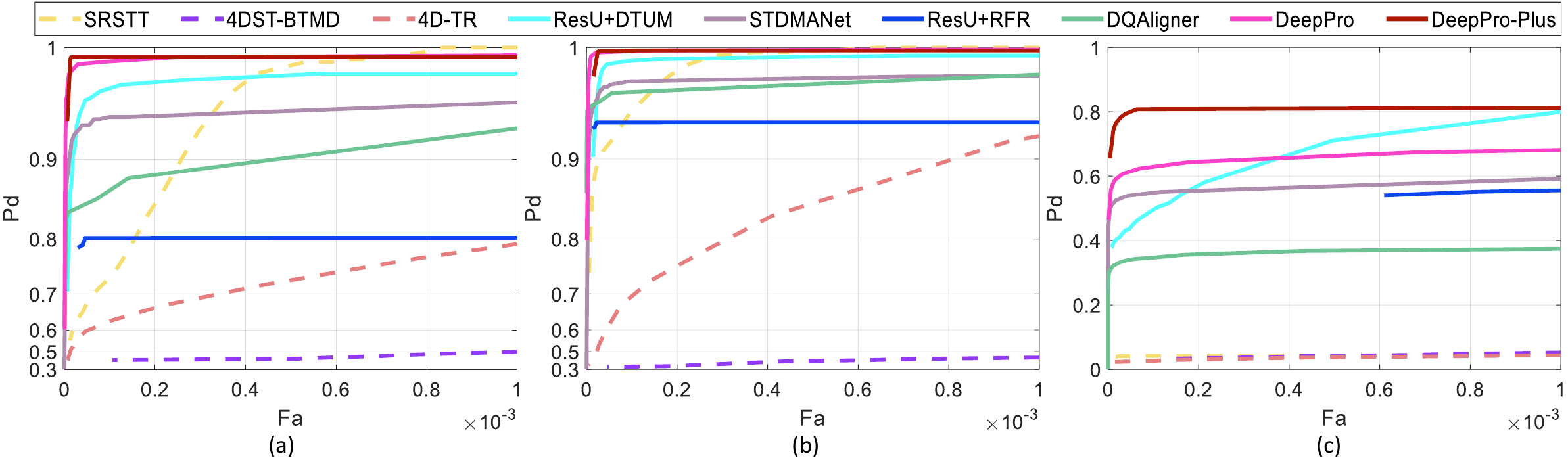}
    \vspace{-6mm}
    \caption{ROC performances of different methods on the NUDT-MIRSDT dataset and the NUDT-MIRSDT-HiNo dataset. (a) NUDT-MIRSDT ($SNR\leq 3$). (b) NUDT-MIRSDT. (c) NUDT-MIRSDT-HiNo.
    }
    \label{fig:ROC}
\end{figure*}

\begin{figure*}[t]
    \centering
    \includegraphics[width=1\linewidth]{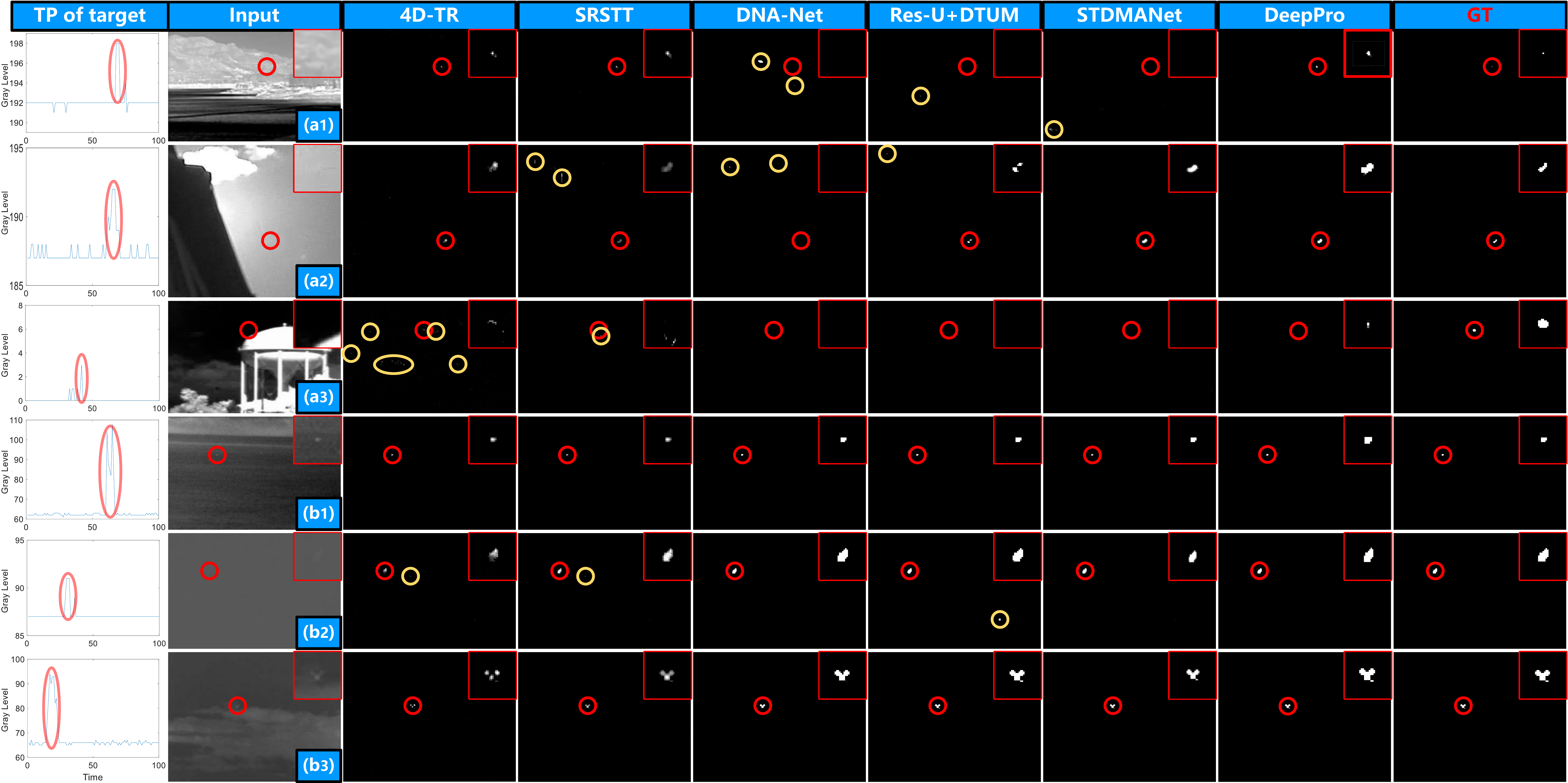}
    \caption{Visual comparison on the NUDT-MIRSDT dataset. (a1-a3) Results on the NUDT-MIRSDT ($SNR\leq 3$) subset. (b1-b3) Results on the NUDT-MIRSDT ($SNR> 3$) subset. The first column shows the temporal profile (TP) of the target central pixel. For better visualization, the target area is enlarged in the top-right corner and highlighted with a red circle. The false alarm area is marked with a yellow circle.
    }
    \label{fig:visual}
\end{figure*}

To demonstrate the effectiveness of our method, we compared our method with state-of-the-art IRST detection methods, including traditional MIRST detection methods (MSLSTIPT \cite{sun2020infrared}, IMNN-LWEC \cite{luo2022imnn}, SRSTT \cite{li2023sparse}, 4DST-BTMD \cite{luo20234dst}, STRL-LBCM \cite{luo2023spatial}, 4D-TR \cite{wu2023infrared}, 4D-TT \cite{wu2023infrared}, and NFTDGSTV \cite{liuT2023infrared}), deep learning-based SIRST detection methods (ACM \cite{dai2021asymmetric}, ALCNet \cite{dai2021attentional}, Res-UNet \cite{xiao2018weighted}, DNA-Net \cite{li2022dense}, ISNet \cite{zhang2022isnet}, UIUNet \cite{wu2022uiu}, AGPCNet \cite{zhang2023attention}, MSHNet \cite{liu2024infraredMSHNet}, SCTransNet \cite{yuan2024sctransnet}, RPCANet \cite{wu2024rpcanet}, and ILNet \cite{li2025ilnet}), and deep learning-based MIRST detection methods (Res-UNet+DTUM \cite{li2023direction}, STDMANet \cite{yan2023stdmanet}, Res-UNet+RFR \cite{ying2025infrared}, and DQAligner \cite{deng2026learning}) on the NUDT-MIRSDT dataset, the NUDT-MIRSDT-HiNo dataset, and the IRDST-simulation dataset. All traditional methods were implemented with their default parameters. We adopted 0 as the threshold for all deep learning-based methods, i.e., 0.5 after sigmoid operation \cite{ying2023mapping, li2023direction}. Quantitative results are presented in Tables \ref{tab:SOTA1}, \ref{tab:SOTA2} and Fig. \ref{fig:ROC}, and qualitative results are shown in Fig. \ref{fig:visual}.

\subsubsection{Quantitative results}

As shown in Table \ref{tab:SOTA1} and Fig. \ref{fig:ROC}, our DeepPro outperforms the state-of-the-art methods in all scenarios. Especially, on the low-SNR subset and the NUDT-MIRSDT-HiNo dataset, DeepPro on average improves $P_d$ by $3.59\%$ and $7.52\%$, while reducing $F_a$ by $2.11\times 10^{-5}$ and $0.19\times 10^{-5}$, respectively, over sub-optimal methods (i.e., DTUM \cite{li2023direction}, STDMANet \cite{yan2023stdmanet}). The results verify that the temporal profile has significant advantages in containing more distinctive characteristics for IRST detection. That is, long-term temporal information is worthy of attention. Besides, among the deep learning-based methods including the single-frame ones, our DeepPro has the fewest parameters and achieves the second-highest inference efficiency and the second-lowest GFLOPs per frame. That is because, DeepPro performs calculations (e.g., addition and multiplication) only in the time dimension for detection, which still benefits from the superiority of the temporal profile.

From the results of the state-of-the-art methods, we can also observe the advantages of temporal information. In most scenarios (e.g., the NUDT-MIRSDT dataset and the NUDT-MIRSDT-HiNo dataset), deep learning-based SIRST detection methods perform better than traditional MIRST detection methods. However, on the low-SNR subset, some traditional methods (e.g., SRSTT, 4D-TR, 4D-TT, and 4DST-BTMD) significantly outperform deep learning-based SIRST detection methods. That is because, it is hard to detect dim targets (e.g., $SNR\leq 3$) without temporal information. Moreover, SRSTT \cite{li2023sparse} in traditional methods and STDMANet \cite{yan2023stdmanet} in deep learning-based methods achieve much better detection performance than other existing methods on dim targets and complex scenarios (e.g., the NUDT-MIRSDT-HiNo dataset), because they can extract longer temporal features from more input frames (i.e., 30 and 20, respectively) in a sequence.


\subsubsection{Qualitative results}

We compared the qualitative results of different detection methods on the NUDT-MIRSDT dataset, as shown in Fig. \ref{fig:visual}. Compared with the state-of-the-art methods, DeepPro generates better visual results, especially on dim targets (e.g., $SNR\leq 3$) nearly invisible spatially, due to the enhanced signal distinguishability in the temporal profile. In complex scenarios (e.g., a1, a2, and a3), traditional methods and deep learning-based SIRST methods have significant disadvantages in suppressing false alarms, while most deep learning-based MIRST detection methods (e.g., STDMANet, and our DeepPro) with long sequence as input perform well. That shows the importance of long temporal information for IRST detection.

\subsection{Model analyses}\label{Model_analysis} 
\subsubsection{Effectiveness of TPro}

\begin{table}[!t]
    \centering
    \caption{Detection results achieved by variants of DeepPro with and without TPro on the NUDT-MIRSDT dataset and the NUDT-MIRSDT-HiNo dataset.}
    \renewcommand{\arraystretch}{1.1}
    \begin{tabular}{c c c c c}
    \hline
    Datasets & Models & $P_d$ & $F_a$ & AUC \\ \hline
    \multirow{2}{*}{NUDT-MIRSDT} & DeepPro w/o TPro & 97.34 & 1.96 & 0.9954 \\
                          & DeepPro & 98.50 & 0.72 & 0.9973 \\ \hline
    \multirow{2}{*}{$SNR\leq 3$} & DeepPro w/o TPro & 92.44 & 3.39 & 0.9902 \\
                          & DeepPro & 95.84 & 0.52 & 0.9952 \\ \hline
    \multirow{2}{*}{NUDT-MIRSDT-HiNo} & DeepPro w/o TPro & 53.21 & 1.75 & 0.9137 \\
                          & DeepPro & 59.17 & 1.76 & 0.9638 \\ \hline
    \end{tabular}
    \label{tab:woTPM}
\end{table}

To verify the effectiveness of TPro in IRST detection, we compared the quantitative results of the variant without TPro on the NUDT-MIRSDT dataset and the NUDT-MIRSDT-HiNo dataset. Specifically, we replaced TPro with the same group number (i.e., $m=4$) of $1\times 1\times T$ convolutions. The results are shown in Table \ref{tab:woTPM}. It can be observed that DeepPro without TPro suffers 3.4\% and 5.96\% reductions in terms of $P_d$ on the $SNR\leq 3$ subset and the NUDT-MIRSDT-HiNo dataset, respectively, and suffers a $2.87\times10^{-5}$ increase in terms of $F_a$ on the NUDT-MIRSDT dataset. This demonstrates that TPro plays an important role in extracting sufficient distinctive characteristics from the temporal profile.

\begin{figure}[!t]
    \centering
    \includegraphics[width=1\linewidth]{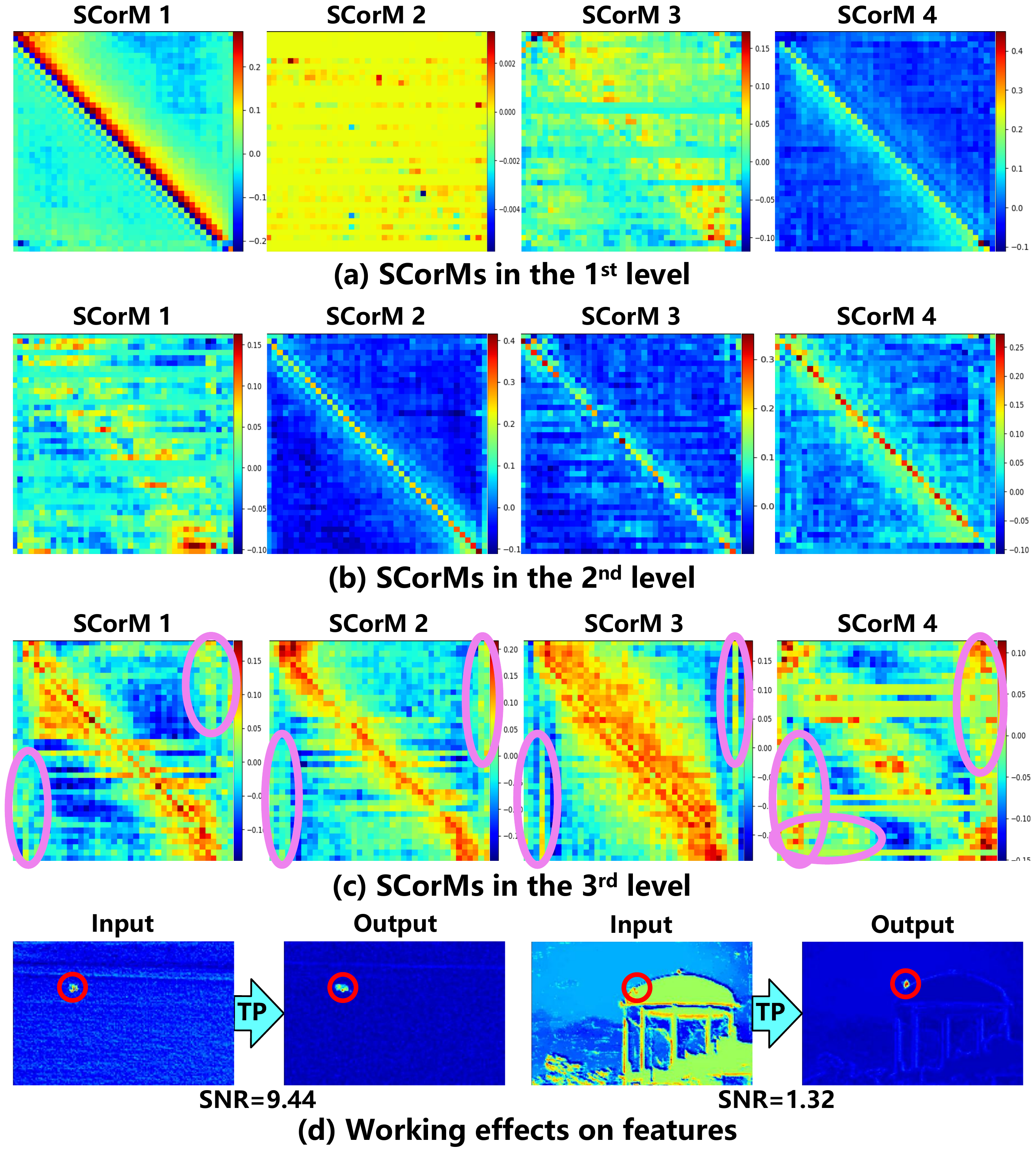}
    \caption{Visualization of statistic-based correlation matrixes in our DeepPro. (a) SCorMs in the first level which focus on all moments and the target occurrence moments (i.e., the diagonal). (b) SCorMs in the second level which focus on the target occurrence moments. (c) SCorMs in the third level which focus on the moments both proximal and distal to the target presence. The violet circles represent significant moments other than the moments near the target presence. (d) Working effects of TPro on features. For better visualization, the target area is highlighted by a red circle.
    }
    \label{fig:visual_TPM}
\end{figure}
To further explore how TPro works, we visualized the statistics-based correlation matrixes (SCorM) and their input and output features in Fig. \ref{fig:visual_TPM}. Different SCorMs contain different statistical properties of the signals. They focus on the intensity variations of signals in different periods and mine different temporal profile features. Fig. \ref{fig:visual_TPM}(a) shows that, in the first level, some SCorMs pay attention to all moments of the input sequence (e.g., SCorMs 2 and 3), and others pay attention to the moments close to the target existence (e.g., SCorMs 1 and 4). In the second level, most SCorMs focus on the target occurrence moment, as shown in Fig. \ref{fig:visual_TPM}(b). In the third level shown in Fig. \ref{fig:visual_TPM}(c), there are some purple circles marking the moments with the locally highest weights except the moments when the targets occur. That is, SCorMs in this level focus on the moments near the target presence and the farthest moments before and after the target presence. That also confirms \textbf{Phenomenon 2} in Section \ref{attrRes_Analysis}, i.e., \textit{distant information is crucial for predictions like recent information}. Different SCorMs work together to extract various temporal profile information and highlight the target area. Especially, the dim target is significantly enhanced after TPro or temporal profile calculations as shown in Fig. \ref{fig:visual_TPM}(d).

\subsubsection{Length of temporal profile information} 

\begin{table}[t]
\centering
    \caption{Detection results achieved by our DeepPro with different lengths of input frames on the NUDT-MIRSDT-HiNo dataset.}
    \label{tab:lenTP}
    \renewcommand{\arraystretch}{1.1}
    \begin{tabular}{c c c c c c c}
    \hline
    $T$ & $P_d$ & $F_a$ & AUC & \#Params (M) & GFLOPs & FPS \\
    \hline
    5 & 23.14 & 2.02 & 0.8339 & 0.119 & 0.92 & 165.53 \\
    10 & 46.44 & 7.18 & 0.9634 & 0.123 & 0.93 & 171.99 \\
    20 & 57.49 & 1.72 & 0.9726 & 0.138 & 0.96 & 177.94 \\
    40 & 59.17 & 1.76 & 0.9638 & 0.197 & 1.01 & 184.55 \\
    60 & 59.63 & 2.68 & 0.9797 & 0.293 & 1.07 & 190.52 \\
    80 & 60.21 & 3.61 & 0.9804 & 0.429 & 1.12 & 183.47 \\
    \hline
    \end{tabular}
\end{table}

To explore the impact of the length of the temporal profile information used for detection, we changed the number $T$ of input frames to 5, 10, 20, 60, and 80, as well as the size of SCorM (i.e., $T\times T$). The results are summarized in Table \ref{tab:lenTP}. It can be observed that with the extension of temporal information, $P_d$ and $F_a$ are significantly improved, and the inference efficiency (i.e., FPS) is also gained, albeit with a slight rise in GFLOPs. However, when the length of the time dimension exceeds 40 (e.g., $T=60, 80$), there is no significant improvement in the detection performance, while the number of parameters and GFLOPs continue to rise gradually. Furthermore, when $T=5$, DeepPro exhibits significantly degraded detection performance due to the insufficient temporal profile information. This underscores that effective temporal probing relies on long-term temporal information.

\begin{figure*}[!t]
    \centering
    \includegraphics[width=1\linewidth]{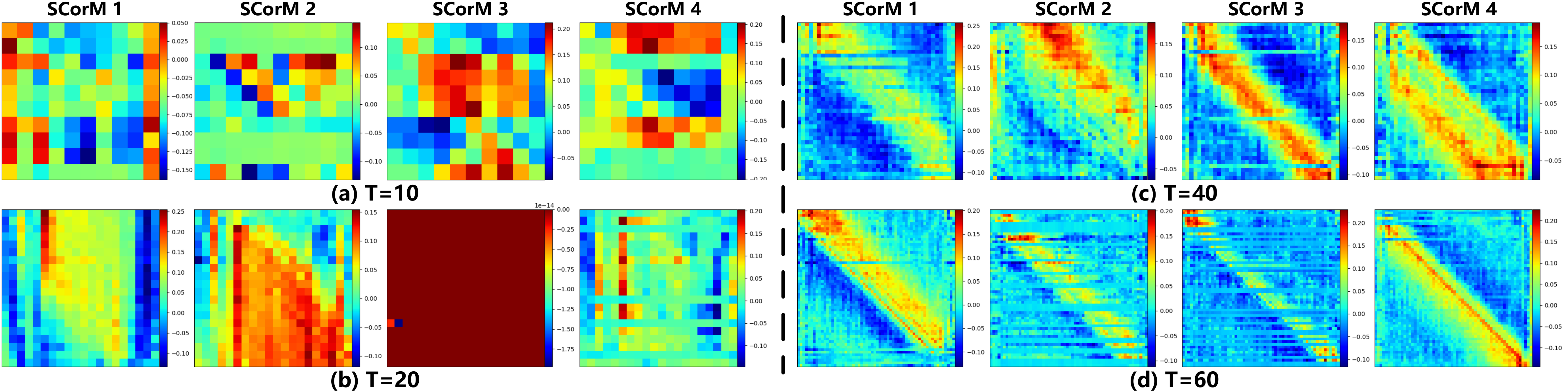}
    \caption{Visualization of the statistic-based correlation matrixes in the third level with different lengths. (a) SCorMs with length 10 and (b) SCorMs with length 20 are irregular. (c) SCorMs with length 40 and (b) SCorMs with length 60 are approximately symmetric about the minor diagonals. (SCorMs with length 20 are also approximately symmetric about the minor diagonals in the second level as shown in the Appendices. This explains why DeepPro with $T=20$ demonstrates acceptable performance as compared to the network with $T=40$, and even performs much better than STDMANet.)
    }
    \label{fig:visual_TP_L}
\end{figure*}

To understand why utilizing short temporal profile information performs worse, we visualized the SCorMs in the third level with different lengths in Fig. \ref{fig:visual_TP_L}. In predicting the signal property at a given moment, the importance of signals at other moments is only related to their relative time distances rather than their absolute times. Consequently, the SCorMs should be symmetric about the minor diagonals. When the length is shorter than 40, the matrices deviate from this expected symmetry in the third level, e.g., $T=10$ and $T=20$ in Fig. \ref{fig:visual_TP_L}. This illustrates that it is difficult to learn consistent essential features or accurate statistical properties from short temporal information.

Notably, we compared the results of STDMANet \cite{yan2023stdmanet} with the input of length 20 in Table \ref{tab:SOTA1} and the results of the variant with $T=20$ in Table \ref{tab:lenTP}. It can be observed that the detection performance of our DeepPro is obviously superior in all terms of $P_d$, $F_a$, AUC, \#Params, GFLOPs, and FPS. This demonstrates that \textit{the temporal profile information is more important than local spatial-temporal information in IRST detection.}

\begin{table}[t]
    \caption{Detection results achieved by our DeepPro with different numbers of SCorMs on the NUDT-MIRSDT-HiNo dataset. The best results are in bold.} \label{tab:numTP}
    \centering
    \renewcommand{\arraystretch}{1.1}
    \begin{tabular}{c c c c c}
    \hline
    $m$ & $P_d$ & $F_a$ & \#Params (M) & FPS \\
    \hline
    0 & - & - & 0.117 & - \\
    1 & 55.12 & 2.44 & \textbf{0.137} & 180.76 \\
    2 & 59.11 & 3.12 & 0.157 & 180.40 \\
    4 & \textbf{59.17} & \textbf{1.76} & 0.197 & 184.55 \\
    8 & \textbf{59.17} & 3.27 & 0.275 & 184.21 \\
    \hline
    \end{tabular}
\end{table}

\subsubsection{Analysis of multiple SCorMs}
To investigate the effect of multiple SCorMs on learning variation characteristics in the temporal profile, we changed the number $m$ to 1, 2, and 8. Note that different number SCorMs have identical GFLOPs. We compared the detection performances of these variants on the NUDT-MIRSDT-HiNo dataset. The results are reported in Table \ref{tab:numTP}. Compared with a single SCorM, multiple SCorMs significantly improve $P_d$. In particular, the variant with four SCorMs achieves the best detection performance in terms of $P_d$ and $F_a$. That is because multiple SCorMs can learn rich correlation properties, which is beneficial for mining temporal profile information, like multi-head attention \cite{vaswani2017attention}. However, more SCorMs are not cost-effective, since there is no obvious performance improvement but increases in the number of parameters. In addition, the parameter amount of one SCorM is only 6.56K, and that of other structures in DeepPro is only 0.117M. Our DeepPro is extremely lightweight and effective.

\subsubsection{Analysis of the level structure}

\begin{table}[t]
    \renewcommand\arraystretch{1.16}
    \caption{Detection results achieved by several variants of DeepPro with different levels on the NUDT-MIRSDT-HiNo dataset. The best results are in bold.} \label{tab:level}
    \centering
    \renewcommand{\arraystretch}{1.1}
    \begin{tabular}{c c c c}
    \hline
    levels in DeepPro & $P_d$ & $F_{a}$ & AUC \\
    \hline
    $1^{st}$ & 53.50 & 13.82 & 0.9355 \\
    $2^{nd}$ & 54.77 & 2.11 & 0.9769 \\
    $3^{rd}$ & 56.45 & 1.87 & 0.95776 \\
    $1^{st} + 2^{nd}$ & 58.13 & 5.96 & \textbf{0.9853} \\
    $2^{nd} + 3^{rd}$ & 57.84 & 1.97 & 0.9784 \\
    $1^{st} + 2^{nd} + 3^{rd}$ & \textbf{59.17} & \textbf{1.76} & 0.9638 \\
    \hline
    \end{tabular}
\end{table}

To explore the effect of the level of the DeepPro structure, we removed some levels shown in Fig. \ref{fig:architecture}, and the comparison results are in Table \ref{tab:level}. It can be observed that fewer levels result in higher false alarm rates and lower detection rates, while the multi-level structure significantly improves the detection performance. This is because, with the decrease in the spatial resolution of the features, the process of a target moving through a specified location becomes slow and obvious. Meanwhile, the interference of noise is reduced since most noise is spatially uncorrelated. Consequently, the signal of the target in the temporal profile becomes broader and more distinctive, making it easier for the model to probe the anomaly signals. Furthermore, the multi-level architecture enables the network to capture more comprehensive temporal profile information of the target regions, enriching the feature representation.

\subsection{Robustness to noise intensities}\label{Robust_toNoise}
\begin{figure}[!t]
    \centering
    \includegraphics[width=1\linewidth]{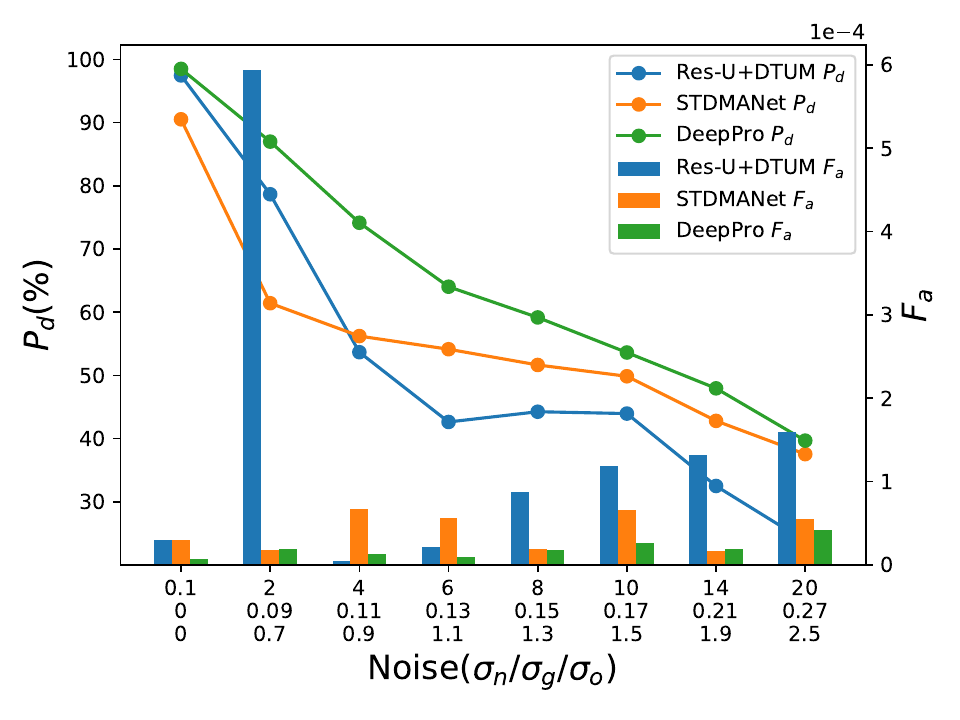}
    \vspace{-8mm}
    \caption{Results of different networks in scenes under different noise intensity conditions.
    }
    \label{fig:Robust-to-N}
\end{figure}
We explored the robustness of different networks to the noises with different intensities. We changed the standard deviation $\sigma_n$ of Gaussian white noise, the distribution range $\sigma_g$ of the multiplicative noise (related to the non-uniformity), and the standard deviation $\sigma_o$ of the additive noise to different values. Then, we evaluated the detection performances of different networks on different conditions, as shown in Fig. \ref{fig:Robust-to-N}.

As the intensity of the noise increases, the detection performances of all methods experience degradations. Under extreme noise conditions, Res-UNet+DTUM suffers an obvious increase in terms of $F_a$, whereas our DeepPro consistently maintains a low $F_a$. In all conditions, our DeepPro achieves significantly superior detection performance. That is because, the temporal probe mechanism can learn more reliable and richer correlation features of different signals in the temporal profile. Besides, the characteristics of long-term temporal variation enable robust discrimination between varying levels of noise, clutter, and target signals. Therefore, our DeepPro is robust to different noise intensity conditions.

\subsection{DeepPro-Plus: Combination with little spatial information}
As aforementioned, we confirm that temporal profile information is very important for IRST detection. Only leveraging the temporal profile information can achieve superior performance as compared with the state-of-the-art methods that use spatial or local spatial-temporal information. Then, a question arises naturally: How about combining temporal profile information with little spatial information? To answer it, we further developed another version of our DeepPro, called DeepPro-Plus, which replaces all temporal convolutions in temporal feature extraction with spatial-temporal convolution and retains only the first level in DeepPro. All TD-ResBlocks in DeepPro are changed into spatial temporal difference residual blocks (STD-ResBlock). The details are described in the Appendices.

\subsubsection{Comparison to the state-of-the-arts}
We compared our DeepPro-Plus with the same state-of-the-art methods in Subsection \ref{ComparisonSORT}. The quantitative results achieved by DeepPro-Plus in different scenes and on different targets are shown in Table \ref{tab:SOTAv2}. Furthermore, we expanded our evaluation  on a more diverse, large-scale dataset (the IRSatVideo-LEO dataset \cite{ying2025infrared}) with smaller targets for a thorough comparison. See the Appendices for further results.

\begin{table*}[t!]
\caption{Detection results achieved by different versions of our DeepPro on the NUDT-MIRSDT dataset and the NUDT-MIRSDT-HiNo dataset.}\label{tab:SOTAv2}
\centering
\renewcommand{\arraystretch}{1.1}
\begin{tabular}{c c c c c c c c c c c c}
\hline
\multirow{2}*{Models} & \multicolumn{2}{c}{$SNR\leq 3$} & \multicolumn{3}{c}{NUDT-MIRSDT} & \multicolumn{3}{c}{NUDT-MIRSDT-HiNo} & \multirow{2}*{\#Params (M)} & \multirow{2}*{GFLOPs} &\multirow{2}*{FPS}\\ 
\cline{2-9}
  & $P_d$  & $F_a $       & $P_d$  & $F_a$ & AUC       & $P_d$  & $F_a$  & AUC & & \\ \hline
DeepPro & 95.84 & 0.52 & 98.50 & 0.72 & 0.9973 & 59.17 & 1.76 & 0.9638 & 0.197 & 1.01 & 184.55 \\
DeepPro-Plus & 99.24 & 1.65 & 99.71 & 2.69 & 0.9978 & 76.23 & 1.69 & 0.9171 & 0.284 & 3.89 & 224.05 \\
DeepPro+S & 98.68 & 0.83 & 99.60 & 1.02 & 0.9976 & 76.52 & 1.03 & 0.9461 & 0.757 & 5.22 & 163.69 \\ \hline
\end{tabular}
\end{table*}

In all scenarios, DeepPro-Plus achieves the highest detection performance with absolute superiority in almost all aspects as compared with the state-of-the-art methods and DeepPro in Table \ref{tab:SOTA1}. Especially in scenes with strong noise, DeepPro with little spatial calculations improves $P_d$ by more than 20\%, while maintaining a lower $F_a$. Although GFLOPs and the number of parameters are increased, the increment is negligible compared to other existing methods. That is because, the spatial receptive field of DeepPro-Plus is only $19\times 19$ in theory. Besides, the real-time performance (i.e., FPS) is improved because of less-level structures. Therefore, it can be concluded that combining temporal profile information with little spatial information is further beneficial for IRST detection. In addition, the temporal profile information is more important.

\subsubsection{Ablation study}\label{Modelv2_analysis}
In this part, we validate that one level is enough when combining little spatial information, and a deep spatial-temporal feature extraction structure is unadvisable design for IRST detection.

\textbf{1) Number of levels:}
From DeepPro to DeepPro-Plus, we maintained only one level in the structure. To explain its effectiveness, we compared DeepPro-Plus with DeepPro+S obtained by only replacing all temporal convolutions with spatial-temporal convolutions in DeepPro. The results are shown in Table \ref{tab:SOTAv2}. We find that the performance gap in IRST detection between DeepPro+S and DeepPro-Plus is minimal, but the structure with three levels (i.e, DeepPro+S) requires substantially more computational overhead. Consequently, little spatial information in one level is enough to help the network to further capture richer temporal profile information of the target region. This reveals a notion that extensive spatial receptive field and complex feature extraction are unessential, but focusing on the temporal profile is crucial.

\begin{table}[t]
    \renewcommand\arraystretch{1.16}
    \caption{Detection results achieved by several variants of DeepPro-Plus with different numbers of STD-Resblocks on the NUDT-MIRSDT-HiNo dataset. The best results are in bold.} \label{tab:numSTD}
    \centering
    \renewcommand{\arraystretch}{1.1}
    \begin{tabular}{c c c c c c}
    \hline
    $n$ & $P_d$ & $F_{a}$ & AUC & \#Params (M) & GFLOPs \\
    \hline
    2 & \textbf{76.23} & \textbf{1.69} & \textbf{0.9171} & \textbf{0.284} & \textbf{3.89} \\
    3 & 73.05 & 2.26 & 0.9100 & 0.515 & 7.68 \\
    4 & 72.76 & 3.46 & 0.9114 & 0.745 & 11.47 \\ \hline
    \end{tabular}
\end{table}

\textbf{2) Depth of spatial-temporal feature extraction:}
To explore the effect of the depth of the spatial-temporal feature extraction structure, we adjusted the number of STD-ResBlocks $n$ to 2, 3, and 4. The results are shown in Table \ref{tab:numSTD}. It can be observed that when $n=2$, the network performs best with a $3.18\%$ $P_d$ increase and a $0.57\times10^{-5}$ $F_a$ decrease as compared with the second best results. This reveals that the deeper structure (e.g., more STD-ResBlocks) or the larger spatial receptive field does not bring a performance improvement in IRST detection, but an efficiency decline. These results can also explain \textbf{Phenomenon 1} that the influential pixels are concentrated on the temporal profile of the target since the temporal profile information is more important.

\subsection{Performance in the real world}
\begin{figure}[t]
    \centering
    \includegraphics[width=1\linewidth]{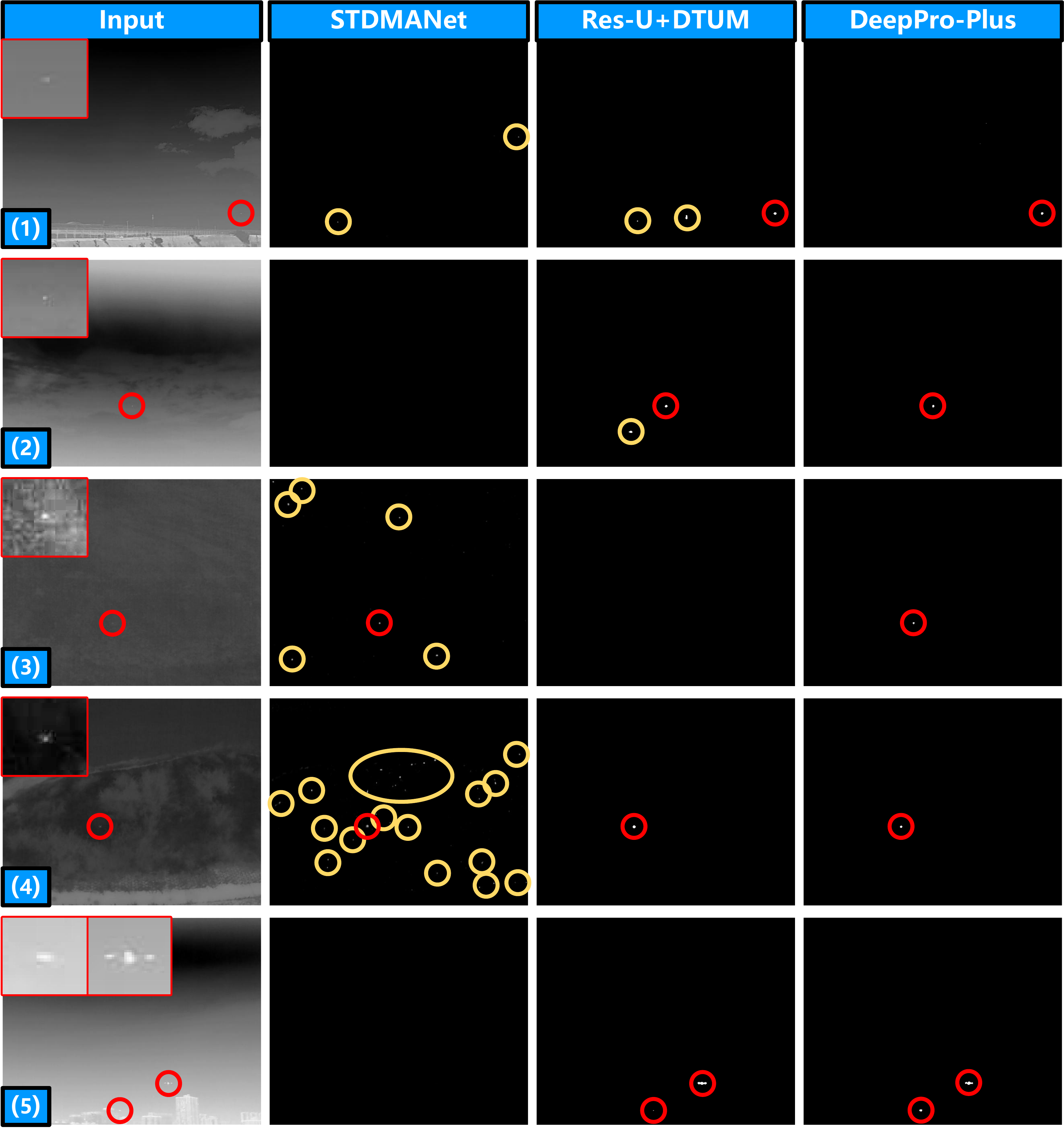}
    \caption{Visual comparison on the real-world RGBT-Tiny dataset. For better visualization, the target area is enlarged in the top-left corner and highlighted with a red circle. The false alarm area is marked with a yellow circle.
    }
    \label{fig:visual_real}
\end{figure}

We tested the detection performance of our method on real scenarios from the RGBT-Tiny dataset \cite{ying2025visible}. We selected 45 and 15 scenarios containing IRSTs from its training set and test set, respectively. In the dataset, there are various small targets (e.g., ships, pedestrians, cars, drones, and planes) with different moving speeds, and diverse scenarios with dynamic backgrounds and camera movements. For training, we generated rough Gaussian masks as labels according to the bounding boxes provided by its authors. The quantitative results are shown in Table \ref{tab:SOTA_real}, and the qualitative results are shown in Fig. \ref{fig:visual_real}.

\begin{table}[t!]
\caption{Detection results achieved by different state-of-the-art methods on the real-world RGBT-Tiny dataset.}\label{tab:SOTA_real}
\centering
\renewcommand{\arraystretch}{1.1}
\begin{tabular}{c l c c c}
\hline
\multicolumn{2}{c}{Methods} & $P_d$  & $F_a$ & AUC \\ \hline  
\multirow{5}{*}{\rotatebox{90}{\textit{SF}}} & AGPCNet \cite{zhang2023attention} & 51.05 & 10436.86 & 0.7296 \\  
        & MSHNet \cite{liu2024infraredMSHNet} & 73.45 & 14.37 & 0.9405 \\
        & SCTransNet \cite{yuan2024sctransnet} & 76.13 & 6.35 & 0.8852 \\
        & RPCANet \cite{wu2024rpcanet} & 1.50 & 0.03 & 0.4824 \\
        & ILNet \cite{li2025ilnet} & 72.72 & 16.57 & 0.9407 \\
\hline
\multirow{5}{*}{\rotatebox{90}{\textit{MF}}} & Res-U+DTUM \cite{li2023direction} & 68.81 & 4.79 & 0.8462 \\  
        & STDMANet \cite{yan2023stdmanet} & 45.76 & 4.61 & 0.8865 \\  
        & Res-U+RFR \cite{ying2025infrared} & 59.42 & 3.48 & 0.8155 \\  
        & DQAligner \cite{deng2026learning} & 58.08 & 97.57 & 0.8264 \\  
        & DeepPro-Plus & 77.83 & 4.47 & 0.9730 \\ \hline  
\end{tabular}
\end{table}

From Table \ref{tab:SOTA_real}, it can be found that the $P_d$ of DeepPro-Plus is 9.02\% higher than that of the sub-optimal multi-frame method (i.e., DTUM) in the case of close $F_a$. This shows that our method can achieve state-of-the-art detection performance in the real world. From Fig. \ref{fig:visual_real}, we can observe that DeepPro-Plus achieves good detection performance in various scenes. In scene (3), the background and target are very dim and the noise is relatively prominent. Other methods produce more false alarms or miss detection, while our method can still maintain good performance. As for scene (4), the background is changed due to camera movement. STDMANet produces a lot of false alarms, while our method and DTUM can still adapt and perform well.

\subsection{Failure cases}
\begin{figure}[t]
    \centering
    \includegraphics[width=1\linewidth]{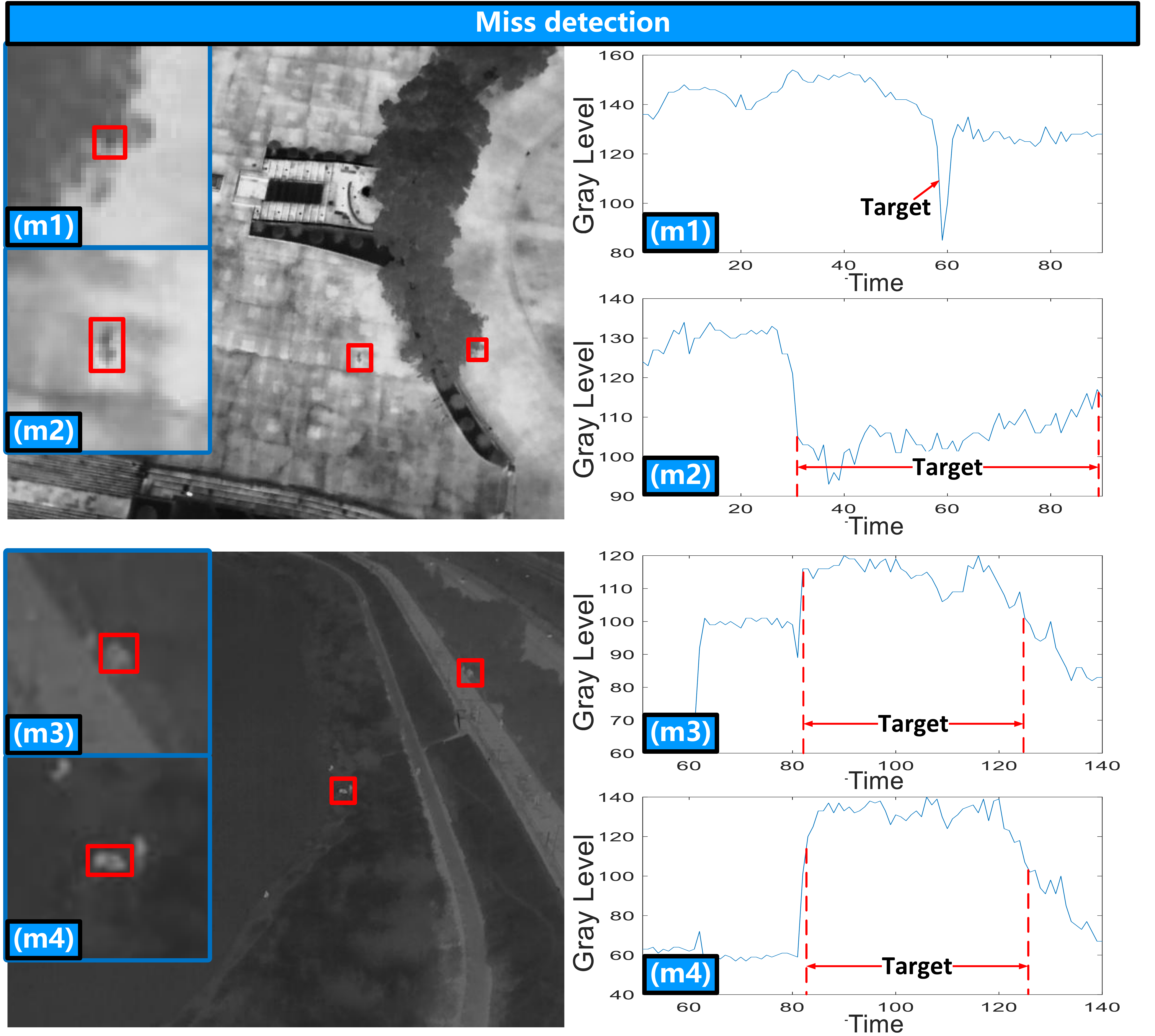}
    \caption{Failure cases in the real world and their gray variations in the temporal profile. (m1) is a pedestrian obscured by trees. (m2), (m3), and (m4) are stationary pedestrians. Their signals in the temporal profile are different from the ordinary moving target signals as shown in Figs. \ref{fig:toy_examples}, \ref{fig:temporal_profile_analysis_real} and \ref{fig:temporal_profile_analysis_corr}.
    }
    \label{fig:visual_lim}
\end{figure}

When testing our method in real-world scenarios, there are some failure cases of miss detection, as shown in Fig. \ref{fig:visual_lim}.

\textbf{Miss detecting obscured targets.} The target (m1) in Fig. \ref{fig:visual_lim} is obscured by trees, leading to an incomplete signal in the temporal profile. The target signal is so narrow as a spike. This target is missed, and the reason can be twofold. On the one hand, this may be due to the incomplete signal. On the other hand, it may be due to the lack of such samples in the training set. 

\textbf{Miss detecting stationary targets.} The targets (m2), (m3), and (m4) are stationary during a period of time. From their temporal profiles, it can be observed that there is no regular temporal profile information like that of the moving targets. For a long time (over 40 frames), the signal is maintained at a fixed gray level. It is hard to detect stationary objects from the temporal profile. Therefore, our method is targeted at moving IRSTs. 

\section{Conclusion}
In this paper, we consider the essential domain and information for precise, robust, and efficient infrared small target (IRST) detection. We observe the targets from the temporal profile. We reveal some overlooked but essential temporal profile information, that is, the global temporal saliency of the target and the correlation information between target signal and other signals. We built the first prediction attribution tool, and validated the robustness of the temporal profile information (long-term temporal information) by applying it to IRST detection networks. Based on our research, we remodel the IRST detection task as a one-dimensional signal anomaly detection task and propose an efficient and effective IRST detection network, deep temporal probe network (DeepPro), in which the calculations (i.e., multiplication and addition) are only in the time dimension. Numerous experiments on widely-used benchmarks were conducted to confirm the effectiveness of our methods. It is exciting that our DeepPro outperforms the state-of-the-art methods especially on dim targets and in complex scenarios with extremely few parameters and extremely high processing efficiency. Through this work, we reveal the paramount importance of temporal profile information.

\bibliographystyle{IEEEtran}
\bibliography{ref_abb}

@article{reed1983application,
  title={Application of three-dimensional filtering to moving target detection},
  author={Reed, Irving S and Gagliardi, Robert M and Shao, HM},
  journal={IEEE TAES},
  number={6},
  pages={898--905},
  year={1983},
  publisher={IEEE}
}

@article{reed1990recursive,
  title={A recursive moving-target-indication algorithm for optical image sequences},
  author={Reed, Irving S and Gagliardi, Robert M and Stotts, Larry B},
  journal={IEEE TAES},
  volume={26},
  number={3},
  pages={434--440},
  year={1990},
  publisher={IEEE}
}

@article{soni1993performance,
  title={Performance evaluation of 2-D adaptive prediction filters for detection of small objects in image data},
  author={Soni, Tarun and Zeidler, James R and Ku, Walter H},
  journal={IEEE TIP},
  volume={2},
  number={3},
  pages={327--340},
  year={1993},
  publisher={IEEE}
}

@inproceedings{deshpande1999max,
  title={Max-mean and max-median filters for detection of small targets},
  author={Deshpande, Suyog D and Er, Meng Hwa and Venkateswarlu, Ronda and Chan, Philip},
  booktitle={Signal and Data Processing of Small Targets 1999},
  volume={3809},
  pages={74--83},
  year={1999},
}

@article{bai2010analysis,
  title={Analysis of new top-hat transformation and the application for infrared dim small target detection},
  author={Bai, Xiangzhi and Zhou, Fugen},
  journal={PR},
  volume={43},
  number={6},
  pages={2145--2156},
  year={2010},
  publisher={Elsevier}
}

@article{2013Infrared_IPI,
  title={Infrared patch-image model for small target detection in a single image},
  author={ Gao, C.  and  Meng, D.  and  Yang, Y.  and  Wang, Y.  and  Zhou, X.  and  Hauptmann, A. G. },
  journal={IEEE TIP},
  volume={22},
  number={12},
  pages={4996-5009},
  year={2013},
}

@article{2018Infrared_NRAM,
  title={Infrared small target detection via non-convex rank approximation minimization joint l2,1 norm},
  author={Zhang, L.  and  Peng, L.  and  Zhang, T.  and  Cao, S.  and  Peng, Z. },
  journal={RS},
  volume={10},
  number={11},
  year={2018},
}

@article{dai2017reweighted_RIPT,
  title={Reweighted infrared patch-tensor model with both nonlocal and local priors for single-frame small target detection},
  author={Dai, Yimian and Wu, Yiquan},
  journal={IEEE JSTARS},
  volume={10},
  number={8},
  pages={3752--3767},
  year={2017},
  publisher={IEEE}
}

@article{chen2013local,
  title={A local contrast method for small infrared target detection},
  author={Chen, CL Philip and Li, Hong and Wei, Yantao and Xia, Tian and Tang, Yuan Yan},
  journal={IEEE TGRS},
  volume={52},
  number={1},
  pages={574--581},
  year={2013},
  publisher={IEEE}
}

@article{liu2018infrared,
  title={Infrared small target detection based on flux density and direction diversity in gradient vector field},
  author={Liu, Depeng and Cao, Lei and Li, Zhengzhou and Liu, Tianmei and Che, Peng},
  journal={IEEE JSTARS},
  volume={11},
  number={7},
  pages={2528--2554},
  year={2018},
  publisher={IEEE}
}

@article{wei2016multiscale,
  title={Multiscale patch-based contrast measure for small infrared target detection},
  author={Wei, Yantao and You, Xinge and Li, Hong},
  journal={PR},
  volume={58},
  pages={216--226},
  year={2016},
  publisher={Elsevier}
}

@article{2015Infrared_PLCM,
  title={Infrared small dim target detection based on local contrast combined with region saliency},
  author={ Wang, X.  and  Peng, Z.  and  Zhang, P.  and  Meng, Y. },
  journal={High Power Laser and Particle Beams},
  volume={27},
  number={9},
  year={2015},
}

@article{zhu2020balanced,
  title={Balanced ring top-hat transformation for infrared small-target detection with guided filter kernel},
  author={Zhu, Hu and Zhang, Jieke and Xu, Guoxia and Deng, Lizhen},
  journal={IEEE TAES},
  volume={56},
  number={5},
  pages={3892--3903},
  year={2020},
  publisher={IEEE}
}

@article{zhao2021three,
  title={Three-order tensor creation and tucker decomposition for infrared small-target detection},
  author={Zhao, Mingjing and Li, Wei and Li, Lu and Ma, Pengge and Cai, Zhaoquan and Tao, Ran},
  journal={IEEE TGRS},
  volume={60},
  pages={1--16},
  year={2021},
  publisher={IEEE}
}

@article{yang2023small,
  title={Small maritime target detection using gradient vector field characterization of infrared image},
  author={Yang, Ping and Dong, Lili and Xu, Wenhai},
  journal={IEEE JSTARS},
  volume={16},
  pages={1827--1841},
  year={2023},
  publisher={IEEE}
}

@article{sun2020infrared,
  title={Infrared dim and small target detection via multiple subspace learning and spatial-temporal patch-tensor model},
  author={Sun, Yang and Yang, Jungang and An, Wei},
  journal={IEEE TGRS},
  volume={59},
  number={5},
  pages={3737--3752},
  year={2020},
  publisher={IEEE}
}

@inproceedings{bearman2016s,
  title={What's the point: Semantic segmentation with point supervision},
  author={Bearman, Amy and Russakovsky, Olga and Ferrari, Vittorio and Fei-Fei, Li},
  booktitle={ECCV},
  pages={549--565},
  year={2016},
}

@article{du2021spatial,
  title={A spatial-temporal feature-based detection framework for infrared dim small target},
  author={Du, Jinming and Lu, Huanzhang and Zhang, Luping and Hu, Moufa and Chen, Sheng and Deng, Yingjie and Shen, Xinglin and Zhang, Yu},
  journal={IEEE TGRS},
  volume={60},
  pages={1--12},
  year={2021},
  publisher={IEEE}
}

@article{dai2021attentional,
  title={Attentional local contrast networks for infrared small target detection},
  author={Dai, Yimian and Wu, Yiquan and Zhou, Fei and Barnard, Kobus},
  journal={IEEE TGRS},
  volume={59},
  number={11},
  pages={9813--9824},
  year={2021},
  publisher={IEEE}
}

@inproceedings{dai2021asymmetric,
  title={Asymmetric contextual modulation for infrared small target detection},
  author={Dai, Yimian and Wu, Yiquan and Zhou, Fei and Barnard, Kobus},
  booktitle={WACV},
  pages={950--959},
  year={2021}
}

@inproceedings{wu2024rpcanet,
  title={{RPCANet}: Deep unfolding {RPCA} based infrared small target detection},
  author={Wu, Fengyi and Zhang, Tianfang and Li, Lei and Huang, Yian and Peng, Zhenming},
  booktitle={WACV},
  pages={4809--4818},
  year={2024}
}

@article{li2022dense,
  title={Dense nested attention network for infrared small target detection},
  author={Li, Boyang and Xiao, Chao and Wang, Longguang and Wang, Yingqian and Lin, Zaiping and Li, Miao and An, Wei and Guo, Yulan},
  journal={IEEE TIP},
  volume={32},
  pages={1745--1758},
  year={2022},
  publisher={IEEE}
}

@inproceedings{zhang2022isnet,
  title={{ISNet}: Shape matters for infrared small target detection},
  author={Zhang, Mingjin and Zhang, Rui and Yang, Yuxiang and Bai, Haichen and Zhang, Jing and Guo, Jie},
  booktitle={CVPR},
  pages={877--886},
  year={2022}
}

@article{sun2021small,
  title={Small aerial target detection for airborne infrared detection systems using LightGBM and trajectory constraints},
  author={Sun, Xiaoliang and Guo, Liangchao and Zhang, Wenlong and Wang, Zi and Yu, Qifeng},
  journal={IEEE JSTARS},
  volume={14},
  pages={9959--9973},
  year={2021},
  publisher={IEEE}
}

@inproceedings{gao2017tvpcf,
  title={{TVPCF}: A spatial and temporal filter for small target detection in {IR} images},
  author={Gao, Jinyan and Lin, Zaiping and Guo, Yulan and An, Wei},
  booktitle={DICTA},
  pages={1--7},
  year={2017},
}

@article{li2021infrared,
  title={Infrared maritime dim small target detection based on spatiotemporal cues and directional morphological filtering},
  author={Li, Yongsong and Li, Zhengzhou and Zhang, Chao and Luo, Zefeng and Zhu, Yong and Ding, Zhiquan and Qin, Tianqi},
  journal={Infrared Phys. Technol.},
  volume={115},
  pages={103657},
  year={2021},
  publisher={Elsevier}
}

@article{zhang2020edge,
  title={Edge and corner awareness-based spatial-temporal tensor model for infrared small-target detection},
  author={Zhang, Ping and Zhang, Lingyi and Wang, Xiaoyang and Shen, Fengcan and Pu, Tian and Fei, Chun},
  journal={IEEE TGRS},
  volume={59},
  number={12},
  pages={10708--10724},
  year={2020},
  publisher={IEEE}
}

@article{wang2021infrared,
  title={Infrared small target detection using nonoverlapping patch spatial-temporal tensor factorization with capped nuclear norm regularization},
  author={Wang, Guanghui and Tao, Bingjie and Kong, Xuan and Peng, Zhenming},
  journal={IEEE TGRS},
  volume={60},
  pages={1--17},
  year={2021},
  publisher={IEEE}
}

@inproceedings{he2015delving,
  title={Delving deep into rectifiers: Surpassing human-level performance on imagenet classification},
  author={He, Kaiming and Zhang, Xiangyu and Ren, Shaoqing and Sun, Jian},
  booktitle={ICCV},
  pages={1026--1034},
  year={2015}
}

@article{sun2023receptive,
  title={Receptive-field and direction induced attention network for infrared dim small target detection with a large-scale dataset IRDST},
  author={Sun, Heng and Bai, Junxiang and Yang, Fan and Bai, Xiangzhi},
  journal={IEEE TGRS},
  volume={61},
  pages={1--13},
  year={2023},
  publisher={IEEE}
}

@article{luo2022imnn,
  title={{IMNN-LWEC}: A novel infrared small target detection based on spatial-temporal tensor model},
  author={Luo, Yuan and Li, Xiaorun and Chen, Shuhan and Xia, Chaoqun and Zhao, Liaoying},
  journal={IEEE TGRS},
  volume={60},
  pages={1--22},
  year={2022},
  publisher={IEEE}
}

@article{li2023sparse,
  title={Sparse regularization-based spatial-temporal twist tensor model for infrared small target detection},
  author={Li, Jie and Zhang, Ping and Zhang, Lingyi and Zhang, Zhiyuan},
  journal={IEEE TGRS},
  volume={61},
  pages={1--17},
  year={2023},
  publisher={IEEE}
}

@article{liu2025graph,
  title={Graph Laplacian regularization for fast infrared small target detection},
  author={Liu, Ting and Liu, Yongxian and Yang, Jungang and Li, Boyang and Wang, Yingqian and An, Wei},
  journal={PR},
  volume={158},
  pages={111077},
  year={2025},
  publisher={Elsevier}
}

@article{wu2022uiu,
  title={{UIU-Net: U-Net in U-Net} for infrared small object detection},
  author={Wu, Xin and Hong, Danfeng and Chanussot, Jocelyn},
  journal={IEEE TIP},
  volume={32},
  pages={364--376},
  year={2022},
  publisher={IEEE}
}

@article{zhou2023deep,
  title={Deep low-rank and sparse patch-image network for infrared dim and small target detection},
  author={Zhou, Xinyu and Li, Peng and Zhang, Ye and Lu, Xin and Hu, Yue},
  journal={IEEE TGRS},
  volume={61},
  pages={1--14},
  year={2023},
  publisher={IEEE}
}

@article{liu2023infrared,
  title={Infrared small and dim target detection with transformer under complex backgrounds},
  author={Liu, Fangcen and Gao, Chenqiang and Chen, Fang and Meng, Deyu and Zuo, Wangmeng and Gao, Xinbo},
  journal={IEEE TIP},
  volume={32},
  pages={5921--5932},
  year={2023},
  publisher={IEEE}
}

@article{yan2023stdmanet,
  title={{STDMANet}: Spatio-temporal differential multiscale attention network for small moving infrared target detection},
  author={Yan, Puti and Hou, Runze and Duan, Xuguang and Yue, Chengfei and Wang, Xin and Cao, Xibin},
  journal={IEEE TGRS},
  volume={61},
  pages={1--16},
  year={2023},
  publisher={IEEE}
}

@article{fawcett2006introduction,
  title={An introduction to {ROC} analysis},
  author={Fawcett, Tom},
  journal={Pattern Recog. Lett.},
  volume={27},
  number={8},
  pages={861--874},
  year={2006},
  publisher={Elsevier}
}

@article{deng2017entropy,
  title={Entropy-based window selection for detecting dim and small infrared targets},
  author={Deng, He and Sun, Xianping and Liu, Maili and Ye, Chaohui and Zhou, Xin},
  journal={PR},
  volume={61},
  pages={66--77},
  year={2017},
  publisher={Elsevier}
}

@article{wei2020end,
  title={End-to-end video saliency detection via a deep contextual spatiotemporal network},
  author={Wei, Lina and Zhao, Shanshan and Bourahla, Omar Farouk and Li, Xi and Wu, Fei and Zhuang, Yueting and Han, Junwei and Xu, Mingliang},
  journal={IEEE TNNLS},
  volume={32},
  number={4},
  pages={1691--1702},
  year={2020},
  publisher={IEEE}
}

@article{cheng2023towards,
  title={Towards large-scale small object detection: Survey and benchmarks},
  author={Cheng, Gong and Yuan, Xiang and Yao, Xiwen and Yan, Kebing and Zeng, Qinghua and Xie, Xingxing and Han, Junwei},
  journal={IEEE TPAMI},
  volume={45},
  number={11},
  pages={13467--13488},
  year={2023},
  publisher={IEEE}
}

@inproceedings{rahman2016optimizing,
  title={Optimizing intersection-over-union in deep neural networks for image segmentation},
  author={Rahman, Md Atiqur and Wang, Yang},
  booktitle={ISVC},
  pages={234--244},
  year={2016},
}

@inproceedings{xiao2018weighted,
  title={Weighted {Res-UNet} for high-quality retina vessel segmentation},
  author={Xiao, Xiao and Lian, Shen and Luo, Zhiming and Li, Shaozi},
  booktitle={ITME},
  pages={327--331},
  year={2018},
}

@article{li2023direction,
  title={Direction-coded temporal {U-shape} module for multiframe infrared small target detection},
  author={Li, Ruojing and An, Wei and Xiao, Chao and Li, Boyang and Wang, Yingqian and Li, Miao and Guo, Yulan},
  journal={IEEE TNNLS},
  year={2025},
  volume={36},
  number={1},
  pages={555--568},
  doi={10.1109/TNNLS.2023.3331004}}

@article{tong2024st,
  title={{ST-Trans}: Spatial-temporal transformer for infrared small target detection in sequential images},
  author={Tong, Xiaozhong and Zuo, Zhen and Su, Shaojing and Wei, Junyu and Sun, Xiaoyong and Wu, Peng and Zhao, Zongqing},
  journal={IEEE TGRS},
  year={2024},
  publisher={IEEE}
}

@inproceedings{ying2023mapping,
  title={Mapping degeneration meets label evolution: Learning infrared small target detection with single point supervision},
  author={Ying, Xinyi and Liu, Li and Wang, Yingqian and Li, Ruojing and Chen, Nuo and Lin, Zaiping and Sheng, Weidong and Zhou, Shilin},
  booktitle={CVPR},
  pages={15528--15538},
  year={2023}
}

@inproceedings{li2023monte,
  title={{Monte Carlo} linear clustering with single-point supervision is enough for infrared small target detection},
  author={Li, Boyang and Wang, Yingqian and Wang, Longguang and Zhang, Fei and Liu, Ting and Lin, Zaiping and An, Wei and Guo, Yulan},
  booktitle={ICCV},
  pages={1009--1019},
  year={2023}
}

@article{wu2023infrared,
  title={Infrared small target detection using spatiotemporal {4-D} tensor train and ring unfolding},
  author={Wu, Fengyi and Yu, Hang and Liu, Anran and Luo, Junhai and Peng, Zhenming},
  journal={IEEE TGRS},
  volume={61},
  pages={1--22},
  year={2023},
  publisher={IEEE}
}

@article{guo2023small,
  title={Small aerial target detection using trajectory hypothesis and verification},
  author={Guo, Liangchao and Sun, Xiaoliang and Zhang, Wenlong and Li, Zhang and Yu, Qifeng},
  journal={IEEE TGRS},
  volume={61},
  pages={1--14},
  year={2023},
  publisher={IEEE}
}

@article{luo20234dst,
  title={{4DST-BTMD}: An infrared small target detection method based on {4-D} data-sphered space},
  author={Luo, Yuan and Li, Xiaorun and Chen, Shuhan and Xia, Chaoqun},
  journal={IEEE TGRS},
  volume={62},
  pages={1--20},
  year={2023},
  publisher={IEEE}
}

@article{luo2023spatial,
  title={Spatial-temporal tensor representation learning with priors for infrared small target detection},
  author={Luo, Yuan and Li, Xiaorun and Yan, Yunfeng and Xia, Chaoqun},
  journal={IEEE TAES},
  volume={59},
  number={6},
  pages={9598--9620},
  year={2023},
  publisher={IEEE}
}

@article{liuT2023infrared,
  title={Infrared small target detection via nonconvex tensor tucker decomposition with factor prior},
  author={Liu, Ting and Yang, Jungang and Li, Boyang and Wang, Yingqian and An, Wei},
  journal={IEEE TGRS},
  volume={61},
  pages={1--17},
  year={2023},
  publisher={IEEE}
}

@article{zhang2023attention,
  title={Attention-guided pyramid context networks for detecting infrared small target under complex background},
  author={Zhang, Tianfang and Li, Lei and Cao, Siying and Pu, Tian and Peng, Zhenming},
  journal={IEEE TAES},
  volume={59},
  number={4},
  pages={4250--4261},
  year={2023},
  publisher={IEEE}
}

@article{zhang2022learning,
  title={Learning nonlocal quadrature contrast for detection and recognition of infrared rotary-wing {UAV} targets in complex background},
  author={Zhang, Yu and Zhang, Yan and Fu, Ruigang and Shi, Zhiguang and Zhang, Jinghua and Liu, Di and Du, Jinming},
  journal={IEEE TGRS},
  volume={60},
  pages={1--19},
  year={2022},
  publisher={IEEE}
}

@article{vaswani2017attention,
  title={Attention is all you need},
  author={Vaswani, Ashish and Shazeer, Noam and Parmar, Niki and Uszkoreit, Jakob and Jones, Llion and Gomez, Aidan N and Kaiser, {\L}ukasz and Polosukhin, Illia},
  journal={NeurIPS},
  volume={30},
  year={2017}
}

@inproceedings{su2021pixel,
  title={Pixel difference networks for efficient edge detection},
  author={Su, Zhuo and Liu, Wenzhe and Yu, Zitong and Hu, Dewen and Liao, Qing and Tian, Qi and Pietik{\"a}inen, Matti and Liu, Li},
  booktitle={ICCV},
  pages={5117--5127},
  year={2021}
}

@book{2016InfraredTargetDetection,
    title={Infrared target detection},
    author={Chen, Qian and Qian, Weixian and Zhang, Wenwen},
    publisher={Publishing House of Electronics Industry},
    year={2016},
}

@inproceedings{wang2015dsNoise,
  author={Wang, Yan and Xie, Xiaofang},
  booktitle={ICNISC},
  title={Dynamic Simulation of Noise in 3D Infrared Scene Simulation},
  year={2015},
  pages={251-254},
  doi={10.1109/ICNISC.2015.101}}

@article{wang2022noise,
  title={Noise parameter estimation two-stage network for single infrared dim small target image destriping},
  author={Wang, Teliang and Yin, Qian and Cao, Fanzhi and Li, Miao and Lin, Zaiping and An, Wei},
  journal={RS},
  volume={14},
  number={19},
  pages={5056},
  year={2022},
  publisher={MDPI}
}

@article{liu2015moving,
  title={Moving target detection by nonlinear adaptive filtering on temporal profiles in infrared image sequences},
  author={Liu, Delian and Li, Zhaohui and Wang, Xiaorui and Zhang, Jianqi},
  journal={Infrared Phys. Technol.},
  volume={73},
  pages={41--48},
  year={2015},
  publisher={Elsevier}
}

@inproceedings{sundararajan2017axiomatic,
  title={Axiomatic attribution for deep networks},
  author={Sundararajan, Mukund and Taly, Ankur and Yan, Qiqi},
  booktitle={ICML},
  pages={3319--3328},
  year={2017},
}

@inproceedings{xu2020attribution,
  title={Attribution in scale and space},
  author={Xu, Shawn and Venugopalan, Subhashini and Sundararajan, Mukund},
  booktitle={CVPR},
  pages={9680--9689},
  year={2020}
}

@article{sturmfels2020visualizing,
  title={Visualizing the impact of feature attribution baselines},
  author={Sturmfels, Pascal and Lundberg, Scott and Lee, Su-In},
  journal={Distill},
  volume={5},
  number={1},
  pages={e22},
  year={2020}
}

@article{miglani2020investigating,
  title={Investigating saturation effects in integrated gradients},
  author={Miglani, Vivek and Kokhlikyan, Narine and Alsallakh, Bilal and Martin, Miguel and Reblitz-Richardson, Orion},
  journal={arXiv preprint arXiv:2010.12697},
  year={2020}
}

@article{hoyer2019grid,
  title={Grid saliency for context explanations of semantic segmentation},
  author={Hoyer, Lukas and Munoz, Mauricio and Katiyar, Prateek and Khoreva, Anna and Fischer, Volker},
  journal={NeurIPS},
  volume={32},
  year={2019}
}

@article{omeiza2019smooth,
  title={{Smooth Grad-CAM++}: An enhanced inference level visualization technique for deep convolutional neural network models},
  author={Omeiza, Daniel and Speakman, Skyler and Cintas, Celia and Weldermariam, Komminist},
  journal={arXiv preprint arXiv:1908.01224},
  year={2019}
}

@article{chrabaszcz2024aggregated,
  title={Aggregated attributions for explanatory analysis of 3d segmentation models},
  author={Chrabaszcz, Maciej and Baniecki, Hubert and Komorowski, Piotr and P{\l}otka, Szymon and Biecek, Przemyslaw},
  journal={arXiv preprint arXiv:2407.16653},
  year={2024}
}

@article{ancona2019gradient,
  title={Gradient-based attribution methods},
  author={Ancona, Marco and Ceolini, Enea and {\"O}ztireli, Cengiz and Gross, Markus},
  journal={Explainable AI: Interpreting, explaining and visualizing deep learning},
  pages={169--191},
  year={2019},
  publisher={Springer}
}

@article{gevrey2003review,
  title={Review and comparison of methods to study the contribution of variables in artificial neural network models},
  author={Gevrey, Muriel and Dimopoulos, Ioannis and Lek, Sovan},
  journal={Ecol. Modell.},
  volume={160},
  number={3},
  pages={249--264},
  year={2003},
  publisher={Elsevier}
}

@inproceedings{simonyan2014deep,
  title={Deep inside convolutional networks: Visualising image classification models and saliency maps},
  author={Simonyan, Karen and Vedaldi, Andrea and Zisserman, Andrew},
  booktitle={ICLR},
  year={2014},
}

@inproceedings{zeiler2014visualizing,
  title={Visualizing and understanding convolutional networks},
  author={Zeiler, Matthew D and Fergus, Rob},
  booktitle={ECCV},
  pages={818--833},
  year={2014},
}

@article{abhishek2022attribution,
  title={Attribution-based {XAI} methods in computer vision: A review},
  author={Abhishek, Kumar and Kamath, Deeksha},
  journal={arXiv preprint arXiv:2211.14736},
  year={2022}
}

@article{wang2024gradient,
  title={Gradient based feature attribution in explainable {AI}: A technical review},
  author={Wang, Yongjie and Zhang, Tong and Guo, Xu and Shen, Zhiqi},
  journal={arXiv preprint arXiv:2403.10415},
  year={2024}
}

@article{springenberg2014striving,
  title={Striving for simplicity: The all convolutional net},
  author={Springenberg, Jost Tobias and Dosovitskiy, Alexey and Brox, Thomas and Riedmiller, Martin},
  journal={arXiv preprint arXiv:1412.6806},
  year={2014}
}

@inproceedings{zhou2016learning,
  title={Learning deep features for discriminative localization},
  author={Zhou, Bolei and Khosla, Aditya and Lapedriza, Agata and Oliva, Aude and Torralba, Antonio},
  booktitle={CVPR},
  pages={2921--2929},
  year={2016}
}

@article{shrikumar2016not,
  title={Not just a black box: Learning important features through propagating activation differences},
  author={Shrikumar, Avanti and Greenside, Peyton and Shcherbina, Anna and Kundaje, Anshul},
  journal={arXiv preprint arXiv:1605.01713},
  year={2016}
}

@inproceedings{shrikumar2017learning,
  title={Learning important features through propagating activation differences},
  author={Shrikumar, Avanti and Greenside, Peyton and Kundaje, Anshul},
  booktitle={ICML},
  pages={3145--3153},
  year={2017},
}

@inproceedings{selvaraju2017grad,
  title={{Grad-CAM}: Visual explanations from deep networks via gradient-based localization},
  author={Selvaraju, Ramprasaath R and Cogswell, Michael and Das, Abhishek and Vedantam, Ramakrishna and Parikh, Devi and Batra, Dhruv},
  booktitle={ICCV},
  pages={618--626},
  year={2017}
}

@inproceedings{chattopadhay2018grad,
  title={{Grad-CAM++}: Generalized gradient-based visual explanations for deep convolutional networks},
  author={Chattopadhay, Aditya and Sarkar, Anirban and Howlader, Prantik and Balasubramanian, Vineeth N},
  booktitle={WACV},
  pages={839--847},
  year={2018},
}

@inproceedings{binder2016layer,
  title={Layer-wise relevance propagation for neural networks with local renormalization layers},
  author={Binder, Alexander and Montavon, Gr{\'e}goire and Lapuschkin, Sebastian and M{\"u}ller, Klaus-Robert and Samek, Wojciech},
  booktitle={ICANN},
  pages={63--71},
  year={2016},
}

@inproceedings{gu2021interpreting,
  title={Interpreting super-resolution networks with local attribution maps},
  author={Gu, Jinjin and Dong, Chao},
  booktitle={CVPR},
  pages={9199--9208},
  year={2021}
}

@article{picard2021torch,
  title={Torch. manual\_seed (3407) is all you need: On the influence of random seeds in deep learning architectures for computer vision},
  author={Picard, David},
  journal={arXiv preprint arXiv:2109.08203},
  year={2021}
}

@article{xiao2024highly,
  title={Highly efficient and unsupervised framework for moving object detection in satellite videos},
  author={Xiao, Chao and An, Wei and Zhang, Yifan and Su, Zhuo and Li, Miao and Sheng, Weidong and Pietik{\"a}inen, Matti and Liu, Li},
  journal={IEEE TPAMI},
  year={2024},
  publisher={IEEE}
}

@article{huang2023anti,
  title={{Anti-UAV410}: A thermal infrared benchmark and customized scheme for tracking drones in the wild},
  author={Huang, Bo and Li, Jianan and Chen, Junjie and Wang, Gang and Zhao, Jian and Xu, Tingfa},
  journal={IEEE TPAMI},
  volume={46},
  number={5},
  pages={2852--2865},
  year={2023},
  publisher={IEEE}
}

@article{sun2024multi,
  title={{Multi-YOLOv8}: An infrared moving small object detection model based on YOLOv8 for air vehicle},
  author={Sun, Shizun and Mo, Bo and Xu, Junwei and Li, Dawei and Zhao, Jie and Han, Shuo},
  journal={Neurocomputing},
  volume={588},
  pages={127685},
  year={2024},
  publisher={Elsevier}
}

@article{ying2025visible,
  title={Visible-thermal tiny object detection: A benchmark dataset and baselines},
  author={Ying, Xinyi and Xiao, Chao and An, Wei and Li, Ruojing and He, Xu and Li, Boyang and Cao, Xu and Li, Zhaoxu and Wang, Yingqian and Hu, Mingyuan and others},
  journal={IEEE TPAMI},
  year={2025},
  publisher={IEEE}
}

@inproceedings{chollet2017xception,
  title={Xception: Deep learning with depthwise separable convolutions},
  author={Chollet, Fran{\c{c}}ois},
  booktitle={CVPR},
  pages={1251--1258},
  year={2017}
}

@inproceedings{nair2010rectified,
  title={Rectified linear units improve restricted boltzmann machines},
  author={Nair, Vinod and Hinton, Geoffrey E},
  booktitle={ICML},
  pages={807--814},
  year={2010}
}

@article{liu2024infrared,
  title={Infrared small target detection via joint low rankness and local smoothness prior},
  author={Liu, Pei and Peng, Jiangjun and Wang, Hailin and Hong, Danfeng and Cao, Xiangyong},
  journal={IEEE TGRS},
  year={2024},
  publisher={IEEE}
}

@article{yang2024pinwheel,
  title={Pinwheel-shaped convolution and scale-based dynamic loss for infrared small target detection},
  author={Yang, Jiangnan and Liu, Shuangli and Wu, Jingjun and Su, Xinyu and Hai, Nan and Huang, Xueli},
  journal={arXiv preprint arXiv:2412.16986},
  year={2024}
}

@inproceedings{zhang2024explore,
  title={Explore Hybrid Modeling for Moving Infrared Small Target Detection},
  author={Zhang, Mingjin and Liu, Shilong and Ouyang, Yuanjun and Guo, Jie and Tang, Zhihong and Li, Yunsong},
  booktitle={ACM MM},
  pages={6172--6181},
  year={2024}
}

@article{wang2024robust,
  title={A robust space target extraction algorithm based on standardized correlation space construction},
  author={Wang, Han and Chen, Siyang and Shen, Zhihua and Wang, Kunpeng and Duan, Meiya and Yang, Wenbo and Lin, Bin and Zhang, Xiaohu},
  journal={IEEE JSTARS},
  year={2024},
  publisher={IEEE}
}

@article{ying2025infrared,
  title={Infrared small target detection in satellite videos: A new dataset and a novel recurrent feature refinement framework},
  author={Ying, Xinyi and Liu, Li and Lin, Zaipin and Shi, Yangsi and Wang, Yingqian and Li, Ruojing and Cao, Xu and Li, Boyang and Zhou, Shilin and An, Wei},
  journal={IEEE TGRS},
  year={2025},
  publisher={IEEE}
}

@article{fan2024diffusion,
  title={Diffusion-based continuous feature representation for infrared small-dim target detection},
  author={Fan, Linyu and Wang, Yingying and Hu, Guoliang and Li, Feifei and Dong, Yuhang and Zheng, Hui and Ling, Changqing and Huang, Yue and Ding, Xinghao},
  journal={IEEE TGRS},
  year={2024},
  publisher={IEEE}
}

@article{yang2025deep,
  title={Deep learning based infrared small object segmentation: Challenges and future directions},
  author={Yang, Zhengeng and Yu, Hongshan and Zhang, Jianjun and Tang, Qiang and Mian, Ajmal},
  journal={IF},
  volume={118},
  pages={103007},
  year={2025},
  publisher={Elsevier}
}

@inproceedings{zhang2022visible,
  title={Visible-thermal {UAV} tracking: A large-scale benchmark and new baseline},
  author={Zhang, Pengyu and Zhao, Jie and Wang, Dong and Lu, Huchuan and Ruan, Xiang},
  booktitle={CVPR},
  pages={8886--8895},
  year={2022}
}

@inproceedings{zhang2025irmamba,
  title={{IRMamba}: Pixel difference mamba with layer restoration for infrared small target detection},
  author={Zhang, Mingjin and Li, Xiaolong and Gao, Fei and Guo, Jie},
  booktitle={AAAI},
  volume={39},
  number={9},
  pages={10003--10011},
  year={2025}
}

@inproceedings{zhang2025mocid,
  title={{MOCID}: Motion context and displacement information learning for moving infrared small target detection},
  author={Zhang, Mingjin and Ouyang, Yuanjun and Gao, Fei and Guo, Jie and Zhang, Qiming and Zhang, Jing},
  booktitle={AAAI},
  volume={39},
  number={10},
  pages={10022--10030},
  year={2025}
}

@inproceedings{zhang2025semi,
  title={Semi-supervised infrared small target detection with thermodynamic-inspired uneven perturbation and confidence adaptation},
  author={Zhang, Mingjin and Shang, Wenteng and Gao, Fei and Zhang, Qiming and Lu, FengQin and Zhang, Jing},
  booktitle={AAAI},
  volume={39},
  number={10},
  pages={10013--10021},
  year={2025}
}

@article{yuan2024sctransnet,
  title={{SCTransNet}: Spatial-channel cross transformer network for infrared small target detection},
  author={Yuan, Shuai and Qin, Hanlin and Yan, Xiang and Akhtar, Naveed and Mian, Ajmal},
  journal={IEEE TGRS},
  year={2024},
  publisher={IEEE}
}

@inproceedings{liu2024infraredMSHNet,
  title={Infrared small target detection with scale and location sensitivity},
  author={Liu, Qiankun and Liu, Rui and Zheng, Bolun and Wang, Hongkui and Fu, Ying},
  booktitle={CVPR},
  pages={17490--17499},
  year={2024}
}

@inproceedings{chapple1999target,
  title={Target detection in infrared and {SAR} terrain images using a non-Gaussian stochastic model},
  author={Chapple, Philip B and Bertilone, Derek C and Caprari, Robert S and Angeli, Steven and Newsam, Garry N},
  booktitle={Targets and backgrounds: characterization and representation V},
  volume={3699},
  pages={122--132},
  year={1999}
}

@article{kou2023infrared,
  title={Infrared small target segmentation networks: A survey},
  author={Kou, Renke and Wang, Chunping and Peng, Zhenming and Zhao, Zhihe and Chen, Yaohong and Han, Jinhui and Huang, Fuyu and Yu, Ying and Fu, Qiang},
  journal={PR},
  volume={143},
  pages={109788},
  year={2023},
  publisher={Elsevier}
}

@article{hou2024unsupervised,
  title={Unsupervised image sequence registration and enhancement for infrared small target detection},
  author={Hou, Runze and Yan, Puti and Duan, Xuguang and Wang, Xin},
  journal={IEEE TGRS},
  year={2024},
  publisher={IEEE}
}

@article{hua2017progress,
  title={The progress of operational forest fire monitoring with infrared remote sensing},
  author={Hua, Lizhong and Shao, Guofan},
  journal={J. For. Res.},
  volume={28},
  number={2},
  pages={215--229},
  year={2017},
  publisher={Springer}
}

@article{zhao2022single,
  title={Single-frame infrared small-target detection: A survey},
  author={Zhao, Mingjing and Li, Wei and Li, Lu and Hu, Jin and Ma, Pengge and Tao, Ran},
  journal={IEEE GRSM},
  volume={10},
  number={2},
  pages={87--119},
  year={2022},
  publisher={IEEE}
}

@article{wang2022review,
  title={A review of vehicle detection techniques for intelligent vehicles},
  author={Wang, Zhangu and Zhan, Jun and Duan, Chunguang and Guan, Xin and Lu, Pingping and Yang, Kai},
  journal={IEEE TNNLS},
  volume={34},
  number={8},
  pages={3811--3831},
  year={2022},
  publisher={IEEE}
}

@article{li2024mixed,
  title={Mixed-precision network quantization for infrared small target segmentation},
  author={Li, Boyang and Wang, Longguang and Wang, Yingqian and Wu, Tianhao and Lin, Zaiping and Li, Miao and An, Wei and Guo, Yulan},
  journal={IEEE TGRS},
  volume={62},
  pages={1--12},
  year={2024},
  publisher={IEEE}
}

@article{mooney1995point,
  title={Point target detection in consecutive frame staring infrared imagery with evolving cloud clutter},
  author={Mooney, Jonathan M and Silverman, Jerry and Caefer, Charlene E},
  journal={Opt. Eng.},
  volume={34},
  number={9},
  pages={2772--2784},
  year={1995},
  publisher={SPIE}
}

@article{silverman1996temporal,
  title={Temporal filters for tracking weak slow point targets in evolving cloud clutter},
  author={Silverman, Jerry and Mooney, Jonathan M and Caefer, Charlene E},
  journal={Infrared Phys. Technol.},
  volume={37},
  number={6},
  pages={695--710},
  year={1996},
  publisher={Elsevier}
}

@article{tzannes2002detecting,
  title={Detecting small moving objects using temporal hypothesis testing},
  author={Tzannes, Alexis P and Brooks, Dana H},
  journal={IEEE TAES},
  volume={38},
  number={2},
  pages={570--586},
  year={2002},
  publisher={IEEE}
}

@article{niu2024high,
  title={A high framerate imaging-based framework for moving point target detection in very low SNR},
  author={Niu, Wenlong and Guo, Yingyi and Han, Xiaoqing and Ma, Ruidi and Zheng, Wei and Peng, Xiaodong and Yang, Zhen},
  journal={IEEE TAES},
  year={2024},
  publisher={IEEE}
}

@article{yu2021searching,
  title={Searching multi-rate and multi-modal temporal enhanced networks for gesture recognition},
  author={Yu, Zitong and Zhou, Benjia and Wan, Jun and Wang, Pichao and Chen, Haoyu and Liu, Xin and Li, Stan Z and Zhao, Guoying},
  journal={IEEE TIP},
  volume={30},
  pages={5626--5640},
  year={2021},
  publisher={IEEE}
}

@inproceedings{zhang2024irprunedet,
  title={{IRPruneDet}: Efficient infrared small target detection via wavelet structure-regularized soft channel pruning},
  author={Zhang, Mingjin and Yang, Handi and Guo, Jie and Li, Yunsong and Gao, Xinbo and Zhang, Jing},
  booktitle={AAAI},
  volume={38},
  number={7},
  pages={7224--7232},
  year={2024}
}

@inproceedings{zhang2025saist,
  title={{SAIST}: Segment any infrared small target model guided by contrastive language-image pretraining},
  author={Zhang, Mingjin and Li, Xiaolong and Gao, Fei and Guo, Jie and Gao, Xinbo and Zhang, Jing},
  booktitle={CVPR},
  pages={9549--9558},
  year={2025}
}

@article{zhang2023dim2clear,
  title={{Dim2Clear} network for infrared small target detection},
  author={Zhang, Mingjin and Zhang, Rui and Zhang, Jing and Guo, Jie and Li, Yunsong and Gao, Xinbo},
  journal={IEEE TGRS},
  year={2023},
  volume={61},
  pages={1-14},
  publisher={IEEE},
  doi={10.1109/TGRS.2023.3263848}
}

@article{li2025dataset,
  title={Infrared video satellite aerial moving target detection dataset},
  author={Li, Ruojing and Zeng, Yaoyuan and Sheng, Weidong and Li, Boyang and Li, Zhaoxu and Chen, Nuo and Guo, Gaowei and Dou, Zechao and Long, Zhengxing and Luo, Yihang and Li, Zhijun and Li, Miao and Long, Yunli and An, Wei},
  doi={10.57760/sciencedb.j00240.00077},
  url={https://doi.org/10.57760/sciencedb.j00240.00077},
  year={2025},
  publisher={Science Data Bank}
}

@article{li2026satvideodataset,
  title={Infrared video satellite aerial moving target detection dataset and its evaluation},
  author={Li, Ruojing and Li, Zhaoxu and Chen, Nuo and Guo, Gaowei and Dou, Zechao and Long, Zhengxing and Luo, Yihang and Zeng, Yaoyuan and Sheng, Weidong and Li, Boyang and others},
  doi={10.11834/jig.250536},
  journal={Journal of Image and Graphics},
  pages={1--15},
  year={2026}
}

@article{deng2026learning,
  title={Learning global dynamic query for large--motion infrared small target detection},
  author={Deng, Chuiyi and Guo, Yanyin and Xu, Xiang and Zhao, Zhuoyi and Xia, Yixin and An, Runxuan and Li, Junwei and Plaza, Antonio},
  journal={IEEE TGRS},
  year={2026},
  volume={64},
  pages={1-16},
  publisher={IEEE}
}

@article{li2026dynamic,
  title={Dynamic high-frequency convolution for infrared small target detection},
  author={Li, Ruojing and Xiao, Chao and Yin, Qian and An, Wei and Chen, Nuo and Ying, Xinyi and Li, Miao and Wang, Yingqian},
  journal={IEEE TCSVT},
  year={2026},
  volume={},
  number={},
  pages={1-1}
}

@article{li2025ilnet,
  title={{ILNet}: Low-level matters for salient infrared small target detection},
  author={Li, Haoqing and Yang, Jinfu and Wang, Runshi and Xu, Yifei},
  journal={IEEE TAES},
  year={2025},
  volume={61},
  number={4},
  pages={8306-8318},
  publisher={IEEE}
}

\begin{IEEEbiography}[{\includegraphics[width=1in,height=1.25in,clip,keepaspectratio]{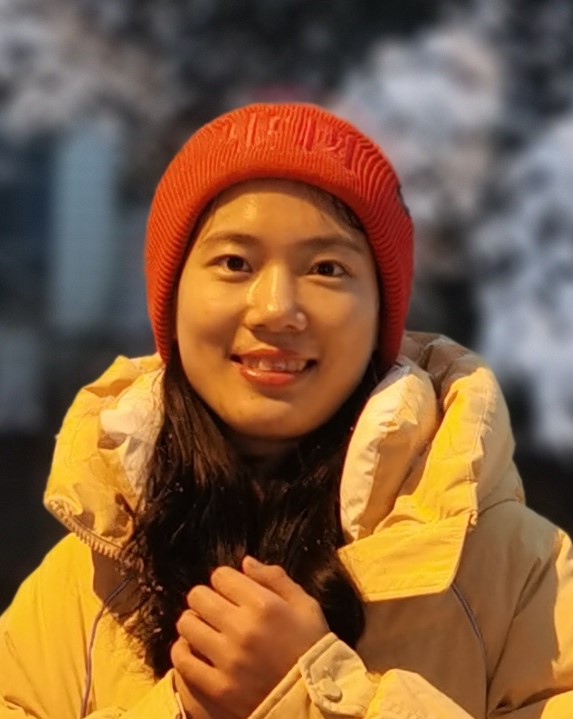}}]{Ruojing Li}
received the B.E. degree in electronic engineering from the National University of Defense Technology (NUDT), Changsha, China, in 2020. She is currently working toward the Ph.D. degree in information and communication engineering from NUDT. Her research interests include infrared small target detection, particularly on multi-frame detection, deep learning. For more information, please visit \href{https://tinalrj.github.io/}{https://tinalrj.github.io/}.
\end{IEEEbiography}

\begin{IEEEbiography}[{\includegraphics[width=1in,height=1.25in,clip,keepaspectratio]{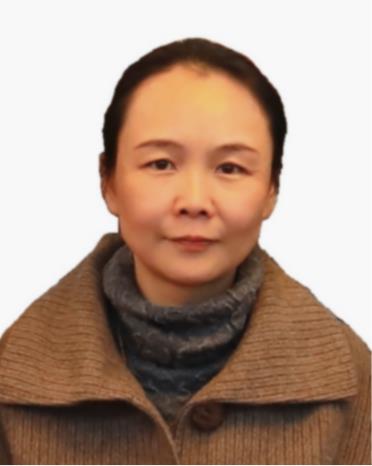}}]{Wei An}
received the Ph.D. degree from the National University of Defense Technology (NUDT), Changsha, China, in 1999. She was a Senior Visiting Scholar with the University of Southampton, Southampton, U.K., in 2016. She is currently a Professor with the College of Electronic Science and Technology, NUDT. She has authored or co-authored over 100 journal and conference publications. Her current research interests include signal processing and image processing.
\end{IEEEbiography}

\begin{IEEEbiography}[{\includegraphics[width=1in,height=1.25in,clip,keepaspectratio]{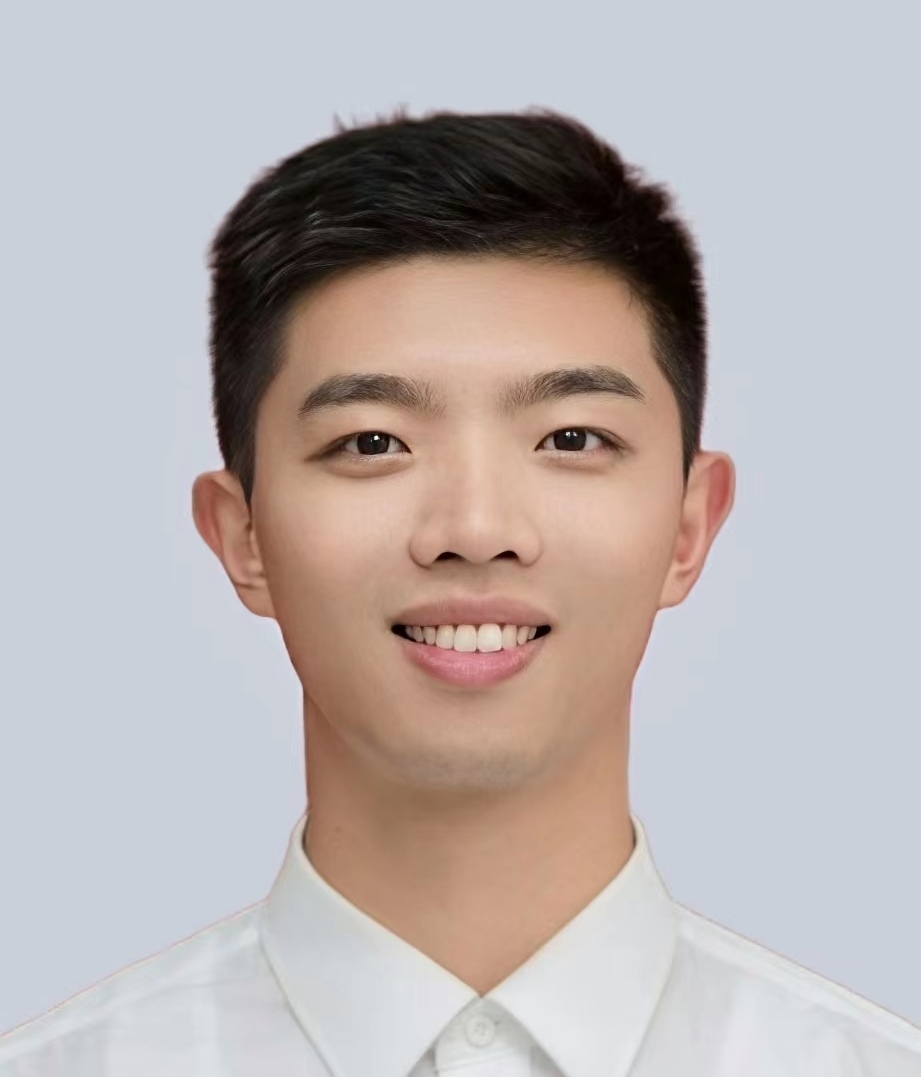}}]{Yingqian Wang}
received his B.E. degree in electrical engineering from Shandong University, Jinan, China, in 2016, the Master and the Ph.D. degrees in information and communication engineering from National University of Defense Technology (NUDT), Changsha, China, in 2018 and 2023, respectively. Dr. Wang is currently an Associate Professor with the College of Electronic Science and Technology, NUDT. His research interests focus on computational photography and low-level vision, particularly on light field image processing and image super-resolution. For more information, please visit \href{https://yingqianwang.github.io/}{https://yingqianwang.github.io/}.
\end{IEEEbiography}

\begin{IEEEbiography}[{\includegraphics[width=1in,height=1.25in,clip,keepaspectratio]{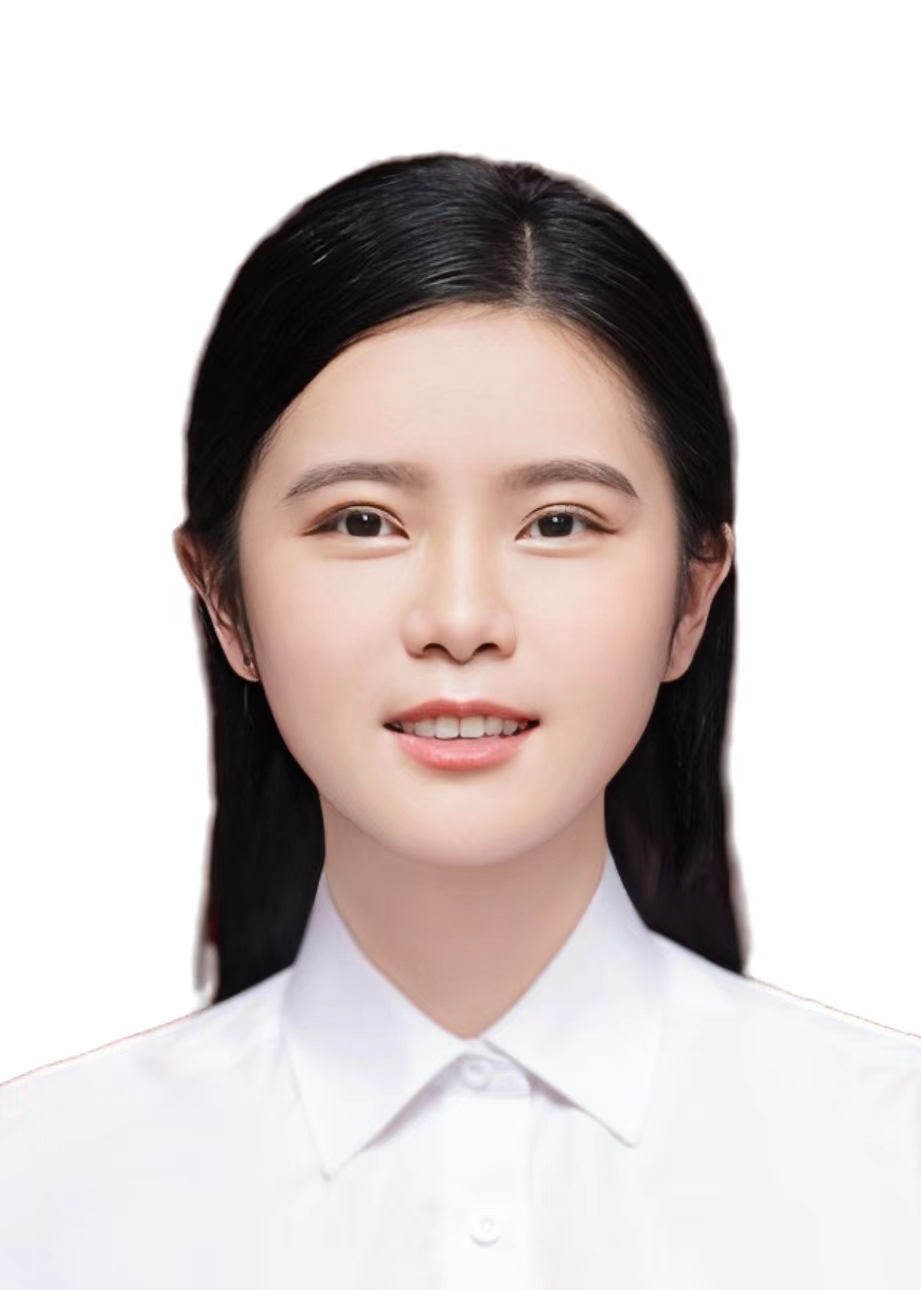}}]{Xinyi Ying}
received the Ph.D. degree in information and communication engineering from the National University of Defense Technology (NUDT), Changsha, China, in 2024. She is currently a Lecturer with the College of Electronic Science and Technology, NUDT. Her research interests focus on space-based optical surveillance, and detection and tracking of infrared small targets.
\end{IEEEbiography}

\begin{IEEEbiography}[{\includegraphics[width=1in,height=1.25in,clip,keepaspectratio]{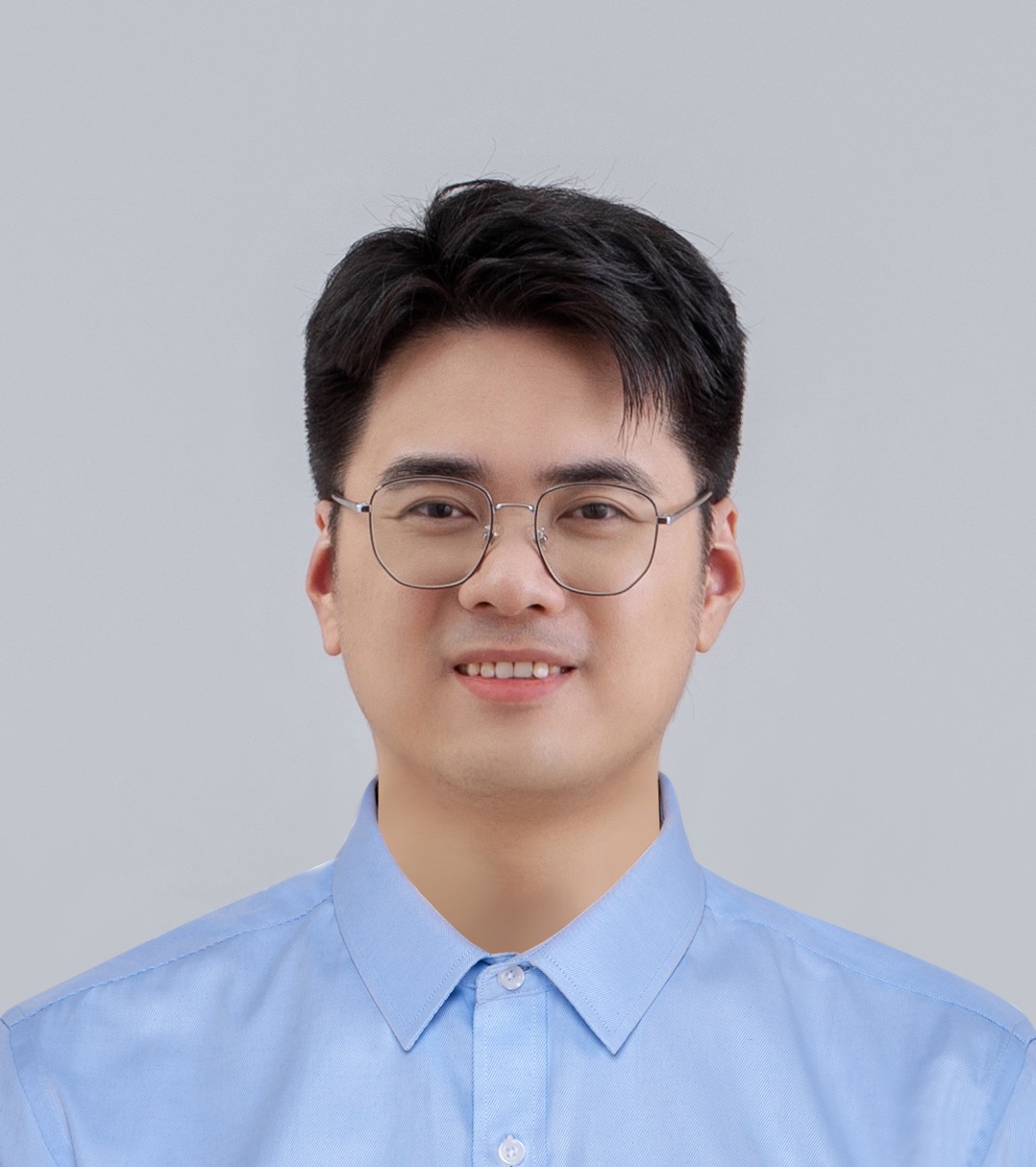}}]{Yimian Dai}
(Member, IEEE) received the B.E. degree in information engineering and the Ph.D. degree in signal and information processing from Nanjing University of Aeronautics and Astronautics, Nanjing, China, in 2013 and 2020, respectively.
From 2021 to 2024, he was a Postdoctoral Researcher with the School of Computer Science and Engineering, Nanjing University of Science and Technology, Nanjing, China.
He is currently an Associate Professor with the College of Computer Science, Nankai University, Tianjin, China.
His research interests include computer vision, deep learning, and their applications in remote sensing.
For more information, please visit the link (\href{https://yimian.grokcv.ai/}{https://yimian.grokcv.ai/}).
\end{IEEEbiography}

\begin{IEEEbiography}[{\includegraphics[width=1in,height=1.25in,clip,keepaspectratio]{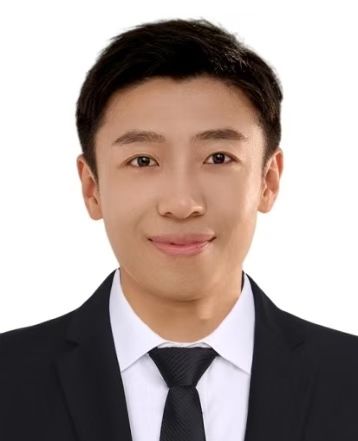}}]{Longguang Wang}
received the B.E. degree in Electrical Engineering from Shandong University (SDU), Jinan, China, in 2015, and the Ph.D. degree in Information and Communication Engineering from National University of Defense Technology (NUDT), Changsha, China, in 2022. His current research interests include low-level vision and 3D vision.
\end{IEEEbiography}

\begin{IEEEbiography}[{\includegraphics[width=1in,height=1.25in,clip,keepaspectratio]{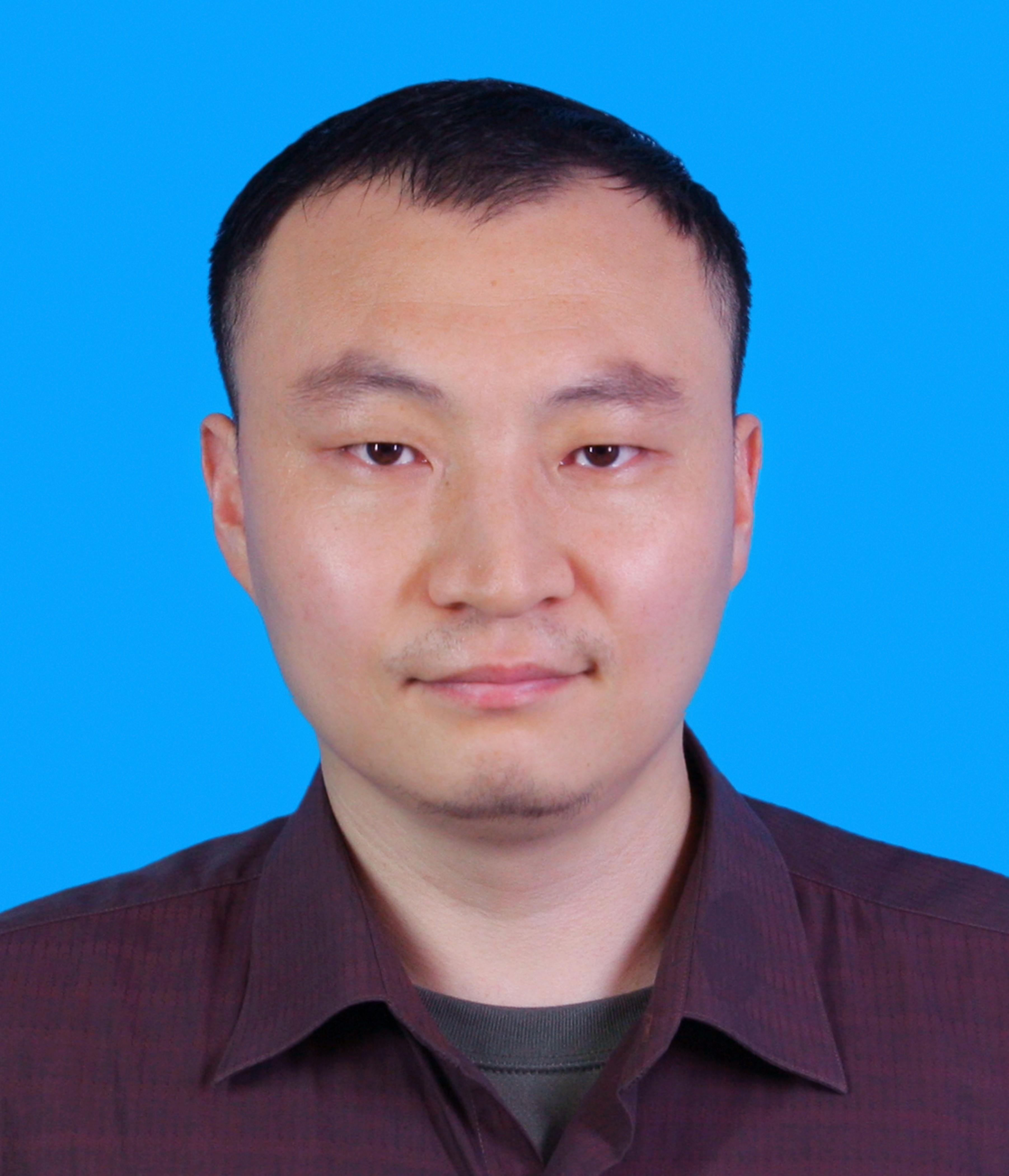}}]{Miao Li}
received the M.E. and Ph.D. degrees from the National University of Defense Technology (NUDT) in 2012 and 2017, respectively. He is currently an Associate Professor with the College of Electronic Science and Technology, NUDT. His current research interests include infrared dim and small target detection and event detection.
\end{IEEEbiography}

\begin{IEEEbiography}[{\includegraphics[width=1in,height=1.25in,clip,keepaspectratio]{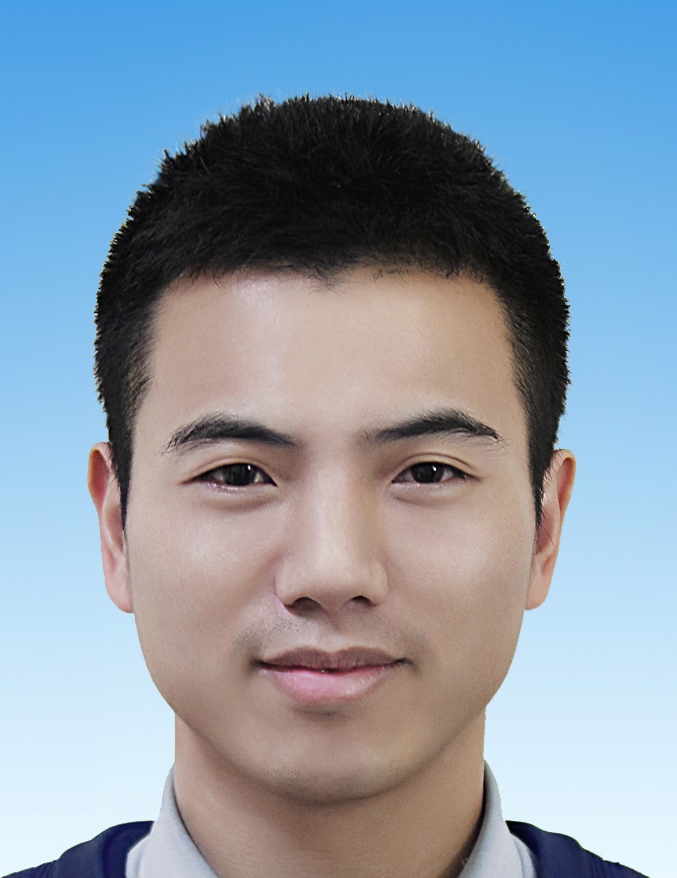}}]{Yulan Guo}
is a full Professor with the School of Electronics and Communication Engineering, Sun Yat-sen University. His research interests lie in computer vision and robotics, particularly in 3D reconstruction, point cloud understanding, and robot systems. He has authored over 200 articles at highly referred journals and conferences. He served as a Senior Area Editor for IEEE Transactions on Image Processing, and an Associate Editor for the Visual Computer, and Computers \& Graphics. He also served as an area chair for CVPR 2025/2023/2021, ICCV 2025/2021, ECCV 2024, NeurIPS 2024, and ACM Multimedia 2021. He organized over 10 workshops, challenges, and tutorials in prestigious conferences such as CVPR, ICCV, ECCV, and 3DV. He is a Senior Member of IEEE and ACM.
\end{IEEEbiography}

\begin{IEEEbiography}[{\includegraphics[width=1in,height=1.25in,clip,keepaspectratio]{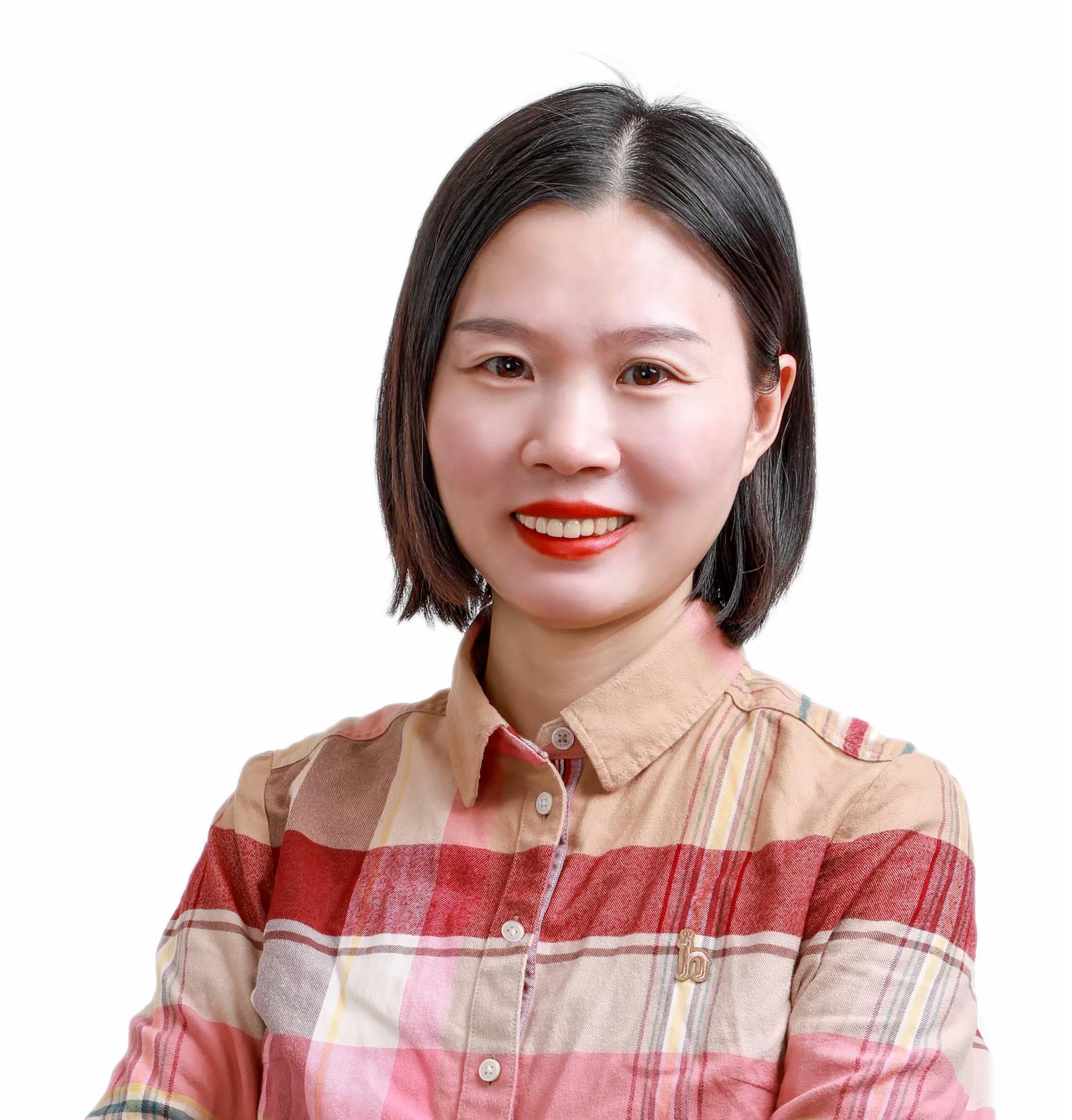}}]{Li Liu}
received the PhD degree in information and communication engineering from the National University of Defense Technology (NUDT), China, in 2012. She is now a full professor with the College of Electronic Science and Technology, NUDT. She has hold research visits at the University of Waterloo, Canada, at the Multimedia Laboratory with the Chinese University of Hong Kong, and the University of Oulu, Finland. She was a cochair of nine International Workshops with CVPR, ICCV, and ECCV. She served as the leading guest editor for special issues in IEEE Transactions on Pattern Analysis and Machine Intelligence (IEEE TPAMI) and International Journal of Computer Vision. Her current research interests include computer
vision, pattern recognition and remote image analysis. Her papers have currently more than 18400 citations according to Google Scholar. She currently serves as associate editor for IEEE Transactions on Geoscience and Remote Sensing
(IEEE TGRS), IEEE Transactions on Circuits and Systems for Video Technology (IEEE TCSVT), and IEEE TMM.
\end{IEEEbiography}

\vfill

\end{document}


%
\title{Appendices of ``Probing Deep into Temporal Profile Makes the Infrared Small Target Detector Much Better''}

\author{Ruojing~Li, Wei~An, Yingqian~Wang, Xinyi~Ying, Yimian~Dai, Longguang~Wang, Miao~Li, Yulan~Guo, Li~Liu 
}

\markboth{Journal of \LaTeX\ Class Files,~Vol.~14, No.~8, August~2015}%
{Li \MakeLowercase{\textit{et al.}}: Probing Deep into Temporal Profile Makes the Infrared Small Target Detector Much Better}


\maketitle

\IEEEdisplaynontitleabstractindextext
\IEEEpeerreviewmaketitle

\section{Correlation and sampling rate}

\begin{figure}[!t]
    \centering
    \includegraphics[width=1\linewidth]{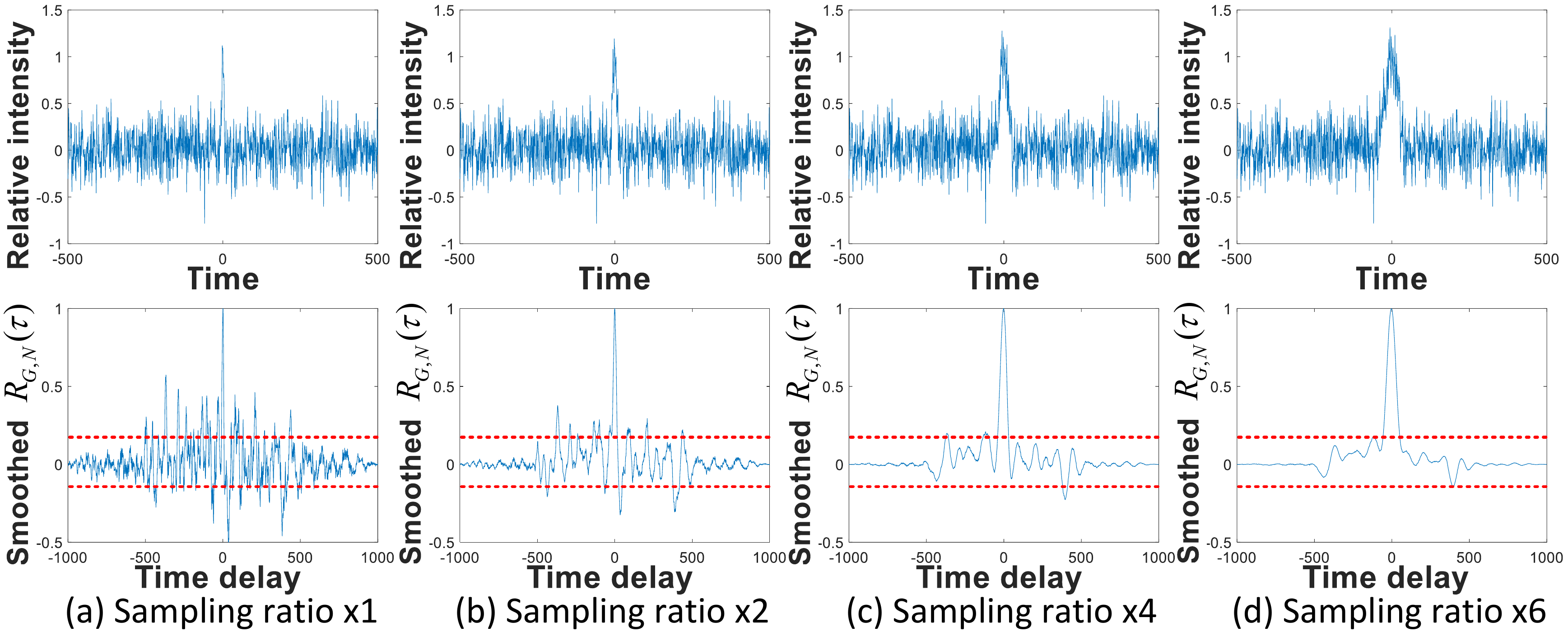}
    \caption{Visualizations of target-in-noise signals and their correlation functions under different sampling rates. From (a) to (d), as the sampling rate increases, the correlation function exhibits reduced fluctuations at non-zero time delays.}
    \label{fig:SampleRatio-Cor}
\end{figure}
Ideally, target and noise signals are uncorrelated \cite{2016InfraredTargetDetection}. It means that the correlation function $R_{G,N}(\tau)$ (Equation 1 in the main text) should equal zero at $\left| \tau \right| > \frac{1}{2}\tau_t$, where $\tau_t$ donates the target duration. However, the practical $R_{G,N}(\tau)$ often exhibits fluctuations at $\left| \tau \right| > \frac{1}{2}\tau_t$ due to signal discretization from sampling, causing deviations from ideal results. To verify that, we calculated $R_{G,N}(\tau)$ for the same target-in-noise signal at varying sampling rates, as shown in Fig. \ref{fig:SampleRatio-Cor}. The analyses show higher sampling rates reduce fluctuation amplitudes at $\left| \tau \right| > \frac{1}{2}\tau_t$, indicating closer approximation to the ideal state. Therefore, these fluctuations do not affect qualitative judgments of the correlation between signal components. This conclusion can apply to target-in-clutter signals, though their fluctuations may additionally depend on inherent signal characteristics.

\section{Noise modeling}
In IRST images, there are mainly three signals from different sources: target $T$, background $B$, and noise $N$ signals. Different signals are usually not correlated with each other, so infrared images $I$ can be represented as $I=T+B+N$.

Noise signals can be divided into random noises and fixed pattern noises (FPNs) according to their spatial-temporal characteristics. Salt and pepper noise is a typical FPN, but it is not considered. That is because, it can be effectively identified and removed without affecting the surrounding normal pixels. According to the sources of signals, random noises include photon noise (i.e., shot noise), dark current noise, readout noise, and noise from subsequent processing circuits \cite{2016InfraredTargetDetection}. Photon noise and dark current noise follow Poisson distributions, while others follow Gaussian distributions. According to the central limit theorem, the combination of these noises follows a Gaussian distribution \cite{wang2015dsNoise}, and the distribution can be represented as
\begin{equation}
    \label{randomN}
    p(n) = \frac{1}{\sqrt{2 \pi \sigma_n}} exp^{-\frac{(n-\mu_n)^2}{2 \sigma_n^2}},
\end{equation}
where $n$ denotes the gray value of noise. $\sigma_n$ is the standard deviation of the noise distribution, and $\mu_n$ is the mathematical expectation of $n$ ($\mu_n=0$).

Besides, the non-uniformity of images is also an important factor affecting image quality \cite{wang2022noise}. The non-uniformity is manifested mainly as the response (i.e., gray value) difference of the same radiation source in an image. The non-uniformity can be expressed as
\begin{equation}
    \label{nonUN}
    y_{ij} = g_{ij} \cdot x_{ij} + o_{ij},
\end{equation}
where $i$ and $j$ denote the row and column coordinates of the pixel. $x_{ij}$ is the ideal response of the pixel (without non-uniformity), and $y_{ij}$ is the actual response. $g_{ij}$ and $o_{ij}$ represent the gain and the offset of the detection unit corresponding to the pixel.

The non-uniformity is constant over a short period of time (e.g., an hour), and thus it can be regarded as an FPN. It is assumed that the multiplicative noise $g_{ij}$ follows a uniform distribution of $(1-\sigma_g, 1+\sigma_g)$ and the additive noise $o_{ij}$ follows a Gaussian distribution with a mean of 0 and a standard deviation of $\sigma_o$.

Therefore, the overall noise of a pixel can be described as
\begin{equation}
    \label{overallN}
    N_{ij} = (g_{ij}-1)x_{ij} + o_{ij} + n.
\end{equation}

\section{Dataset summary}
\begin{figure}[!t]
    \centering
    \includegraphics[width=1\linewidth]{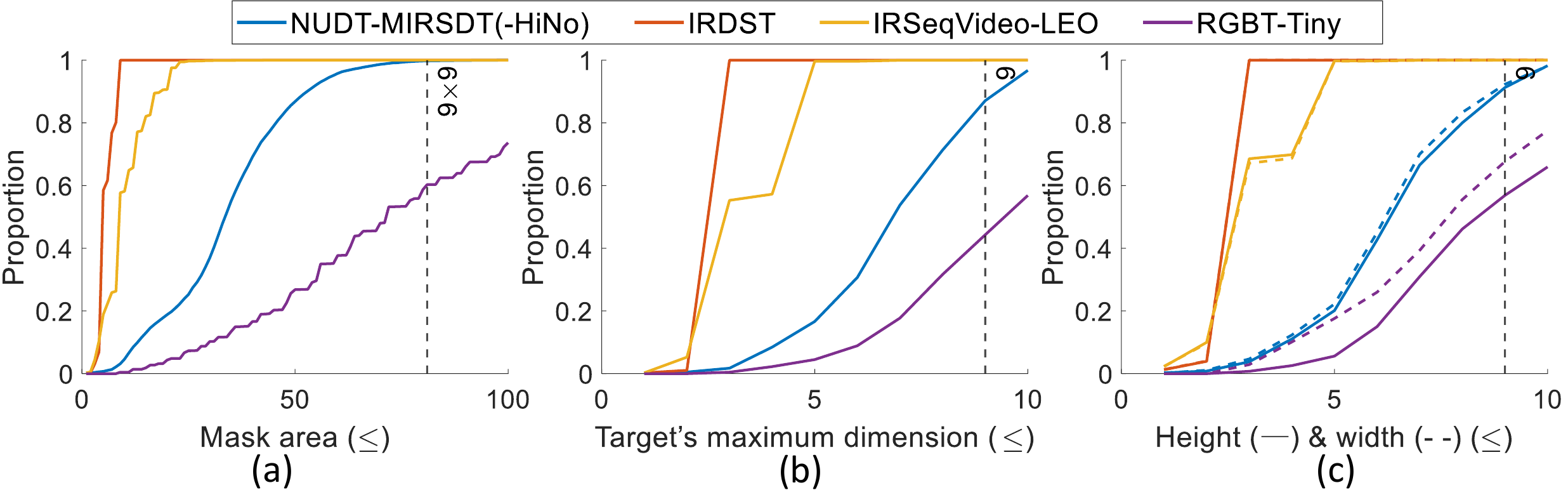}
    \caption{Distribution curves of statistical characteristics of the used datasets. (a) Target size distribution. (b) Target’s maximum dimension distribution. (c) Target height and width distribution. The NUDT-MIRSDT and the NUDT-MIRSDT-HiNo datasets share identical ground truths, thus their results are represented by the same curves, i.e., NUDT-MIRSDT(-HiNo).}
    \label{fig:dataset_summary}
\end{figure}

In this section, we summary the distribution characteristics of the utilized datasets, including the target size\footnote{Target size distribution: The cumulative distribution of target mask area, representing the proportion of targets with a mask area below a given threshold.}, the maximum dimension of the target\footnote{Target’s maximum dimension distribution: The cumulative distribution of target bounding box major axis length, indicating the proportion of targets whose longer sides are below a given threshold.}, and target height/width distributions\footnote{Target height and width distribution: The cumulative distribution of bounding box dimensions, describing the proportion of targets with a height or width below a given threshold.}, as illustrated in Fig. \ref{fig:dataset_summary}. Additionally, We analyze the overall statistical properties such as mean target height, mean target width, mean mask area, and average SNR, as presented in Table \ref{tab:DataSum}. The target SNR (peak signal-to-noise ratio) is defined as
\begin{equation}
    \label{snr}
    SNR = \frac{|m_t-\mu_b|}{\sigma_b},
\end{equation}
where $m_t$ is the maximum grayscale value of a target. $\mu_b$ and $\sigma_b$ denote the mean and the standard deviation of the grayscale values in the annular background region (ranging from $9\times 9$ to $15\times 15$) centered on the target.

\begin{table}[t!]
\caption{
Overall statistical properties of the used datasets, including mean target height, mean target width, mean mask area, and average SNR.}\label{tab:DataSum}
\centering
\begin{threeparttable}
\renewcommand{\arraystretch}{1.1}
\begin{tabular}{c c c c c}
\hline
Datasets & Height & Weight & Mask area & SNR \\ \hline
NUDT-MIRSDT \cite{li2023direction} & 6.86 & 6.72 & 34.37 & 11.67 \\
NUDT-MIRSDT-HiNo & 6.86 & 6.72 & 34.37 & 3.86 \\
IRSDT-simulation \cite{sun2023receptive} & 2.95 & 2.95 & 6.12 & 24.95 \\
IRSatVideo-LEO \cite{ying2025infrared} & 3.50 & 3.53 & 10.73 & 11.91 \\
RGBT-Tiny (parts) \cite{ying2025visible} & 9.72 & 9.24 & 95.10\tnote{*} & 19.64 \\
\hline
\end{tabular}
\begin{tablenotes}
\footnotesize
    \item[*] The RGBT-Tiny dataset only has bounding box annotations. The mean mask area of targets in RGBT-Tiny dataset is smaller than 95.10.
\end{tablenotes}
\end{threeparttable}
\end{table}

From Fig. \ref{fig:dataset_summary} and Table \ref{tab:DataSum}, we can observe that almost all targets in the NUDT-MIRSDT dataset and the NUDT-MIRSDT-HiNo dataset are smaller than $9\times 9$ pixels, and all targets in the IRSDT-simulation dataset and the IRSatVideo-LEO dataset are smaller than $5\times 5$ pixels. The average SNR values of the used datasets range from 3 to 25. There is also the dedicated test set with $SNR \leq 3$ and the dataset with high-intensity noise.

Therefore, these datasets are representative, encompassing both simulated and measured data. The targets range from tiny ones with only a few pixels to those with dozens of pixels, while the datasets cover various scenarios from extremely low SNR conditions to normal SNR conditions.

\section{SCorMs visualization}

\begin{figure}[!t]
    \centering
    \includegraphics[width=1\linewidth]{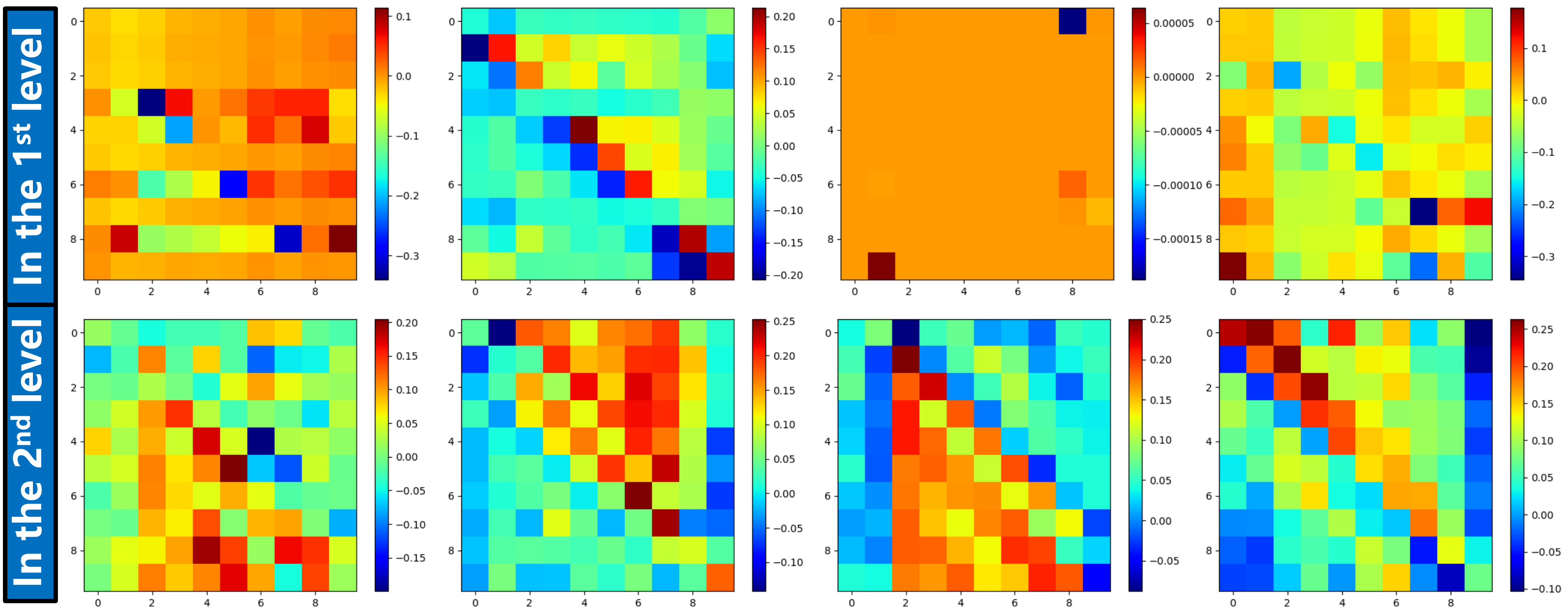}
    \caption{Visualization of the statistic-based correlation matrixes in the first and the second levels with length 10. SCorMs are always irregular in different levels.)
    }
    \label{fig:visual_TP_L10}
\end{figure}

\begin{figure}[!t]
    \centering
    \includegraphics[width=1\linewidth]{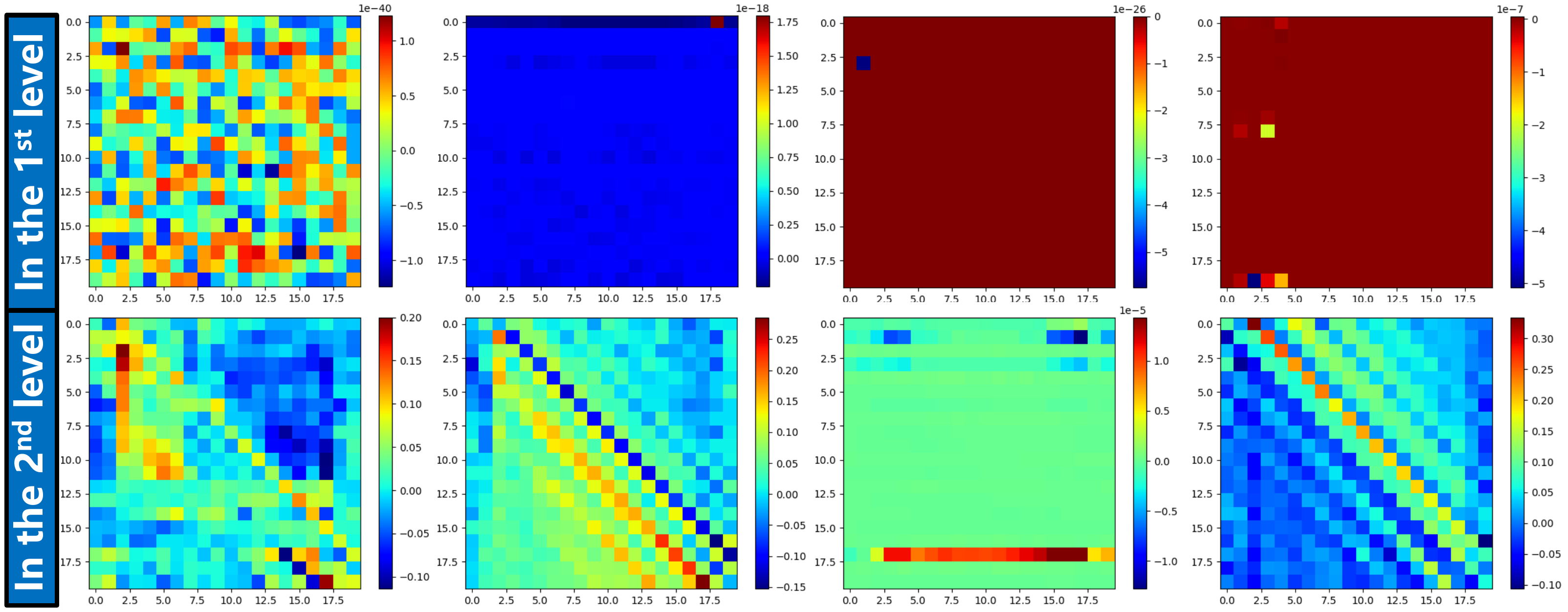}
    \caption{Visualization of the statistic-based correlation matrixes in the first and the second levels with length 20. SCorMs are approximately symmetric about the minor diagonals in the second level. This explains why our DeepPro with $T=20$ demonstrates acceptable performance as compared to network with $T=40$, and even performs much better than STDMANet.)
    }
    \label{fig:visual_TP_L20}
\end{figure}

In the main text, when the length of the temporal profile information is 10, the performance of the network is significantly degraded. When the length is 20, the performance is still acceptable. To explain why, we visualized all SCorMs with lengths of 10 and 20 in the first and the second levels, as shown in Figs. \ref{fig:visual_TP_L10} and \ref{fig:visual_TP_L20}.

The SCorMs with length 20 are approximately symmetric about the minor diagonals in the second level. It reflects that DeepPro can learn some consistent essential features or statistical properties from the temporal profile information of length 20. Therefore, DeepPro with $T=20$ achieves acceptable performance. However, the network can learn little essential characteristics from the temporal profile information of length 10, as the SCorMs are always irregular in different levels.

\section{DeepPro-Plus}
\subsection{Network design}

\begin{figure*}[!t]
    \centering
    \includegraphics[width=1\linewidth]{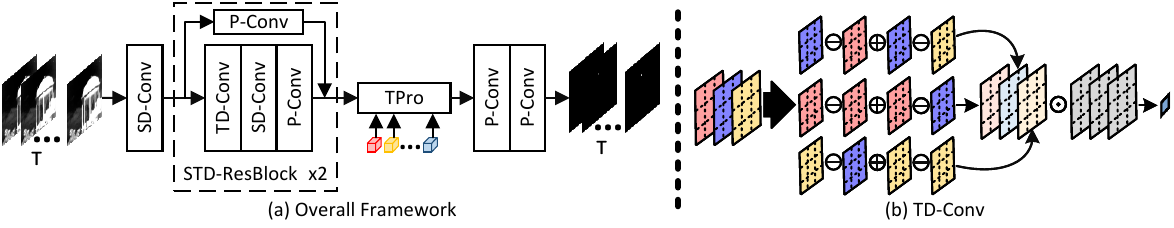}
    \caption{The framework of our DeepPro-Plus. Different from DeepPro which only calculates in the time dimension, this version incorporates few spatial computations to acquire little spatial information as supplementation.
    }
    \label{fig:architectureV2}
\end{figure*}

The structure of DeepPro-Plus is shown in Fig. \ref{fig:architectureV2}. There is only one level in DeepPro-Plus, since the original multi-level design is for rich information from temporal profiles of IRSDTs which can also be obtained by supplementing spatial calculations. Before TPro, there is still one convolution layer and two same blocks for spatial-temporal information extraction. Differently, all T-Convs change to spatial difference convolutions (SD-Conv) $\rho(\cdot)$ \cite{su2021pixel}, and all TD-Convs extend their kernels in the space domain.

Specifically, SD-Conv can help to maintain and emphasize the spatial saliency of dim targets by applying convolution operation to center pixel difference feature maps. The process can be described as
\begin{equation}
    \label{equ_SD}
    \begin{array}{rl}
    \rho(\cdot, \mathbf{W_{SD}}) & = \mathbf{x_c}\cdot \mathbf{w_c^{\prime}} + \sum\limits_{i=1, i\not=c}^{k\times k}\mathbf{w_i^{\prime}} \cdot (\mathbf{x_c}-\mathbf{x_i}) \\
     & = \mathbf{x_c}\cdot \sum\limits_{i=1}^{k\times k}\mathbf{w_i^{\prime}} - \sum\limits_{i=1, i\not=c}^{k\times k}\mathbf{w_i^{\prime}} \cdot \mathbf{x_i},
    \end{array}
\end{equation}
where $c=(k\times k+1)/2$ denotes the index of the spatial center. $\mathbf{W_{SD}}\in\mathbb{R}^{l\times k\times k}$ represents the convolution kernel, and $\mathbf{w_i^{\prime}}\in\mathbb{R}^{l\times 1\times 1}$ is the $i^{th}$ weight vector in it. $\mathbf{x_i}\in\mathbb{R}^{l\times 1\times 1}$ is the $i^{th}$ ($i=1,2,...,k\times k$) pixel vector in sliding window. In the first SD-Conv, $l=5, k=7$. In others, $l=3, k=3$ with a spatial dilation of 2.

The kernel size of TD-Conv is changed from $3\times 1\times 1$ to $3\times 3\times 3$. Then, the difference calculation between pixels in different frames is turned to that between pixel matrix in different frames. The process can be described as
\begin{equation}
    \label{equ_TD2}
    \begin{array}{rl}
    \psi(\cdot, \mathbf{W_{TD}}) & =\sum\limits_{t=1}^{l}\mathbf{w_t} \cdot (2\cdot \mathbf{s_t} - \mathbf{s_{t+1}} - \mathbf{s_{t-1}}) \\
     & = \sum\limits_{t=1}^{l}(2\cdot \mathbf{w_t} - \mathbf{w_{t+1}} - \mathbf{w_{t-1}}) \cdot \mathbf{s_t},
    \end{array}
\end{equation}
where $\mathbf{s_t}\in\mathbb{R}^{1\times k\times k}$ is the pixel matrix from the $t^{th}$ ($t=1,2,...,l$) frame in sliding window. $\mathbf{w_t}\in\mathbb{R}^{1\times k\times k}$ is the $t^{th}$ weight matrix of $\mathbf{W_{TD}}$.

The input of DeepPro-Plus was also set to 40 consecutive frames ($T=40$), but the number of temporal probes was set to 8 ($m=8$) since only one level here.

\subsection{More comparisons to the state-of-the-arts}
For a more comprehensive evaluation, we conducted additional experiments comparing DeepPro-Plus with the state-of-the-art methods (from Section 4.2 in the main text) on the IRSatVideo-LEO dataset \cite{ying2025infrared}. Specifically, the IRSatVideo-LEO dataset is a large-scale, high-fidelity simulation derived from Low Earth Orbit satellite motion dynamics. It comprises 200 video sequences characterized by slow background drift and tiny targets with an average size of 10.73 pixels. In this dataset, target motion velocities reach up to 20 pixels per frame, exceeding the $19\times 19$ spatial receptive field of DeepPro-Plus. To address this mismatch, we applied a $2\times$ downsampling strategy during preprocessing for both training and inference. Experimental results on this dataset are presented in Table \ref{tab:SOTA_LEO}. 

\begin{table}[t!]
\caption{Detection results achieved by different multi-frame state-of-the-art methods on the IRSatVideo-LEO dataset.}\label{tab:SOTA_LEO}
\centering
\renewcommand{\arraystretch}{1.1}
\begin{tabular}{c c c c}
\hline
Methods & $P_d$  & $F_a$ & AUC \\ \hline  
Res-U+DTUM \cite{li2023direction} & 87.00 & 1.59 & 0.9579 \\  
STDMANet \cite{yan2023stdmanet} & 93.92 & 1.38 & 0.9907 \\  
Res-U+RFR \cite{ying2025infrared} & 91.70 & 2.07 & 0.9755 \\  
DQAligner \cite{deng2026learning} & 60.66 & 0.97 & 0.9088 \\  
DeepPro-Plus & 96.37 & 1.00 & 0.9925 \\ \hline  
\end{tabular}
\end{table}


From Table \ref{tab:SOTA_LEO}, it can be observed that DeepPro-Plus achieves superior detection performance, significantly improving the detection rate while maintaining a low false alarm rate. This underscores our method's exceptional adaptability and robustness in diverse complex scenarios and against tiny-dim targets. Specifically, DeepPro-Plus exhibits high sensitivity in capturing and identifying tiny targets amidst dynamic backgrounds. These findings further validate the critical importance of local temporal profile features for IRST detection.

\subsection{Ablation study}

\subsubsection{Length of temporal profile information}
Since the integration of spatial computations, it is necessary to further explore whether it alters the demand of the network for the length of temporal profile information. Therefore, we modified the input frame length (as well as the size of SCorM) and observed the resulting shifts in experimental results. The results are summarized in Table \ref{tab:lenTP2}, which are consistent with the results of DeepPro. 40 length may be the most cost-effective option for this framework, and longer information is not cost-effective without significant performance improvement but with larger computation.

Notably, we compared the results of other state-of-the-art methods in Table 2 in the main text and the results of the variant with $T=10$ in Table \ref{tab:lenTP2}. It can be found that the detection performance of DeepPro-Plus probing only ten frames is still significantly superior to other state-of-the-art methods, especially STDMANet \cite{yan2023stdmanet} which uses 20 frames for the detection. Meanwhile, the parameters are much smaller and the computational efficiency is much higher than others. This also illustrates that \textit{compared to a large spatial-temporal receptive field and fine feature extraction, mining temporal profile information is much more important for IRST detection}, especially for targets with low SNR.

\begin{table}[t]
\centering
    \caption{Detection results achieved by our DeepPro-Plus with different lengths of input frames on the NUDT-MIRSDT-HiNo dataset. The best results are in bold. Notably, GFLOPs is expressed in units of GFLOPs per sample.}
    \label{tab:lenTP2}
    \renewcommand{\arraystretch}{1.1}
    \begin{tabular}{c c c c c c c}
    \hline
    $T$ & $P_d$ & $F_a$ & AUC & \#Params (M) & GFLOPs & FPS \\
    \hline
    10 & 55.06 & 1.52 & 0.8401 & \textbf{0.235} & \textbf{3.83} & 227.76 \\
    20 & 71.66 & 1.55 & 0.8947 & 0.245 & 3.85 & 220.61 \\
    40 & \textbf{76.23} & 1.69 & \textbf{0.9171} & 0.284 & 3.89 & 224.05 \\
    60 & 69.69 & 1.66 & 0.9162 & 0.348 & 3.93 & 220.10 \\
    80 & 68.58 & 1.25 & 0.8914 & 0.439 & 3.98 & 202.06 \\
    100 & 68.83 & \textbf{1.19} & 0.8960 & 0.554 & 4.02 & 201.71 \\
    \hline
    \end{tabular}
\end{table}

\begin{table}[t]
    \caption{Detection results achieved by our DeepPro-Plus with different numbers of SCorMs on the NUDT-MIRSDT-HiNo dataset. The best results are in bold.} \label{tab:numTP2}
    \centering
    \renewcommand{\arraystretch}{1.1}
    \begin{tabular}{c c c c c}
    \hline
    $m$ & $P_d$ & $F_a$ & \#Params (M) & FPS \\
    \hline
    0 & - & - & 0.231 & - \\
    1 & 72.35 & 2.06 & \textbf{0.238} & 219.22 \\
    2 & 75.19 & 2.42 & 0.244 & 220.75 \\
    4 & 74.38 & 2.38 & 0.257 & 218.99 \\
    8 & \textbf{76.23} & \textbf{1.69} & 0.284 & 224.05 \\
    \hline
    \end{tabular}
\end{table}

\subsubsection{Number of SCorMs}
Originally, the network is built with three levels, and each level includes four SCorMs (totally 12 SCorMs), to extract various temporal profile information. With the architecture now reduced to a single level, only four SCorMs might not suffice to capture comprehensive temporal profile correlation features. Consequently, we adjusted the number of SCorMs in this new configuration (i.e., DeepPro-Plus) to investigate the demand for SCorMs in the single-level setup, and the results are shown in Table \ref{tab:numTP2}. The variant with 8 SCorMs achieves the best detection performance in terms of $P_d$ and $F_a$. That is because, the structure with fewer levels needs more SCorMs in each level to maintain rich temporal profile information. In addition, the parameters of one SCorM are only 6.56K, and the parameters of other structures in DeepPro-Plus are only 0.231M. Multiple temporal probes cannot increase computational costs but detection performance.

\subsubsection{Difference convolutions}
\begin{figure*}[!t]
    \centering
    \includegraphics[width=1\linewidth]{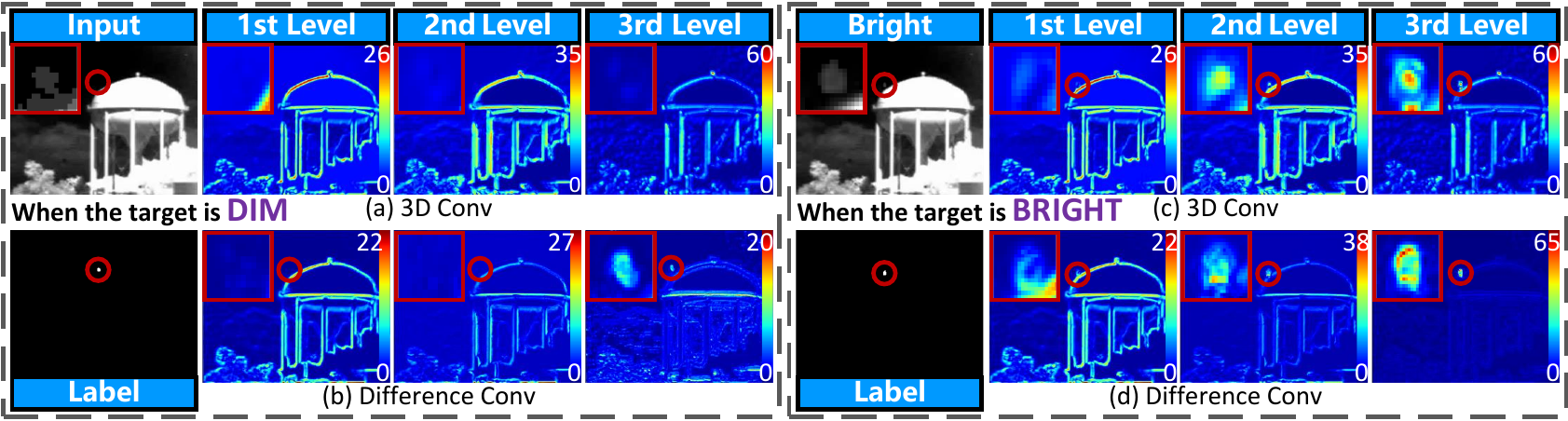}
    \caption{Output feature maps of the four layers (blocks) with and without difference convolutions before TPro. For better visualization, the target area is enlarged in the top-left corner, and some visible on maps are highlighted by a red circle.
    }
    \label{fig:Vis_DifferenceConv}
\end{figure*}
To show the effect of difference convolutions on dim targets, we replaced all difference convolutions in DeepPro-Plus with general 3D convolutions with the same kernel size, and then visualized the output feature maps of each convolution layer. As shown in Fig. \ref{fig:Vis_DifferenceConv}, when the target is dim, the variant of DeepPro-Plus without difference convolutions loses the target from the feature map in all layers, while DeepPro-Plus with difference convolutions can preserve the target in each layer. When the target is bright, the variant can maintain the target and enhance it in the last layer, while DeepPro-Plus with difference convolutions can significantly enhance the target in every layer. Therefore, difference convolutions can highlight targets (especially dim targets) in low levels, which is beneficial for IRST detection.

\begin{table}[!t]
\caption{Detection results achieved by main variants of DeepPro-Plus and difference convolutions on the NUDT-MIRSDT ($SNR\leq 3$) subset and the NUDT-MIRSDT-HiNo dataset.}\label{tab:DifferenceConv}
\centering
\renewcommand{\arraystretch}{1.1}
\begin{threeparttable}
\begin{tabular}{c c c c c c}
\hline
\multicolumn{2}{c}{Variants of DeepPro-Plus} & \multicolumn{2}{c}{$SNR\leq 3$} & \multicolumn{2}{c}{-HiNo} \\ \hline
TD-Conv    & SD-Conv         & $P_d$  & $F_a $                 & $P_d$  & $F_a$ \\ \hline
3D   & 3D              & 97.35  & 1.69                   & 77.15  & 2.15  \\
\ding{51} & 3D         & 98.49  & 1.04                   & 78.31  & 2.61  \\
\ding{51} & TD         & 98.30  & 1.47                   & 70.00  & 5.65  \\
3D    & \ding{51}      & 98.30  & 1.41                   & 71.31  & 2.70  \\
SD    & \ding{51}      & 94.14  & 2.14                   & 68.42  & 1.59  \\ \hline
\ding{51} & \ding{51}  & 99.24  & 1.65                   & 76.23  & 1.69  \\
3D-CDC-ST \cite{yu2021searching} & 3D\tnote{*} & 99.43  & 1.67 & 76.58  & 2.04  \\
3D-CDC-T \cite{yu2021searching} & 3D\tnote{*} & 99.24  & 1.06 & 77.62  & 3.02  \\
3D-CDC-TR \cite{yu2021searching} & 3D\tnote{*} & 99.24  & 2.00 & 76.11  & 2.85  \\ \hline
\end{tabular}
\begin{tablenotes}
\footnotesize
    \item[*] Replace SD-Conv only in the STD-ResBlock.
\end{tablenotes}
\end{threeparttable}
\end{table}

To verify the effectiveness of difference convolutions, we replaced TD-Convs (SD-Convs) in DeepPro-Plus with 3D convolutions or SD-Convs (TD-Convs), and compared the main variants on the NUDT-MIRSDT dataset and the NUDT-MIRSDT-HiNo dataset. The results are shown in Table \ref{tab:DifferenceConv}, from which we can find that using both SD-Convs and TD-Convs makes the network perform better than using none or one of them in weak-noise scenes. For scenes with high-intensity noise, difference convolutions can better balance reducing miss detection and false alarm. In particular, the TD-Conv can partially offset the random fluctuation of noise by adding the changes among multiple frames, and thus reducing the adverse impact of noise. These illustrate that SD-Convs and TD-Convs can retain and enhance dim targets for TPro in different ways, which are both important for the detection.

The STD-ResBlock (a combination of TD-Conv and SD-Conv) approximates a spatial-temporally decoupled version of 3D-CDC \cite{yu2021searching}, where 3D-CDC captures central difference information within local spatial-temporal regions. Its purpose is to retain and enhance target saliency. Based on that, the design enables multiple alternatives for difference convolutions. To validate this, we replaced TD-Conv in the STD-ResBlock with any convolution from the 3D-CDC family and substituted SD-Conv with 3D convolution. The results are presented in Table \ref{tab:DifferenceConv}. It can be observed that the detection performance using arbitrary type of 3D-CDC and 3D convolution combination is close to that of the TD-Conv and SD-Conv combination.

\section{Discussion}
Based on the analysis of failure cases in the main text, we formulate the following considerations for future research.

First, constructing high-quality real-world infrared small and dim target datasets remains crucial, as it facilitates a more comprehensive understanding of the problems and challenges in IRST detection—such as those revealed by the failure cases. Existing datasets generally suffer from limited variations in targets and scenes, whereas real-world scenarios are often more complex. This condition is fully demonstrated in the description of the SatVideoIRSDT dataset \cite{li2025dataset, li2026satvideodataset}.

Second, the efficient and effective utilization of longer temporal information is of significant importance. Although the exploration on the length of temporal profile information for the DeepPro (Section 4.3.2 in the main text) shows that longer temporal profile information did not yield significant performance improvement, theoretically, longer (more complete) sequences enable more accurate predictions. For instance, in the second failure case, access to longer temporal information would confirm that the currently stationary target exhibit motion at other time instances. Thus, exploring the efficient and effective ways to utilize longer temporal information may lead to unexpected breakthroughs.

\bibliographystyle{IEEEtran}
\bibliography{ref_abb}

\newpage

\vfill